\documentclass{article}

\usepackage[preprint]{neurips_2020}
\usepackage[utf8]{inputenc} 
\usepackage[T1]{fontenc}    
\usepackage{hyperref}       
\usepackage{url}            
\usepackage{booktabs}       
\usepackage{amsfonts}       
\usepackage{nicefrac}       
\usepackage{microtype}      
\usepackage{amsmath, amssymb}
\usepackage{amsthm}

\usepackage{subfigure}
\usepackage{graphicx}
\usepackage{float}

\usepackage{microtype}
\usepackage{booktabs} 
\usepackage{multirow} 
\usepackage{makecell} 

\usepackage{hyperref}

\usepackage{amsmath}
\usepackage{amssymb}
\usepackage{mathtools}
\usepackage{amsthm}

\usepackage[capitalize,noabbrev]{cleveref}
\usepackage[OT2, T1]{fontenc}
\theoremstyle{plain}
\newtheorem{theorem}{Theorem}[section]

\theoremstyle{definition}

\theoremstyle{remark}

\usepackage[textsize=tiny]{todonotes}

\allowdisplaybreaks[1] 

\title{Trained Model in Supervised Deep Learning is a Conditional Risk Minimizer}

\begin{document}

\author{%
  Yutong Xie\\
  Academy for Advanced Interdisciplinary Studies\\
  Peking University\\
  Beijing, 100871 \\
  \And
  Dufan Wu\\
  MGH/BWH Center for Advanced Medical Computing and Analysis\\ Gordon Center for Medical Imaging, Department of Radiology\\
  Massachusetts General Hospital and Harvard Medical School\\
  Boston, MA 02114\\
  \And
  Bin Dong\\
  Beijing International Center for Mathematical Research\\
  Peking University\\
  Beijing, 100871\\
  \texttt{dongbin@bicmr.pku.edu.cn}
  \And
  Quanzheng Li \\
  MGH/BWH Center for Advanced Medical Computing and Analysis\\ Gordon Center for Medical Imaging, Department of Radiology\\
  Massachusetts General Hospital and Harvard Medical School\\
  Boston, MA 02114\\
  \texttt{li.quanzheng@mgh.harvard.edu} \\
}

\maketitle

\begin{abstract}
We proved that a trained model in supervised deep learning minimizes the conditional risk for each input (Theorem 2.1). This property provided insights into the behavior of trained models and established a connection between supervised and unsupervised learning in some cases. In addition, when the labels are intractable but can be written as a conditional risk minimizer, we proved an equivalent form of the original supervised learning problem with accessible labels (Theorem 2.2). We demonstrated that many existing works, such as Noise2Score, Noise2Noise and score function estimation can be explained by our theorem. Moreover, we derived a property of classification problem with noisy labels using Theorem 2.1 and validated it using MNIST dataset. Furthermore, We proposed a method to estimate uncertainty in image super-resolution based on Theorem 2.2 and validated it using ImageNet dataset. Our code is available on github\footnote{\hyperlink{https://github.com/Theodore-PKU/theorem-1-verification-python.}{https://github.com/Theodore-PKU/theorem-1-verification-python.}}.
\end{abstract}

\section{Introduction}
\label{sec:introduction}

Supervised learning is one of the most widely-used paradigm in deep learning, where the network tries to learn a mapping from inputs to labels. In addition to the mapping itself, people are also interested in its properties, particularly under a posterior distribution, which is usually non-trivial and intractable. A common example is the supervised learning with noisy label. In practice it is almost impossible to get a clean label without any noise. Therefore, it is important to analyse the network's behavior trained under noisy labels \citep{chen2019understanding}, and then design proper loss functions to eliminate the noises' influence \citep{lehtinen2018noise2noise}. Another example is the uncertainty estimation, where we are interested in predicting the conditional variance of network output \citep{kendall2017what}. 

To describe how the trained network will behave when the dependency between labels and inputs is no longer deterministic but follows a non-trivial posterior distribution, we proposed a theorem (Theorem 2.1) which shows the trained networks will minimize the conditional risk with given network input. Based on this property, we can design a class of supervised learning methods with more accessible labels to solve challenging learning problems that have intractable labels (Theorem 2.2). A stronger conclusion was also given when the training loss is a L2 norm. 

Many existing works related to noisy and intractable labels are special cases of our theory. Theoretical relationship was proved between the proposed methods and Noise2Noise, Noise2Score, and score function estimation \citep{lehtinen2018noise2noise, kim2021noise2score, vincent2011connection}. Besides theoretical analysis, we further experimentally validated the proposed theorems on two problems: (1) MNIST classification with noisy labels in which we compared the network's results with theoretically derived results; (2) uncertainty estimation of single image super-resolution in which we compared the uncertainty learned based on our theory with the one calculated by sampling the DDPM-SR model \citep{dhariwal2021diffusion, ho2020denoising}.

Our main contributions are: (1) we proved a trained model in supervised deep learning will minimize the conditional risk, which provided insights into the behavior of trained model and connected supervised and unsupervised learning in some cases; (2) we proposed a new training strategy for supervised learning when labels are intractable, and proved the equivalence between the original learning problem and the new training strategy. (3) Our theorems explained many existing works and was applied to analyze image classification models trained with noisy labels and uncertainty estimation of single image super-resolution.

This paper is organized as follows. In \cref{sec:theory} we give the main theorems, build theoretical relationship between proposed methods and existing works. We introduce the details of two verification tasks in \cref{sec:verification_tasks}. Experiment results of them are shown in \cref{sec:experiments}. At last, we show the related works in \cref{sec:related_works} and conclusion in \cref{sec:conclusion}.

\begin{table}[ht]
\caption{The concrete meanings of the same notations in different scenarios.}
\label{tab:notation}
\begin{center}
\begin{small}
\begin{sc}
\begin{tabular}{m{75pt} m{75pt} m{75pt} c m{55pt}}
\toprule
        Scenario & Input: $y$ &  Variable: $x$ & Label: $g(x, y)$ & Loss: $L$ \\
        \midrule
        \hyperref[classification]{Image classification} & image & one-hot vector & $x$ & cross-entropy \\
        \hyperref[noise2score]{Noise2Score} & noisy image & clean image & $x$ & L2 norm \\
        \hyperref[noise2noise]{Noise2Noise} & noisy image & noisy image & $x$ & L2 norm \\
        \hyperref[score]{Score function estimation} & random variable & random variable & $\nabla_y \log p(y \mid x)$ & L2 norm \\
        \hyperref[uncertainty]{Uncertainty estimation in super-resolution} & low-resolution image & high-resolution image & \makecell[c]{$x$ or \\ $\left(x - \mathbb{E}_{x\mid y}[x] \right)^2$} & L2 norm \\
\bottomrule
\end{tabular}
\end{sc}
\end{small}
\end{center}
\end{table}

\section{Theory}
\label{sec:theory}

\subsection{Notation}
\label{sec:notation}

We will generally use $y$ for input and $g(x, y)$ for labels. We use more general form $g(x, y)$ instead of $x$ to denote labels in order to cover a wide range of applications in this work. In most cases (particularly related to Theorem 2.1) $g(x, y) = x$ and $x$ is the label. In some cases related to Theorem 2.2 $g(x, y)$ could be a more complicated function. To avoid confusion, the definition of major notations under different scenarios is summarized in \cref{tab:notation}, which will also be further explained again in each section.

\subsection{Understanding Model Trained in Supervised Learning}
\label{sec:theorem_left2right}

In this section we consider a general case of supervised learning in which the labels follow a distribution conditioned on the inputs. To understand what the model will learn under this scenario, we propose the following  \cref{theorem:left2right}:

\begin{theorem}
\label{theorem:left2right}
Assuming that $X$, $Y$ and $W$ are measurable spaces; $x$ and $y$ are random variables defined in $X$ and $Y$respectively; $g(x, y): X \times Y \rightarrow W$ is a measurable function; $L(a, b): W \times W \rightarrow \mathbb{R}$ is a loss function which satisfies $L(a, b) \geq L(a, a)$ and $L(a, b) \geq L(b, b)$;
$f(y; \theta): Y \rightarrow W$ is a model parameterized by $\theta$ and for any measurable function $\tilde{f}(y): Y \rightarrow W$, there exists some $\theta$ such that $f(\cdot; \theta) =\tilde{f}(\cdot)$. 
Then the optimal solution $\theta^*$ to the following problem:
\begin{equation}
\label{eq:left_training}
    \min_{\theta} \mathbb{E}_{x, y} \left[L \left(f\left(y; \theta \right), g\left(x, y\right)\right) \right],
\end{equation}
satisfies that $f(y; \theta^*) = z^*(y)$
where 
\begin{equation}
\label{eq:z_star}
    z^*(y) = \arg\min_{z} \mathbb{E}_{x \mid y} \left[ L \left(z, g \left(x, y \right) \right) \right]. 
\end{equation}
\end{theorem}
The proof of \cref{theorem:left2right} is in \cref{sec:proof_left2right}. In addition, a more generalized form of \cref{theorem:left2right} without the parameterized model is given in \cref{sec:another_version_left2right}.

\cref{theorem:left2right} states that for a supervised learning problem $y \rightarrow g(x,y)$, the learned model $f(y;\theta^*)$ is equal to $z^*(y)$, a conditional risk minimizer. $z^*(y)$ is selected so that the loss between $z^*(y)$ and the labels $g(x,y)$ is minimized over the conditional distribution $x \mid y$. \cref{theorem:left2right} will be trivial if there exists a mapping from $y$ to $x$, and the trained model will be a data fitting process from $y$ to $g(x, y)$ in this case (see \cref{sec:proof_trivial_conclusion}).

Noting that $z^*(y)$ is a function of $y$ and is connected to the conditional distribution $x \mid y$ instead of specific labels $g(x,y)$, \cref{theorem:left2right} implies two applications. (1) \emph{analyzing the behavior of trained model}: when the label $g(x,y)$ follows a probability distribution (e.g. noisy label), we can analyze the property of the trained model $f(y; \theta^*)$ through $z^*(y)$; (2) \emph{connecting supervised and unsupervised learning} : if $z^*(y)$ can be directly estimated from the dataset of inputs $\left\{ y \right\}$, the model obtained by supervised learning can be trained from $\left\{ y \right\}$ using unsupervised learning without labels $g(x,y)$.

\subsubsection{Image classification with noisy labels}
\label{classification}
Image classification with noisy labels is an example of the first application. In this case, $y$ represents the image to be classified, $x$ is the one-hot vector of image class, $g(x,y)=x$ is the label, and $L$ is cross-entropy loss. When labels are correct, there exists a mapping from $y$ to $x$ and we can map $y$ to the right category based on the optimal model $f(y; \theta^*)$. However, if the annotation of dataset is inaccurate and noisy, \cref{theorem:left2right} can be used to analyze the behavior of $f(y;\theta^*)$ through $z^*(y)$. Given the noisy label distribution $p(x \mid y)$, one can calculate $z^*(y)$ according to \cref{eq:z_star}. Examples of different label noise distributions will be given in \cref{sec:classification_experiment_design} with their corresponding experimental results in \cref{sec:mnist_experiment}.

\subsubsection{Noise2Score}
\label{noise2score}
Noise2Score \citep{kim2021noise2score} is an example of the second application. It is an unsupervised learning method for image denoising task. In this scenario, $x$ is the clean image, $y=x+n$ is the noisy image with noise $n$,  $g(x,y) = x$ is the label, and $L$ is L2 norm loss. Supervised learning by \cref{eq:left_training} leads to $z^*(y) = \mathbb{E}_{x\mid y}[x]$\footnote{The proof is in \cref{sec:proof_denoising}.}. If $n$ follows an exponential family distribution, according to the Tweedie's formula \citep{efron2011tweedie, robbins2020empirical}, $\mathbb{E}_{x \mid y}\left[ x \right]$ has a closed form and is related to $\nabla_{y} \log p(y)$, the score function of $y$.
For instance, $\mathbb{E}_{x \mid y}\left[ x \right] = y + \sigma^2 \nabla_{y} \log p(y) $ if $n \sim \mathcal{N}(0, \sigma^2 I)$. In Noise2Score, $\nabla_{y} \log p(y)$ is estimated by AR-DAE \citep{lim2020ar} where only $\left\{ y \right\}$ is used. Therefore, $z^*(y)$ can be learned using only $\left\{ y \right\}$ and thus the original supervised learning problem is converted to unsupervised learning.

\subsection{Equivalent Training Target for Intractable Labels}
\label{sec:theorem_right2left} 
There are a class of deep learning problems that could be considered as to train a network to predict the minimal conditional risk $z^*(y)$. However, $z^*(y)$ could be very difficult to compute or even intractable. In this case we propose the following \cref{theorem:right2left} which shows that one can train the model using labels $g(x,y)$ instead. Furthermore, a stronger conclusion on the equivalency can be drawn under L2 norm. 

\begin{theorem}
\label{theorem:right2left}
Given all the assumptions in \cref{theorem:left2right} and a target function $z^*(y): Y \rightarrow W$ which satisfies that:
\begin{equation}
\label{eq:condition_for_right2left}
    z^*(y) = \arg\min_{z} \mathbb{E}_{x \mid y} \left[ L \left(z, g \left(x, y \right) \right) \right],
\end{equation}
then the following equation holds:
\begin{equation}
\label{eq:right2left}
\begin{split}
    \arg \min_{\theta} \mathbb{E}_{y} \left[ L (f(y; \theta), z^*(y) \right]  = \arg \min_{\theta} \mathbb{E}_{x, y} \left[ L\left( f(y; \theta), g(x, y) \right) \right].
\end{split}
\end{equation}
If $L$ is L2 norm loss, we further have that $z^*(y) = \mathbb{E}_{x \mid y}\left[ g(x, y)\right]$ and for any model $f(y; \theta):Y \rightarrow W$, 
the following equation holds:
\begin{equation*}
\begin{split}
    \mathbb{E}_{y}  \left\|f(y; \theta) - z^*(y) \right\|_2^2 + C
    = \mathbb{E}_{x, y} \left\| f(y; \theta) - g(x, y) \right\|_2^2 ,
\end{split}
\end{equation*}
where $C$ is a constant. Therefore, the following two optimization problems are equivalent:
\begin{equation}
\label{eq:right2left_l2_norm_case_optimziation_problem}
\begin{split}
    \min_{\theta} \mathbb{E}_{y} \left\|f(y; \theta) - z^*(y) \right\|_2^2
    \Longleftrightarrow \min_{\theta} \mathbb{E}_{x, y} \left\| f(y; \theta) - g(x, y) \right\|_2^2 .
\end{split}
\end{equation}
\end{theorem}
The proof of \cref{theorem:right2left} is in \cref{sec:proof_right2left}. 

\cref{eq:right2left} implies that one can achieve the same model by using either $z^*(y)$ or $g(x,y)$ as labels, which is extremely powerful when $z^*(y)$ is intractable. For example, $z^*(y)$ needs to be calculated by averaging over the conditional distribution $x \mid y$. Usually we only have one sample of $x$ for each $y$, which prevents the direct calculation of $z^*(y)$. \cref{theorem:right2left} tells us that we can use $g(x,y)$ as labels, which relies on single pairs of $x$ and $y$, instead. When L2 norm is used as loss function, it is a special case of \cref{eq:right2left}. Under L2 norm, the two object functions are differed by merely a constant. Hence, not only the final optimal solutions will be the same, but also any intermediate solutions during the optimization, e.g. if the training is early stopped. 

In the following subsections, we will show that Noise2Noise and current method to estimate score function are special cases of \cref{theorem:right2left}; we will also apply \cref{theorem:right2left} to compute uncertainty of single image super-resolution.

\subsubsection{Noise2Noise}
\label{noise2noise}
Noise2Noise \citep{lehtinen2018noise2noise} claims that one can train a denoising model by mapping a noisy image to another noisy image, where the two images share the same content but independent noises. In this case, let $y = s + n_1$ and $x = s + n_2$, 
where $s$ is the underlying clean image, $n_1$ and $n_2$ are two independent noises. Let $g(x,y) = x$ and $L$ be the L2 norm, we have:
\begin{equation}
    z^*(y) = \arg\min_{z} \mathbb{E}_{x \mid y} \left\| z - x\right\|_2^2 = \mathbb{E}_{x \mid y}\left[x\right].
\end{equation}
From \cref{theorem:right2left}, we have:
\begin{equation}
\label{eq:denoising_equivalent_1}
\begin{split}
    \min_{\theta} \mathbb{E}_{x, y} \left\| f(y; \theta) - x \right\|_2^2 
    \Longleftrightarrow \min_{\theta} \mathbb{E}_{y} \left\| f(y; \theta) - \mathbb{E}_{x \mid y} \left[x\right] \right\|_2^2.
\end{split}
\end{equation}
On the other hand, let $x = s$ and we can similarly get that:
\begin{equation}
\label{eq:denoising_equivalent_2}
\begin{split}
    \min_{\theta} \mathbb{E}_{s, y} \left\| f(y; \theta) - s \right\|_2^2
    \Longleftrightarrow  \min_{\theta} \mathbb{E}_{y} \left\| f(y; \theta) - \mathbb{E}_{s \mid y} \left[s\right] \right\|_2^2.
\end{split}
\end{equation}
If $n_2$ is zero mean, we have
\begin{equation}
\label{eq:noise2noise}
    \mathbb{E}_{x\mid y} \left[ x \right] = \mathbb{E}_{s \mid y} \left[ s\right].
\end{equation}
The proof to equations (\ref{eq:denoising_equivalent_1}), (\ref{eq:denoising_equivalent_2}), and (\ref{eq:noise2noise}) are given in \cref{sec:proof_noise2noise}. Combining equations (\ref{eq:denoising_equivalent_1}), (\ref{eq:denoising_equivalent_2}), and (\ref{eq:noise2noise}) gives the equivalence between training with $x$ and $s$, which is the Noise2Noise training:
\begin{equation*}
\begin{split}
    \min_{\theta} \mathbb{E}_{x, y} \left\| f(y; \theta) - x \right\|_2^2
    \Longleftrightarrow \min_{\theta} \mathbb{E}_{s, y} \left\| f(y; \theta) - s \right\|_2^2
\end{split}
\end{equation*}
In summary, \cref{theorem:right2left} helped to build the connection between the intractable target $s$ and the noisy labels $x$ through $z^*(y)$, which explained how the Noise2Noise training works.

\subsubsection{Score Function Estimation}
\label{score}
Estimation of the score function is essential in some applications such as image denoising and generation \citep{kim2021noise2score, song2020score}. For a random variable $y$, its score function is defined as $\nabla_{y} \log p(y)$. Let $y$ and $x$ be two random variables, \cite{vincent2011connection} has proved that:
\begin{equation}
\label{eq:score_function_estimation}
\begin{split}
    \min_{\theta} \mathbb{E}_{y} \left[\left\| f(y; \theta) - \nabla_{y} \log p(y) \right\|_{2}^2 \right] 
    \Longleftrightarrow  \min_{\theta} \mathbb{E}_{x, y} \left[\left\| f(y; \theta) - \nabla_{y} \log p(y \mid x ) \right\|_{2}^2 \right].
\end{split}
\end{equation}
$\nabla_{y} \log p(y)$ is usually inaccessible but $\nabla_{y} \log p(y \mid x)$ can sometimes be computed by selecting appropriate $x$. \cref{{eq:score_function_estimation}} indicates that the score function $\nabla_{y} \log p(y)$ can be estimated using the substituted labels $\nabla_{y} \log p(y \mid x )$; this equation is the basis of all score function based methods. 

\cref{theorem:right2left} provides a simple alternative proof to \cref{eq:score_function_estimation}. Let $g(x,y) = \nabla_{y} \log p(y \mid x )$ and $L$ be the L2 norm loss, we have
\begin{align}
\label{eq:score_function_estimation_key_part}
    z^*(y) &= \arg\min_z\mathbb{E}_{x \mid y}\| z - g(x,y)  \|_2^2 =\mathbb{E}_{x \mid y}\left[g(x, y) \right] \notag \\
    &=\mathbb{E}_{x \mid y}\left[\nabla_{y} \log p(y \mid x ) \right] =\nabla_{y} \log p(y).
\end{align}

Hence, $z^*(y)$ provides the left hand side of \cref{eq:score_function_estimation} and $g(x,y)$ provides its right hand side, and the equivalence in \cref{eq:score_function_estimation} is proved according to \cref{theorem:right2left}. The proof to \cref{eq:score_function_estimation_key_part} is give in \cref{sec:proof_score_function}.

\subsubsection{Uncertainty Estimation of Single Image Super-Resolution}
\label{uncertainty}
\cref{theorem:right2left} can be further applied to the uncertainty estimation of single image super-resolution. In this case, $x$ is a high-resolution image and $y$ is the corresponding low-resolution image. Let $z^*(y)$ be some data uncertainty we want to estimate over the conditional distribution $x \mid y$, which is intractable because the conditional distribution is unknown. However, if we can find suitable $g(x, y)$ and $L$ such that $z^*(y)$ satisfies \cref{eq:condition_for_right2left}, $z^*(y)$  can be estimated using $g(x,y)$ as the training label. For example, for pixel-wise variance, our target $z^*(y)$ is $\mathbb{E}_{x \mid y}\left( x - \mathbb{E}_{x \mid y }[x] \right)^2$, which satisfies \cref{eq:condition_for_right2left} when $g(x, y) = \left( x - \mathbb{E}_{x \mid y }[x] \right)^2$ and $L$ is the L2 norm loss. More details are in \cref{sec:uncertainty_experiment_design}. 



\subsection{Summary}
\label{sec:theorem_summary}

So far, we introduced our main theoretical results and their relation to previous works. \cref{theorem:left2right} characterizes the property of trained model in supervised learning problem given by \cref{eq:left_training}. One can analyze the behavior of the trained model through $z^*(y)$, which gives insights into noisy label problems, as demonstrated in \cref{sec:classification_experiment_design}. Based on \cref{theorem:left2right}, a supervised learning problem can also be converted to unsupervised learning problem if one can fit $z^*(y)$ from the inputs $\{y\}$ as shown in Noise2Score.

\cref{theorem:right2left} states the equivalence between the two training approaches, where one can convert the intractable labels $z^*(y)$ to the tractable ones $g(x,y)$. We found that it explains previous works including Noise2Noise and score function estimation. We further applied it to uncertainty estimation of single image super-resolution as in \cref{sec:uncertainty_experiment_design}.

The successes of previous works, including Noise2Score, Noise2Noise, and score function estimation confirmed the proposed theorems. In the following sections, we will further verify our theorems and demonstrate their utility for practical applications by carrying out experiments on two tasks: image classification with noisy labels and uncertainty estimation of single image super-resolution.

\section{Two Verification Tasks as Examples of Practical Applications}
\label{sec:verification_tasks}
In this section, we discuss two verification tasks in detail, which are examples of practical applications of our theorems: (1) image classification with noisy labels, where \cref{theorem:left2right} is used to analyze the trained model's behavior through $z^*(y)$; (2) uncertainty estimation of single image super-resolution, where \cref{theorem:right2left} is used to perform supervised learning when the labels are intractable.

\subsection{Image Classification with Noisy Labels}
\label{sec:classification_experiment_design}
Suppose there are $n$ classes for classification, $y$ is the image to be classified, and $x$ represents an $n$-dimensional one-hot vector of image class $c$ where $c \in \left\{1, 2, \ldots, n \right\}$ and the components of $x$ are all $0$ but $1$ at the position $c$. Usually, $g(x, y) = x$ is the label, $L$ is cross-entropy loss $\mathrm{CE}\left(\cdot, \cdot \right)$, and the classification model $f(y; \theta)$ is trained by:
\begin{equation}
\label{eq:ce_loss}
    \min_{\theta} \mathbb{E}_{{x}, y} \left[ \mathrm{CE}\left(f(y; \theta), {x} \right) \right].
\end{equation} 
The output of $f(y; \theta)$ is also an $n$-dimensional vector representing the predictive probabilities for each class. During inference, usually the class with the highest predictive probability will be chosen as the classification result. If the labels are correct, the optimal model $f(y; \theta^*)$, i.e. $z^*(y)$, will map $y$ to the one-hot vector $x$ and predict the right category\footnote{Practically, because the output of $f$ is usually the result of $\mathrm{softmax}$ operation, the predictive probabilities are all larger than $0$. Therefore, output of $f(y; \theta^*)$ will be very close to $x$.}.

When the labels are noisy, the annotations of $y$ may be wrong and is sampled from $\left\{1,2,\ldots,n\right\}$ following a discrete distribution. Without loss of generality, we assume that the noise distribution is dependent on $y$. We denote the distribution as an $n$-dimensional vector $q_y$, which satisfies that $\sum_{i=1}^{n} q_{y,i} = 1$ and $q_{y,i} \geq 0$. $q_{y,i}$ is the $i$th element of $q_y$ and represents the probability that $y$ is labeled as class $i$. $z^*(y)$ in \cref{theorem:left2right} can be calculated as:
\begin{equation}
\label{eq:minimul_ce_loss}
    z^*(y) = \arg \min_{{z}} \mathbb{E}_{x\mid y} \left[ \mathrm{CE}\left( {z}, {x} \right) \right] = {q}_y. 
\end{equation}
The proof of \cref{eq:minimul_ce_loss} is in \cref{sec:proof_classification_noise_label}. \cref{eq:minimul_ce_loss} shows that the optimal model $f(y; \theta^*)$ trained by noisy labels will map $y$ to $q_y$. In other word, the predictive probability distribution of $f(y; \theta^*)$ is exactly the label noise distribution. Assuming the model performs the same on the testing and training datasets, and the class with the highest predictive probability is used, \cref{eq:minimul_ce_loss} indicates that different label noise distributions ${q}_y$ may affect the testing accuracy differently.

We consider three different label noise distribution $q_y$ to experimentally verify \cref{theorem:left2right}. Let ${x}^{(i)}$ represent the one-hot vector corresponding to class $i$. Given $y$ and its correct class $c$, we define the three types of label noise as follows:
\begin{enumerate}
    \item \emph{Uniform noise}: there is equal possibility to misclassify $y$ to any incorrect class,
    \begin{equation}
        q_{y, i} = p({x}^{(i)} \mid y) = \begin{cases}
        \alpha & \text{ if } i = c, \\ 
        \frac{1 - \alpha}{n-1} & \text{ if } i \neq c.
        \end{cases}
    \end{equation}
    where $0 < \alpha < 1$ is the probability of correct classification.
    \item \emph{Biased noise}: $y$ may only be misclassified to a neighboring class,
    \begin{equation}
        q_{y, i} = p({x}^{(i)} \mid y) = \begin{cases}
        \alpha & \text{ if } i = c, \\ 
        1 - \alpha & \text{ if } i= c_{+1}, \\
        0 & \text{ otherwise }.
        \end{cases}
    \end{equation}
    where $c_{+1} = (c+1) \mod n$ and $0 < \alpha < 1$ is the probability of correct classification.
    \item \emph{Generated noise}: assuming there is another trained classification model $M(y)$ whose output is also a predictive probabilities for each class, ${q}_{y}$ is defined as follows,
    \begin{equation}
    \label{eq:generated_noise}
        q_{y, i} = p({x}^{(i)} \mid y) =  \begin{cases}
        \beta M(y)_{c} & \text{ if } i = c, \\ 
        \frac{1- \beta M(y)_{c}}{1 - M(y)_{c}}M(y)_{i} & \text{ if } i\neq c.
        \end{cases}
    \end{equation}
    where $M(y)_{i}$ represents the predictive probability of class $i$ and $0< \beta < 1$. We assume that $\mathrm{softmax}$ operation is used in $M(y)$, therefore $M(y)_{c} < 1$ and \cref{eq:generated_noise} is well defined.
\end{enumerate}
Uniform noise and biased noise are class-dependent whereas generated noise is instance-dependent, i.e. $q_y$ is different for different samples. Experimental details and results are shown in \cref{sec:mnist_experiment}.

\subsection{Uncertainty Estimation of Single Image Super-resolution}
\label{sec:uncertainty_experiment_design}
Let $x$ be a high-resolution image and $y$ be the corresponding low-resolution image. They satisfy $y = Ax$ where $A$ is a down-sampling operator. Given $y$ there are many $x$ that share the same low-resolution image $y$ because $A$ is underdetermined. Hence, it is worth to measure the uncertainty of the super-resolved images given $y$ to estimate how reliable it is.  The uncertainty here is defined as a value that can be calculated from the statistical distribution $x\mid y$. For example, the most widely used pixel-wise variance is defined as:
\begin{equation}
\label{eq:uncertainty_pixel_var}
    \mathrm{Var}_{x \mid y}\left[ x \right] = \mathbb{E}_{x \mid y}\left( x - \mathbb{E}_{x \mid y }[x] \right)^2
\end{equation}
Normally the uncertainty cannot be calculated directly because $x\mid y$ is unknown. An alternative method is the Monte Carlo approach where a generative super-resolution model is used to generate multiple samples from $x \mid y$ to calculate the statistics \citep{dhariwal2021diffusion}. However, it is time-consuming and the accuracy extremely dependents on the performance of generative model.

\cref{theorem:right2left} can be used to solve the uncertainty estimation problem. As discussed in \cref{uncertainty}, let $z^*(y) = \mathrm{Var}_{x \mid y}\left[ x \right]$, $g(x, y) = \left(x - \mathbb{E}_{x\mid y}\left[ x \right]\right)^2$, and $L$ be the L2 norm loss, we can train a model to learn the pixel-wise variance using:
\begin{equation}
\label{eq:sr_overall_train}
    \min_{\theta} \mathbb{E}_{x,y}\| f(y; \theta) - \left( x - \mathbb{E}_{x \mid y }[x] \right)^2 \|_2^2.
\end{equation}
Note that $\mathbb{E}_{x \mid y }[x]$ is still intractable but we can apply \cref{theorem:right2left} again to train a mean-estimating model: 
\begin{equation}
\label{eq:sr_mean_train}
    \min_{\theta_1} \mathrm{E}_{x, y}\left\| f_\mathrm{mean}(y; \theta_1) - x \right\|_2^2,
\end{equation}
and $f_\mathrm{mean}(y,\theta_1)$ will fit to $\mathbb{E}_{x \mid y }[x]$. Then we can substitute $\mathbb{E}_{x \mid y }[x]$ in \cref{eq:sr_overall_train} with $f_\mathrm{mean}(y,\theta_1)$ and train the variance-estimating model:
\begin{equation}
\label{eq:sr_var_train}
    \min_{\theta_2} \mathrm{E}_{x, y}\left\| f_{\mathrm{var}}(y; \theta_2) - \left( x -  f_{\mathrm{mean}}(y; \theta_1\right))^2 \right\|_2^2.
\end{equation}
Details and results of relevant experiments are given in \cref{sec:imagenet_experiment}. Some other statistics on $x \mid y$ as listed in \cref{sec:details_sr} \cref{tab:examples_of_statistics} may be estimated in the similar way, but they are remained for future works.

\section{Experiments}
\label{sec:experiments}
The notations in \cref{sec:mnist_experiment} and \cref{sec:imagenet_experiment} are the same to \cref{sec:classification_experiment_design} and \cref{sec:uncertainty_experiment_design}, respectively.

\subsection{Image Classification with Noisy Labels}
\label{sec:mnist_experiment}

\subsubsection{Experiment Setup}

We conducted the experiment on MNIST dataset \citep{lecun1998mnist}, which contains 60k images of 10 digits, from $0$ to $9$. The noisy labels were constructed as follows: for each $y$, we sampled $c$ from $q_y$ once and $(y, c)$ was used as a training pair. Several groups of the hyperparameters $\alpha$ and $\beta$ were selected to model different noise levels, where in each group different noise types shared the same ratio of correct labels $\eta$. To effectively match the noise levels, we set $\beta$ from $0.2$ to $0.9$ with an interval of $0.1$. For each $\beta$, we first calculated the correct label rate $\eta$ of the generated noise , then set $\alpha = \eta$ because $\mathbb{E}[\eta] = \alpha$ for uniform and biased noise. The specific values of $\alpha$ and $\beta$ are listed in \cref{sec:details_mnist} \cref{tab:noise_type}.

A CNN with two convolutional layers and three fully-connected layers was used as the classification model $f(y; \theta)$. We trained $f(y; \theta)$ for 20k iterations with a batch size of 128. Adam optimizer with learning rate of $0.0001$ was used. $M$ was trained in the same manner but without label noise. The models were saved every 2k iterations and we selected the ones with the minimal loss as final models, since we focus on the optimal model of \cref{eq:ce_loss}. 

To verify \cref{eq:minimul_ce_loss}, we need to evaluate the closeness between the predictive probabilities of trained model $f(y; \theta^*)$ and ${q}_{y}$. We computed the average of cross-entropy loss over the training set:
\begin{equation}
    \overline{\mathrm{CE}}_f = \frac{1}{N} \sum_{i=1}^{N} \mathrm{CE}\left(f(y; \theta^*), q_y \right),
\end{equation}
and compared it to the theoretical minimum value
\begin{equation}
    \overline{\mathrm{CE}}_q = \frac{1}{N} \sum_{i=1}^{N} \mathrm{CE}\left(q_y, q_y \right),
\end{equation}
where $N$ is the size of the training set. In addition, we use the accuracy of classification over test set to evaluate and compare the performance of models trained by different types of noisy labels.

\subsubsection{Experiment Results}
\label{sec:mnist_experiment_result}

\begin{figure}[t]
\begin{center}
    \centerline{\includegraphics[width=0.75\columnwidth]{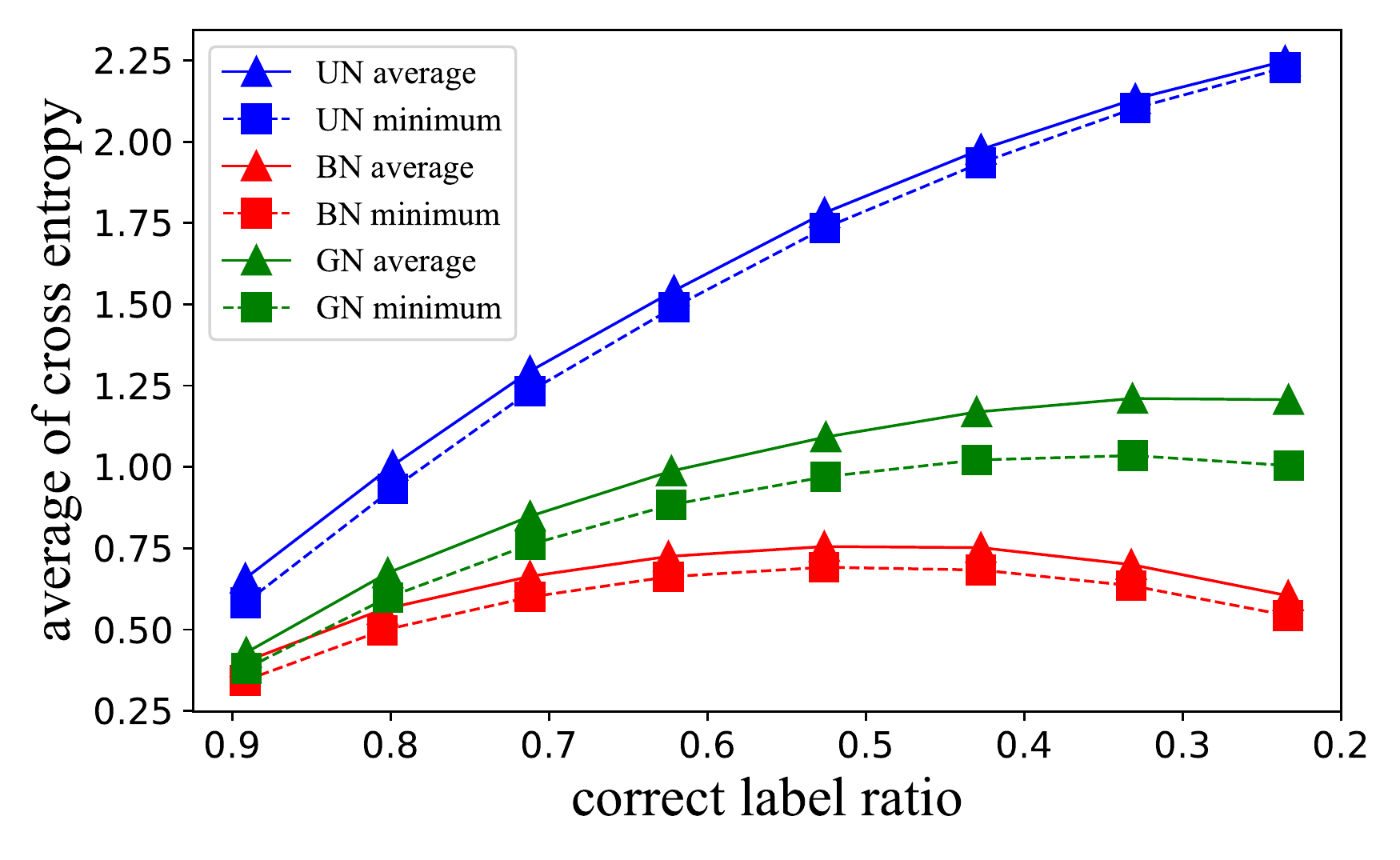}}
    \caption{The comparison between $\overline{\mathrm{CE}}_f$ and $\overline{\mathrm{CE}}_q$ for three types of label noise. The solid lines represent $\overline{\mathrm{CE}}_f$ and  the dash lines represent $\overline{\mathrm{CE}}_q$. UN: uniform noise; BN: biased noise; GN: generated noise. }
    \label{fig:avg_ce}
\end{center}
\end{figure}

\begin{figure}[t]
\begin{center}
    \subfigure[average for UN]{
    \centering{\includegraphics[width=0.42\columnwidth]{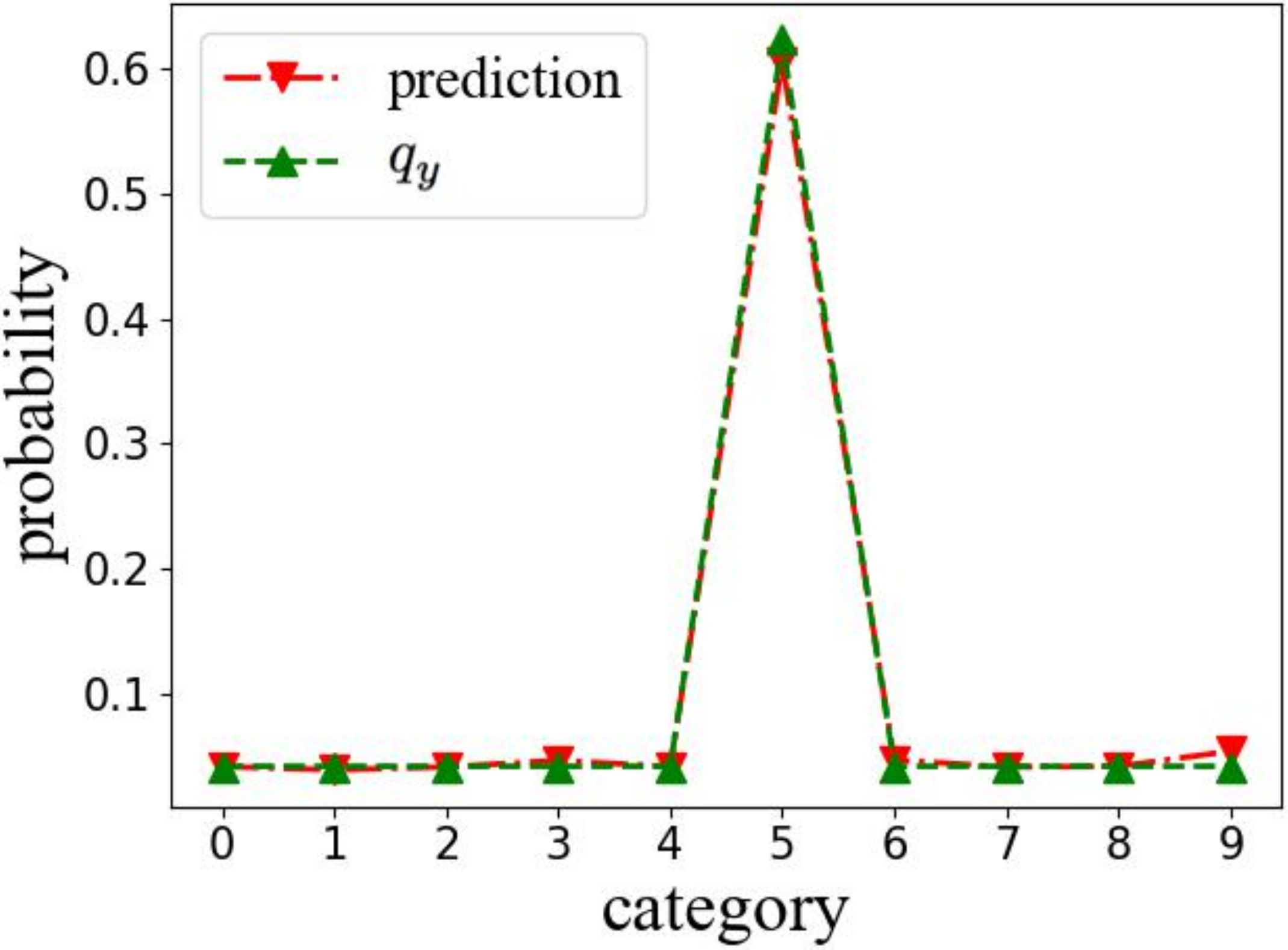}}}
    \subfigure[an instance for UN]{
    \centering{\includegraphics[width=0.42\columnwidth]{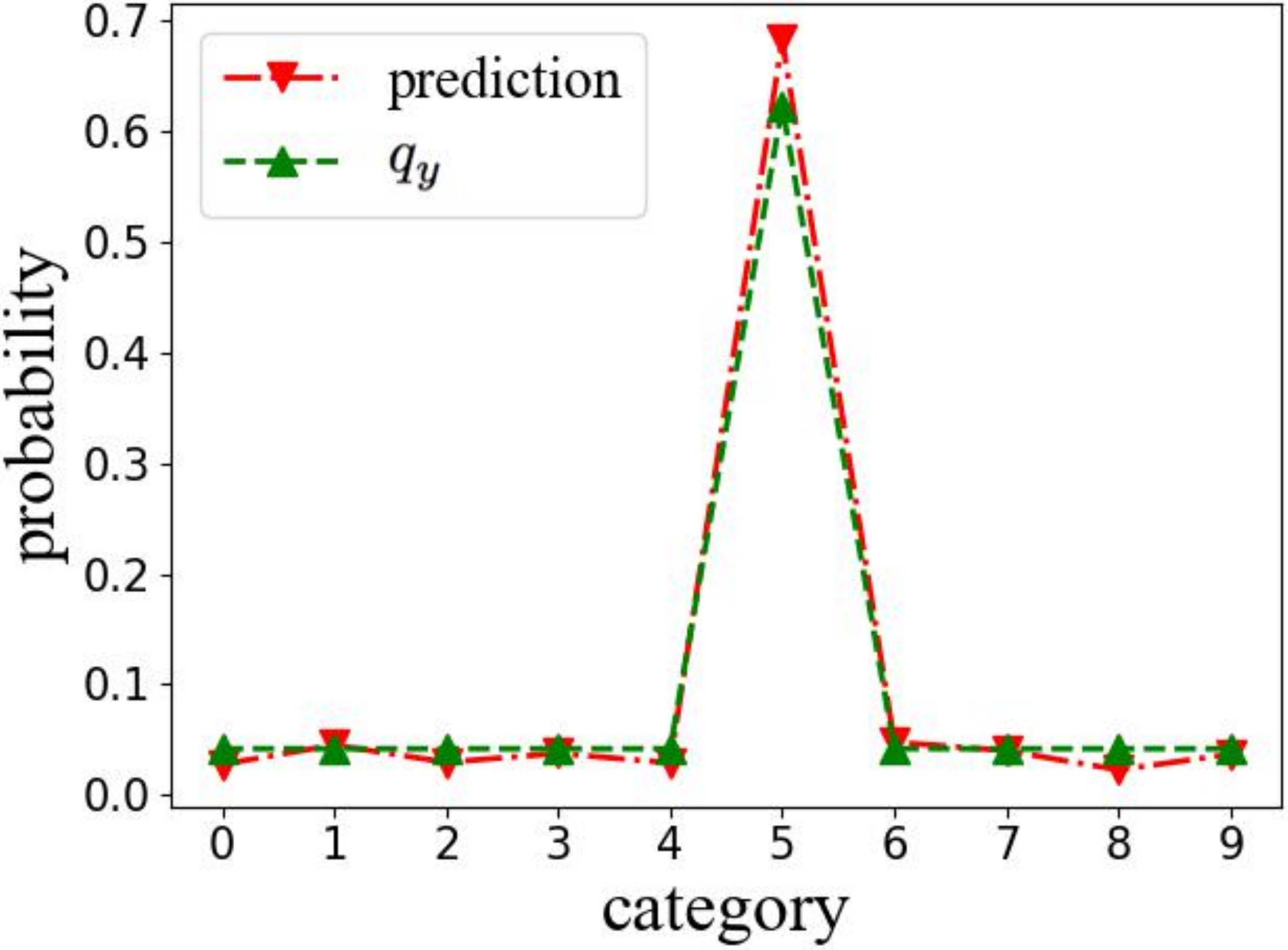}}}
    
    \subfigure[average for BN]{
    \centering{\includegraphics[width=0.42\columnwidth]{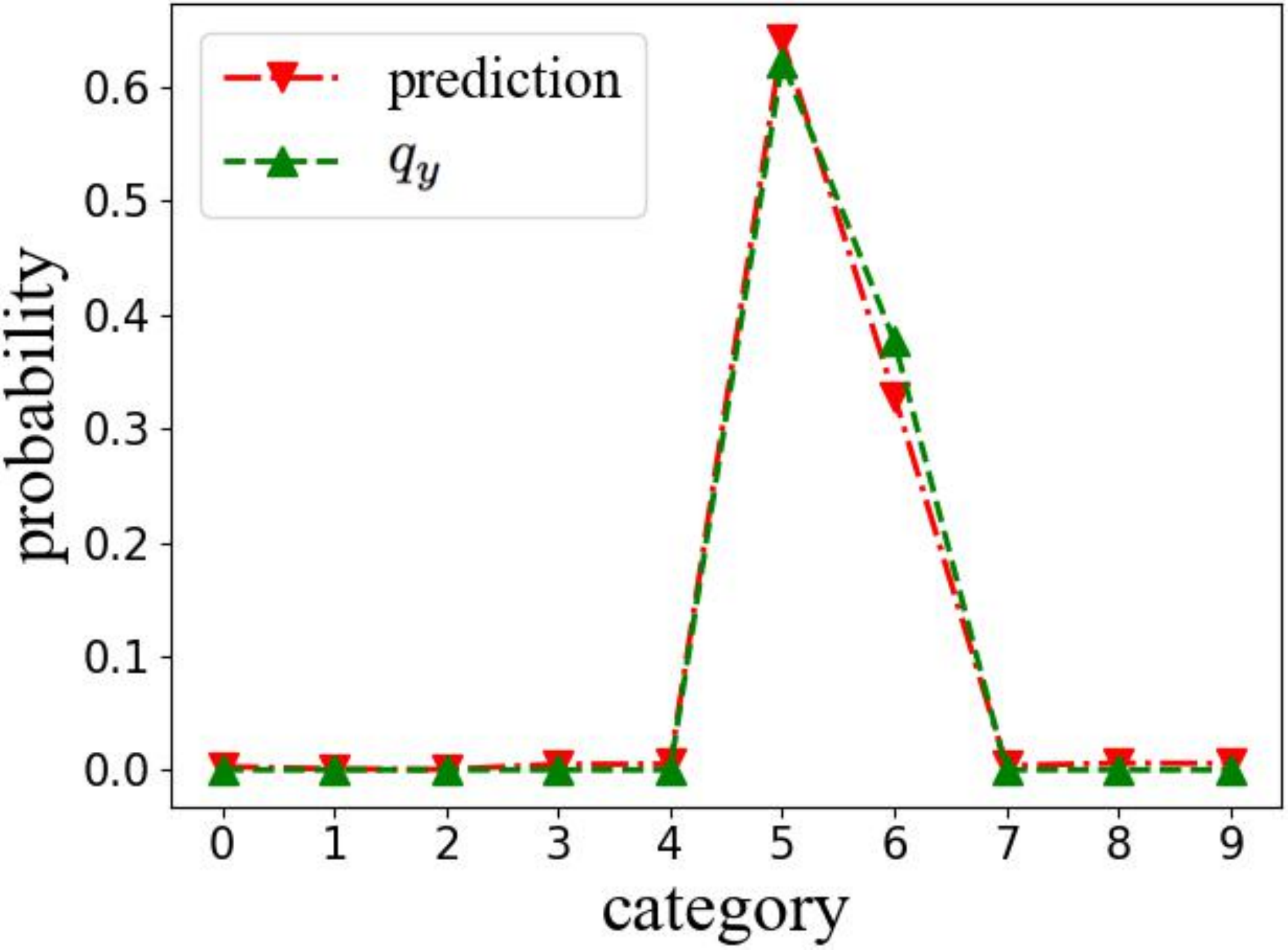}}}
    \subfigure[an instance for BN]{
    \centering{\includegraphics[width=0.42\columnwidth]{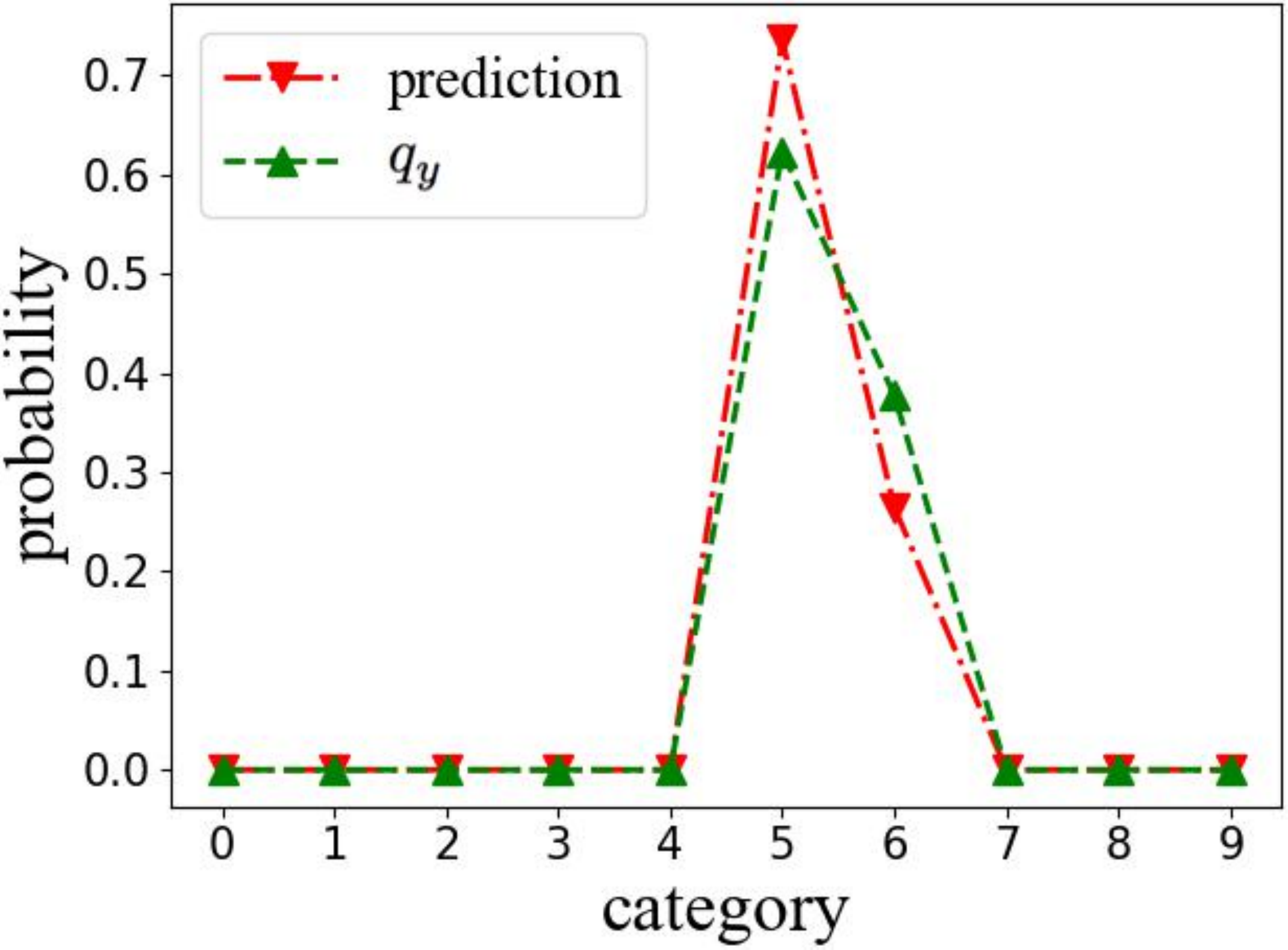}}}
    
    \subfigure[average for GN]{
    \centering{\includegraphics[width=0.42\columnwidth]{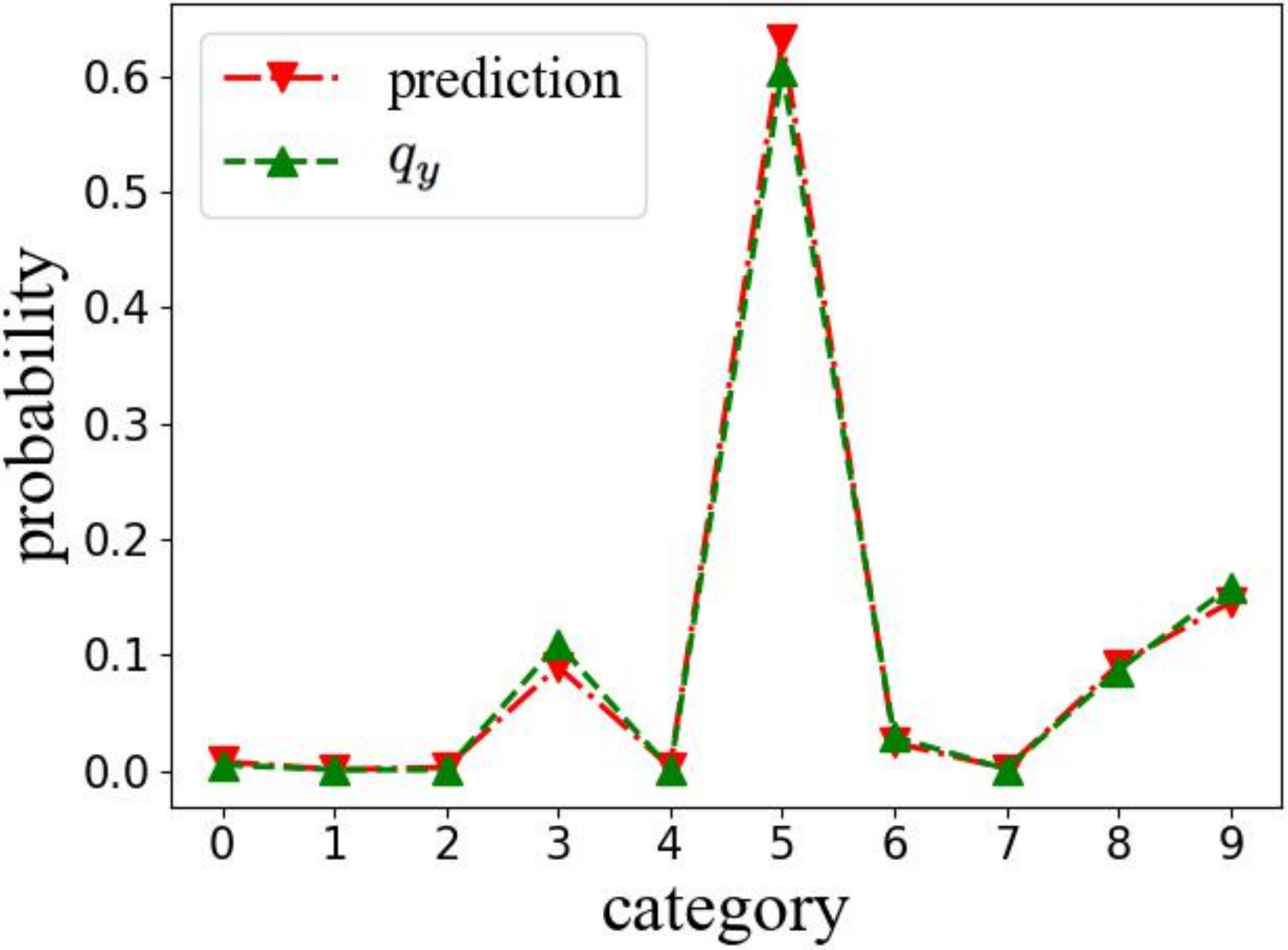}}}
    \subfigure[an instance for GN]{
    \centering{\includegraphics[width=0.42\columnwidth]{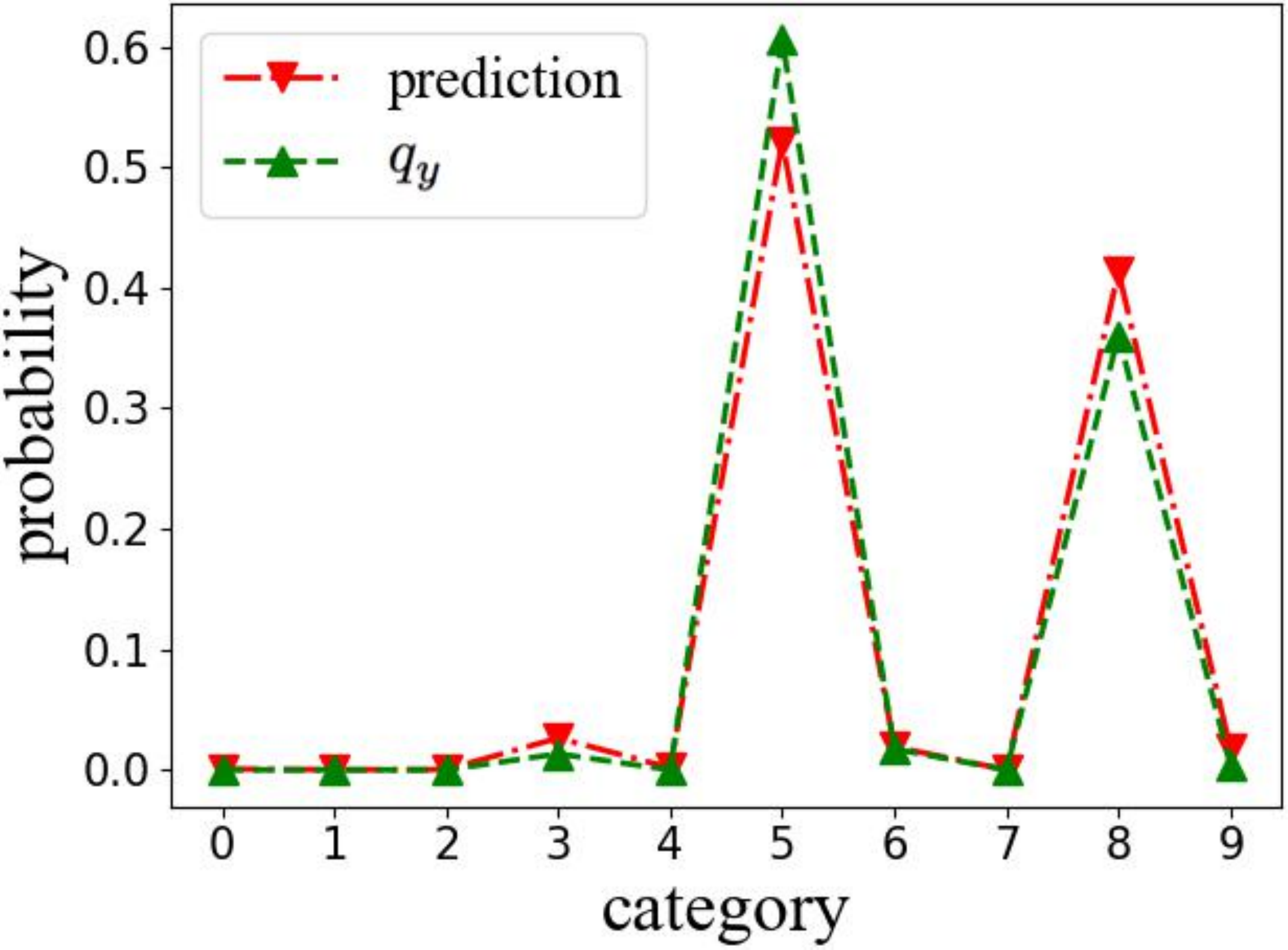}}}

\caption{The predicted probabilities by the trained models compared to $q_y$ for digit $5$ at $\beta=0.6$. The left column is the average for all images of digit $5$. The right column plots a single instance. The red lines represent $f(y; \theta)$ whereas the green lines represent $\mathbf{q}_{y}$. UN: uniform noise; BN: biased noise; GN: generated noise.}
\label{fig:predict_probability}
\end{center}
\end{figure}

\cref{fig:avg_ce} shows that $\overline{\mathrm{CE}}_f$ and $\overline{\mathrm{CE}}_q$ follows the same trend and have small margins at different noise levels for all three types of noises due to the limit of dataset size and imperfect training. It verifies that the predictive probabilities of the trained model is close to the theoretical results $q_y$. In \cref{fig:predict_probability}, we further show the predictive probabilities and $q_y$ for digit $5$ and $\beta = 0.6$. The averages of all images of digit $5$ are listed in the left column and three instances are listed in the right column. It can be seen that the model predictions (red lines) and the calculated $q_y$ (green lines) are in good accordance with each other for both the averaged and individual curves, which further verified our theoretical result in \cref{eq:minimul_ce_loss}. More results are given in \cref{sec:more_results_mnist} \cref{fig:uniform_noise}, \cref{fig:biased_noise} and \cref{fig:generated_noise}.

\begin{figure}[t]
\begin{center}
    \centerline{\includegraphics[width=0.75\columnwidth]{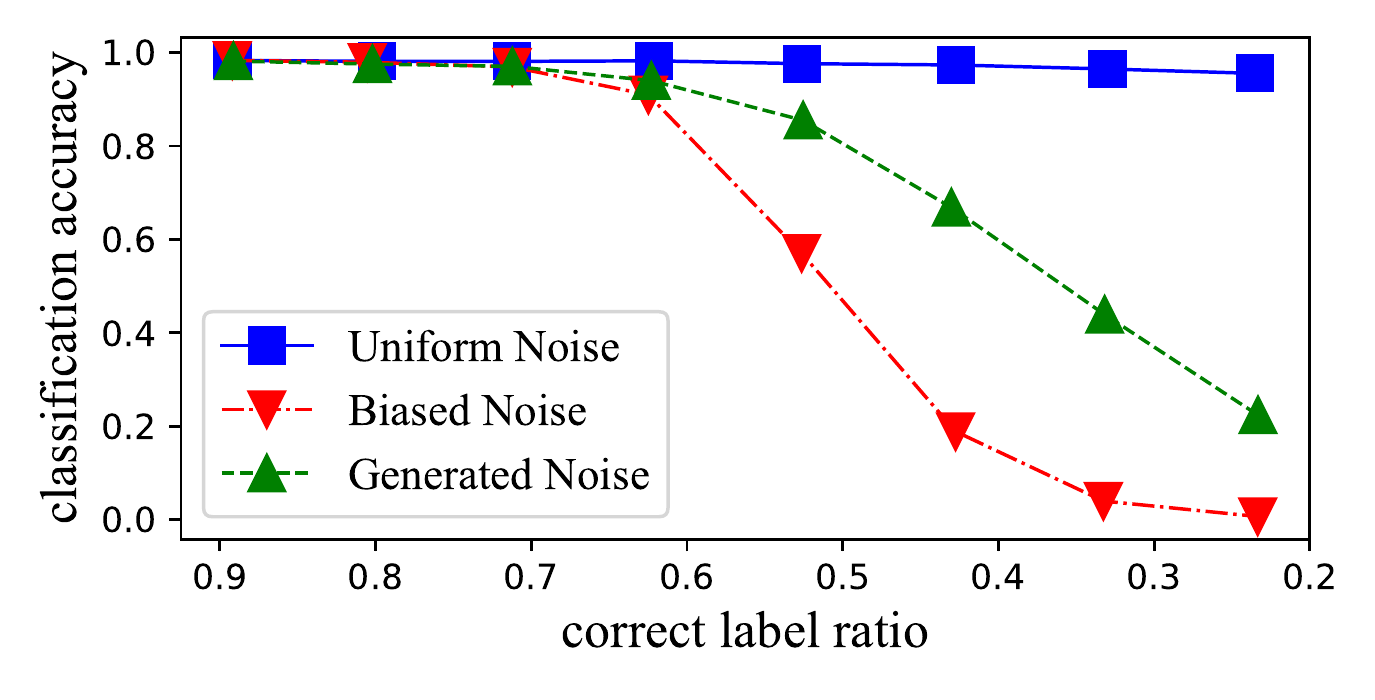}}
    \caption{The accuracies of classification on test data for three types of label noise and different noise level.}
    \label{fig:mnist_classfication}
\end{center}
\end{figure}

The accuracy of classification at different noise levels is illustrated in \cref{fig:mnist_classfication}, which demonstrates the different behavior of the trained models under different noise distributions. Uniform noise has little influence on the accuracy. It is because that we were taking the maximum element of $q_y$ as the prediction, but in uniform noise, it needs $\alpha < 0.1$ for the probability of the incorrect classes to surpass that of the correct class in $q_y$. For the biased noise type, it only requires that $\alpha < 0.5$ for that to happen, and that is why the accuracy with biased noise drops drastically at the correct label rate of $0.5$. For generated noise, it is easier than uniform noise for the magnitude of the wrong classes in $q_y$ to surpass the correct ones, but it is harder than the biased noise. Hence the generated noise has an accuracy curve in between. 

\subsubsection{Discussion}
The experiment provides evidence and explanation to deep learning's resilience to label noises, which is unavoidable in practical applications.As demonstrated in \cref{fig:mnist_classfication}, the trained models had very decent performance at a labeling error rate of $30\%$ for all three noise levels. Because the model has different resilience to noise levels under different noise types, it is important to understand the type of labeling noises and design appropriate quality assurance procedure accordingly. For example, if the labels are derived from automatically collected data it may suffer from uniform noise due to measurement errors, and the model can tolerate higher label noise level. However, for manually annotated data the label noise may be closer to generated noise (recognition error) or biased noise (misoperation during labeling), where we need to pay more attention to the correctness of the label because the model is less resistant to labeling noises. 

\subsection{Pixel-wise Variance Estimation of Single Image Super-resolution}
\label{sec:imagenet_experiment}

\subsubsection{Experiment Setup}

We conducted an experiment on the ImageNet dataset \citep{deng2009imagenet}, which contains 1.28 million images of 1k categories. We randomly chose 2 images from each category in the original training set for testing, whereas the remaining images were used for training. The task was $4 \times$ super-resolution from $64\times 64$ to $256 \times 256$. As reference, we employed an SR model based on DDPM to generate multiple high-resolution images given a low-resolution input \citep{dhariwal2021diffusion, ho2020denoising}. Some generated samples are given in \cref{sec:more_results_sr} \cref{fig:super-resolution_samples_ddpm}. The mean and pixel-wise variance of $x \mid y$ were estimated by the Monte Carlo method using $100$ samples. For convenience, we call this model DDPM-SR.

For our proposed direct estimation method, the network architecture of $f_{\mathrm{mean}}$ and $f_{\mathrm{var}}$ was modified from DDPM-SR by removing the embedding part of time $t$ and categorical information. Because the network architecture is similar to a U-Net \citep{ronneberger2015u}, we upsampled $y$ to $256\times 256$ resolution as the input of networks. A smaller network compared to DDPM-SR was also adopted to reduce the training time.

We first trained $f_\mathrm{mean}(\cdot;\theta_1)$ by \cref{eq:sr_mean_train}, followed by training $f_\mathrm{var}(\cdot;\theta_2)$ by \cref{eq:sr_var_train}. During the training of the variance network, we concatenated $f_\mathrm{mean}(y;\theta_1)$ with $y$ along the channel dimension and fed it as the input to $f_\mathrm{var}$. We trained $f_\mathrm{mean}(\cdot;\theta_1)$ for 90k steps and $f_\mathrm{var}(\cdot;\theta_2)$ for 50k steps with a batch size of 64. The AdamW optimizer with learning rate of 0.0001 was used for both $f_{\mathrm{mean}}$ and $f_{\mathrm{var}}$. The models at the end of training iterations were selected as the final models.

PSNR, MSE, and NMSE were used to evaluate the models' performance. We used the squared root of the pixel-wise variance, i.e. pixel-wise standard deviation to compute these metrics.

\subsubsection{Experiment Result}

The quantitative comparison among $x$, DDPM-SR and the proposed direct estimation are given in \cref{tab:metrics}. For the mean estimation, $f_\mathrm{mean}$ is much closer to the mean estimated from DDPM-SR compared to the high-resolution image $x$, which verifies that $f_\mathrm{mean}(\cdot;\theta_1)$ trained by \cref{eq:sr_mean_train} fits to the conditional expectation $\mathbb{E}_{x \mid y }[x]$ instead of the target $x$. This also explains the smoothness that people observed when training SR models with L2 norm loss. The directly estimated variance $f_\mathrm{var}$ is also very close to the variance calculated from DDPM-SR, with a very small NMSE of 0.1145. 

\begin{figure}[t]
\begin{center}
    \subfigure[ground truth]{
    \centering{\includegraphics[width=0.31\columnwidth]{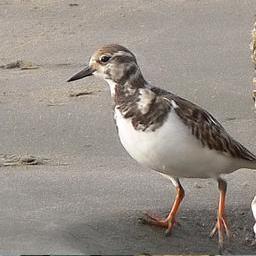}}}
    \subfigure[DDPM-SR mean]{
    \centering{\includegraphics[width=0.31\columnwidth]{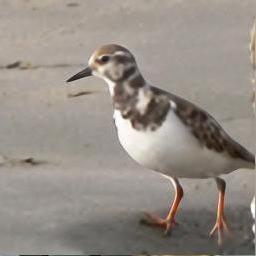}}}
    \subfigure[DDPM-SR var]{
    \centering{\includegraphics[width=0.31\columnwidth]{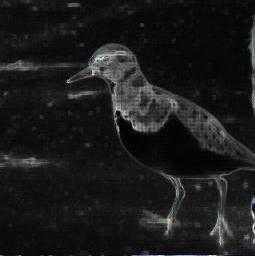}}}
    
    \subfigure[label of $f_{\text{var}}$]{
    \centering{\includegraphics[width=0.31\columnwidth]{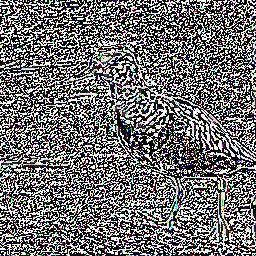}}}
    \subfigure[$f_{\text{mean}}$ mean]{
    \centering{\includegraphics[width=0.31\columnwidth]{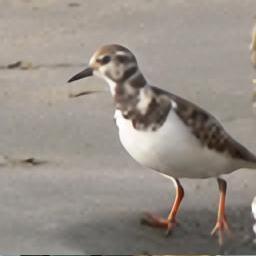}}}
    \subfigure[$f_{\text{var}}$ var]{
    \centering{\includegraphics[width=0.31\columnwidth]{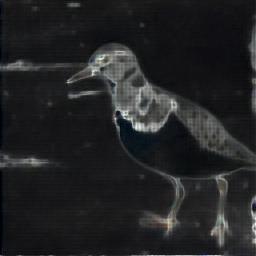}}}
    
\caption{Images: (a) the high-resolution image $x$; (b) estimated $\mathbb{E}_{x\mid y}\left[x\right]$ by DDPM-SR; (c) estimated pixel-wise standard deviation by DDPM-SR; (d) square root of label when training $f_{\mathrm{var}}$, i.e. $\left| x - \mathbb{E}_{x\mid y}\left[x\right] \right|$; (e) estimated $\mathbb{E}_{x\mid y}\left[x\right]$ by $f_{\mathrm{mean}}$; (f) estimated pixel-wise standard deviation by $f_{\mathrm{var}}$. Images in the bottom row are rescaled for better display.}
\label{fig:super-resolution}
\end{center}
\end{figure}

An instance is given in \cref{fig:super-resolution}. As expected, $f_{\mathrm{mean}}(y;\theta_1)$ is smoother than $x$ since it is predicting the conditional expectation instead of $x$. The pixel-wise variance from $f_\mathrm{var}(y;\theta_2)$ and DDPM-SR are visually very close to each other. The training label, $\left| x - \mathbb{E}_{x\mid y}\left[x\right] \right|$, is also given in the figure and demonstrates huge difference from the estimated variance, which indicates that the proposed model would fit to the uncertainty instead of single data points. More examples are given in \cref{sec:more_results_sr} \cref{fig:super-resolution_samples_1} and \cref{fig:super-resolution_samples_2}.

\begin{table}[t]
\caption{Evaluation metrics. The latter one of comparison objects is regarded as reference when computing. DDPM-SR stands for the mean or variance calculated from 100 samples depending on the comparing results. }
\label{tab:metrics}
\begin{center}
\begin{small}
\begin{sc}
\begin{tabular}{l rrr}
\toprule
        Metrics & PSNR ($\uparrow$) & MSE ($\downarrow$) & NMSE ($\downarrow$) \\
        \midrule
        DDPM-SR vs $x$ & 27.6426 & 37.3984 & 2.5395 \\
        $f_{\text{mean}}$ vs $x$ & 27.3060 & 38.3413 & 2.2611 \\
        $f_{\text{mean}}$ vs DDPM-SR & 37.4526 & 9.7725 & 1.0446 \\
        $f_{\text{var}}$ vs DDPM-SR & 23.9540 & 15.0560 & 0.1145 \\
\bottomrule
\end{tabular}
\end{sc}
\end{small}
\end{center}
\end{table}

\subsubsection{Discussion}
\label{sec:imagenet_discussion}

Experimental results demonstrated the feasibility of using our method to estimate the uncertainty in single image super-resolution. Compared to the generative model, the proposed model trades flexibility for speed during inference. In practice, we may only need one sample of the super-resolved images accompanied by the uncertainty (pixel-wise variance) map, and the proposed method will avoid the high computational cost required by the Monte Carlo sampling.

The proposed method can be easily modified to compute the expected error of a trained deterministic SR model against the ground truth $x$, where one can replace the $f_\mathrm{mean}$ in \cref{eq:sr_var_train} with the trained target model. Beyond variance, it is also possible to train other statistical values such as skewness as listed in \cref{sec:more_results_sr} \cref{tab:examples_of_statistics}. In addition, this method can be also applied to other inverse problem, including denoising, deblurring, and image reconstruction.

\section{Related Works}
\label{sec:related_works}

Theoretical results and experiments in this paper are related to many previous works in different areas, including unsupervised denoising \citep{lehtinen2018noise2noise, kim2021noise2score}, score function estimation \citep{hyvarinen2005estimation, vincent2011connection, song2020score}, classification with noisy labels \citep{arpit2017closer, rolnick2017deep, chen2019understanding}, uncertainty estimation of inverse problem \citep{adler2018deep}.

In \cref{theorem:left2right}, when $g(x, y) = x$ and $L$ is L2 norm loss, \cref{eq:left_training} is the least squares regression problem whose optimal solution, $\mathbb{E}_{x \mid y} [x]$, is well-known \citep{murphy2012machine}. It has been used to explain why the trained model by L2 norm loss tends to predict a smooth result in many low-level vision tasks. \cite{adler2018deep} expanded the result with more general $g(x, y)$ under L2 norm loss, which is the stronger conclusion of \cref{theorem:left2right}. The relationship between supervised learning and unsupervised learning for image denoising task was also discussed in Noise2Score \citep{kim2021noise2score}. 

Noise2Noise \citep{lehtinen2018noise2noise} was proposed to train a denoising model when clean images are inaccessible, which is an application of \cref{theorem:right2left}. Score function is an important concept in statistics \citep{robbins2020empirical, efron2011tweedie} and its estimation is used in generative models recently \citep{hyvarinen2005estimation, song2020score}. \cite{vincent2011connection} proved the equivalence of \cref{eq:score_function_estimation} and many works follows the training strategy \citep{song2020score}. \cref{theorem:right2left} provided an alternative proof to \cref{eq:score_function_estimation}.

Classification with noisy labels has been studied in many works \citep{arpit2017closer, rolnick2017deep, chen2019understanding}. \cite{chen2019understanding} claimed that when the classification model is trained by labels with class-dependent noise, the predictive probability follows the same distribution as the label noise, which is a special case of our analysis in \cref{sec:classification_experiment_design}. Our experiment in \cref{sec:mnist_experiment} is similar to the ones conducted in \citep{rolnick2017deep} and our experiment results are consistent to theirs. \cite{adler2018deep} also proposed a method to estimate the pixel-wise variance in computed tomography (CT) reconstruction.

\section{Conclusion}
\label{sec:conclusion}

In this work, we first proved that a trained model in supervised deep learning minimizes the conditional risk for each input. The then illustrated the equivalence between supervised learning problems with intractable labels and its computationally feasible substitution. In addition, we explained many different existing works, such as Noise2Score, Noise2Noise and score function estimation through our theorems. In addition, by applying our theorems we showed that the predictive probability of image classification models trained with noisy labels is related to the noise distribution theoretically and experimentally. Furthermore, we proposed and validated a method to accurately estimate uncertainty of single image super-resolution based on our theorems.

\newpage
\appendix
\section{Proofs}
\label{sec:appendix_proofs}

\subsection{Proof of \cref{theorem:left2right}}
\label{sec:proof_left2right}

\begin{proof}
For any $\theta$ and function $f(\cdot; \theta)$, we have that:
\begin{align*}
    \mathbb{E}_{x, y} \left[L\left( f(y; \theta), g(x, y) \right)\right]
    = & \mathbb{E}_{y} \left[ \mathbb{E}_{x \mid y} \left[ L \left(f(y; \theta), g(x, y) \right) \right] \right] \notag \\
    \geq & \mathbb{E}_{y} \left[ \min_{z} \mathbb{E}_{x \mid y} \left[ L\left(z, g(x, y)\right)\right]\right] \notag \\
    = & \mathbb{E}_{y} \left[ \mathbb{E}_{x \mid y} \left[ L \left(\arg \min_{z} \mathbb{E}_{x \mid y} \left[ L\left(z, g(x, y)\right)\right] , g(x, y) \right) \right] \right] \notag \\
    = & \mathbb{E}_{x, y} \left[L\left( \arg \min_{z} \mathbb{E}_{x \mid y} \left[ L\left(z, g(x, y)\right)\right], g(x, y)  \right)\right].
\end{align*}
Let $z^*(y) = \arg \min_{z} \mathbb{E}_{x \mid y} \left[ L\left(z, g(x, y)\right)\right]$. According to the assumption of $f(\cdot; \theta)$, there exists $\theta^*$ such that $f(y; \theta^*) = z^*(y)$ which is the optimal solution.
\end{proof}

\subsection{Another Version of \cref{theorem:left2right}}
\label{sec:another_version_left2right}
\begin{theorem}
Assume that:
\begin{itemize}
    \item $X$, $Y$ and $W$ are measurable spaces;
    \item $x$ and $y$ are random variables defined in $X$ and $Y$ respectively;
    \item $g(x, y): X \times Y \rightarrow W$ is a measurable function;
    \item $L(a, b): W \times W \rightarrow \mathbb{R}$ is a loss function and satisfies that 
    \begin{equation}
        L(a, b) \geq L(a, a), L(a, b) \geq  L(b, b).
    \end{equation}
    \item $f(y): Y \rightarrow W$ is a measurable function.
\end{itemize}
Consider the following optimization problem:
\begin{equation}
    \min_{f} \mathbb{E}_{x, y} \left[L \left(f\left(y \right), g\left(x, y\right)\right) \right],
\end{equation}
the optimal solution $f^*$ satisfies:
\begin{equation}
    f^*(y) = z^*(y) = \arg\min_{z} \mathbb{E}_{x \mid y} \left[ L \left(z, g \left(x, y \right) \right) \right].
\end{equation}
\end{theorem}

\subsection{The derivation of $z^*(y) = g(x, y)$ when there exists a mapping from $y$ and $x$}
\label{sec:proof_trivial_conclusion}
Suppose $x = h(y)$, we have that
\begin{equation*}
    p(x \mid y) = 
    \begin{cases}
    1, & \text{ if } x=h(y) \\ 
    0 & \text{ otherwise} 
    \end{cases}
\end{equation*}
Then, we derive that
\begin{align*}
    z^*(y) & = \arg \min_{z} \mathbb{E}_{x \mid y} \left[ L\left(z, g(x, y)\right)\right] = \arg \min_{z} \left[ L\left(z, g(x, y)\right)\right] = g(x, y) 
\end{align*}
Therefore, $z^*(y) = g(x, y)$ holds.

\subsection{Proof of $z^*(y) = \mathbb{E}_{x\mid y}[x]$}
\label{sec:proof_denoising}
\begin{proof}
For any $z$, we have that
\begin{align*}
    \mathbb{E}_{x \mid y} \left[\left\| z - x \right\|^{2}_{2} \right]
    = & \mathbb{E}_{x \mid y} \left[\left\| z - \mathbb{E}_{x \mid y}\left[x\right] + \mathbb{E}_{x \mid y}\left[x\right] - x \right\|^{2}_{2} \right] \notag \\
    = & \mathbb{E}_{x \mid y} \left[ \left\| z - \mathbb{E}_{x \mid y} \left[x\right] \right\|_{2}^2 +  2 \left \langle z - \mathbb{E}_{x \mid y}\left[x\right] , \mathbb{E}_{x \mid y}\left[x\right] - x \right \rangle + 
    \left\| \mathbb{E}_{x \mid y}\left[x\right] - x \right\|^{2}_{2} \right] \notag \\
    \geq & \mathbb{E}_{x \mid y} \left[ \left\| \mathbb{E}_{x \mid y} \left[x\right] - x \right\|^{2}_{2} \right] + 2 \mathbb{E}_{x \mid y} \left[ \left \langle z - \mathbb{E}_{x \mid y}\left[x\right] , \mathbb{E}_{x \mid y}\left[x\right] - x \right \rangle \right] \notag \\
    = & \mathbb{E}_{x \mid y} \left[ \left\| \mathbb{E}_{x \mid y} \left[x\right] - x \right\|^{2}_{2} \right].
\end{align*}
The last equality holds because
\begin{align*}
     \mathbb{E}_{x \mid y} \left[ \left \langle z - \mathbb{E}_{x \mid y}\left[x\right] , \mathbb{E}_{x \mid y}\left[x\right] - x \right \rangle \right] 
     = & \left \langle z - \mathbb{E}_{x \mid y}\left[x\right] , \mathbb{E}_{x \mid y}  \left[ \mathbb{E}_{x \mid y}\left[x\right] - x \right] \right \rangle \\
     = & \left \langle z - \mathbb{E}_{x \mid y}\left[x\right] , \mathbb{E}_{x \mid y} \left[x\right] - \mathbb{E}_{x \mid y} \left[x \right] \right \rangle \\
     = & \left \langle z - \mathbb{E}_{x \mid y}\left[x\right] , 0 \right \rangle \\
     = & 0.
\end{align*}
Therefor,
\begin{equation*}
    \arg \min_{z} \mathbb{E}_{x \mid y}[\left\| z - x \right\|_2^2 ] = \mathbb{E}_{x\mid y} \left[ x \right].
\end{equation*}
\end{proof}

\subsection{Proof of \cref{theorem:right2left}}
\label{sec:proof_right2left}
\begin{proof}
According to \cref{theorem:left2right}, the optimal solution of $\min_{\theta} \mathbb{E}_{x, y} \left[ L(f(y; \theta), g(x, y) \right]$, $\theta^*$, satisfies that $f(y; \theta^*) = z^*(y)$. For any $\theta$, we have
\begin{align*}
    \mathbb{E}_{y} \left[L\left(f(y;\theta\right), z^*(y) \right] \geq \mathbb{E}_{y} \left[L\left(z^*(y), z^*(y) \right) \right] = \mathbb{E}_{y} \left[L\left(f(y;\theta^*), z^*(y) \right) \right].
\end{align*}
That is to say, $\theta^*$ is also the the optimal solution of $\min_{\theta} \mathbb{E}_{y} \left[ L(f(y; \theta), z^*(y) \right]$. Therefore, \cref{eq:right2left} holds.

Similar to the proof in \cref{sec:proof_denoising}, we can prove that
\begin{equation*}
    \arg \min_{z} \mathbb{E}_{x \mid y}[\left\| z - g(x, y) \right\|_2^2 ] = \mathbb{E}_{x\mid y} \left[ g(x, y) \right].
\end{equation*}
Then, we can derive that:
\begin{align*}
    & \mathbb{E}_{y} \left[ \left\|f(y; \theta) - z^*(y) \right\|_2^2 \right] - \mathbb{E}_{x, y} \left[\left\| f(y; \theta) - g(x, y) \right\|_2^2 \right] \notag \\
    = & \mathbb{E}_{y} \left[ \left\|f(y; \theta)\right\|_2^2 \right]  - 2 \mathbb{E}_{y}\left[  \left\langle f(y; \theta), \mathbb{E}_{x \mid y}\left[ g(x, y)\right]  \right\rangle  \right] + \mathbb{E}_{y} \left[ \left\|\mathbb{E}_{x \mid y}\left[ g(x, y)\right] \right\|_2^2 \right] - \mathbb{E}_{x, y} \left[ \left\|f(y; \theta)\right\|_2^2 \right]\notag \\
    & \quad + 2 \mathbb{E}_{x, y} \left[ \left\langle f(y; \theta),  g(x, y) \right\rangle  \right] - \mathbb{E}_{x, y} \left[ \left\| g(x, y) \right\|_2^2 \right] \notag \\
    = & - 2 \mathbb{E}_{y}\left[  \left\langle f(y; \theta), \mathbb{E}_{x \mid y}\left[ g(x, y)\right]  \right\rangle  \right] + C_1 + 2 \mathbb{E}_{x, y} \left[ \left\langle f(y; \theta),  g(x, y) \right\rangle  \right] - C_2,
\end{align*}
where $\mathbb{E}_{y} \left[ \left\|\mathbb{E}_{x \mid y}\left[ g(x, y)\right] \right\|_2^2 \right]$ and $\mathbb{E}_{x, y} \left[ \left\| g(x, y) \right\|_2^2 \right]$ are constants denoted as $C_1$ and $C_2$ respectively.

Because 
\begin{align*}
     \mathbb{E}_{x, y} \left[ \left\langle f(y; \theta),  g(x, y) \right\rangle  \right] = & \int \int p(x, y)\left\langle f(y; \theta),  g(x, y) \right\rangle \mathrm{d}x \mathrm{d} y \notag \\
     = & \int \int  p(y)p(x\mid y)\left\langle f(y; \theta),  g(x, y) \right\rangle \mathrm{d}x \mathrm{d} y \notag \\
     = & \int \int p(y) \left\langle f(y; \theta), p(x\mid y) g(x, y) \right\rangle \mathrm{d}x \mathrm{d} y \notag \\
     = & \int p(y) \left( \int  \left\langle f(y; \theta), p(x|y) g(x, y) \right\rangle \mathrm{d}x \right) \mathrm{d} y \notag \\
     = & \int p(y) \left\langle f(y; \theta), \int  p(x|y) g(x, y) \mathrm{d}x  \right\rangle \mathrm{d} y \notag \\
     = & \int p(y) \left\langle f(y; \theta),  \mathbb{E}_{x \mid y} \left[ g(x, y) \right] \right\rangle \mathrm{d} y \notag \\
     = & \mathbb{E}_{y}\left[  \left\langle f(y; \theta), \mathbb{E}_{x \mid y}\left[ g(x, y)\right]  \right\rangle  \right],
\end{align*}
we have that
\begin{equation*}
\begin{split}
    \mathbb{E}_{y} \left[ \left\|f(y; \theta) - f^*(y) \right\|_2^2 \right] - \mathbb{E}_{x, y} \left[\left\| f(y; \theta) - g(x, y) \right\|_2^2 \right] =  C_1 - C_2.
\end{split}    
\end{equation*}
Let $C = C_2 - C_1$, then the following equation holds:
\begin{equation*}
\begin{split}
    \mathbb{E}_{y} \left[ \left\|f(y; \theta) - z^*(y) \right\|_2^2 \right] + C =  \mathbb{E}_{x, y} \left[\left\| f(y; \theta) - g(x, y) \right\|_2^2 \right],
\end{split}
\end{equation*}
where $C$ is a constant. Therefore, \cref{eq:right2left_l2_norm_case_optimziation_problem} is proved.
\end{proof}

\subsection{Proof of \cref{eq:denoising_equivalent_1}, \cref{eq:denoising_equivalent_2} and \cref{eq:noise2noise}}
\label{sec:proof_noise2noise}
\begin{proof}
Let $g(x, y) = x$, then \cref{eq:denoising_equivalent_1} holds according to \cref{theorem:right2left}. \cref{eq:denoising_equivalent_2} is proved as long as replacing $x$ in \cref{eq:denoising_equivalent_1} by $s$.

Next, we prove \cref{eq:noise2noise}. Suppose $n_2$ is sampled from a random variable $n$, which represents the noise. Because $\mathbb{E}[n] = 0$, then 
\begin{equation*}
    \mathbb{E}_{x \mid s}\left[x\right] = \mathbb{E}_{n} \left[ s + n\right] = s.
\end{equation*}
We can derive that
\begin{align*}
    \mathbb{E}_{x\mid y} \left[ x \right] 
    & = \int p(x \mid y) x \mathrm{d} x \notag \\
    & = \int \left( \int p(x, s \mid y) \mathrm{d} s \right) x \mathrm{d} x \notag \\
    & = \int \left( \int p(x \mid s, y) p(s \mid y) \mathrm{d} s \right) x \mathrm{d} x \notag \\
    & = \int \left( \int p(x \mid s) p(s \mid y) \mathrm{d} s \right) x \mathrm{d} x \notag \\
    & = \int p(s \mid y) \left( \int p(x \mid s) x  \mathrm{d} x \right)  \mathrm{d} s \notag \\
    & = \int p(s \mid y) \mathbb{E}_{x \mid s} \left[x\right] \mathrm{d} s \notag \\
    & = \int p(s \mid y) s \mathrm{d} s \notag \\
    & = \mathbb{E}_{s \mid y} \left[ s \right].
\end{align*}
Thus, \cref{eq:noise2noise} holds.
\end{proof}

\subsection{Proof of \cref{eq:score_function_estimation_key_part}}
\label{sec:proof_score_function}
\begin{proof}
We only need to prove that $ \mathbb{E}_{x \mid y} \left[ \nabla_{y} \log p(y \mid x )\right] = \nabla_{y} \log p(y)$. We have the following derivation:
\begin{align*}
    \mathbb{E}_{x \mid y} \left[ \nabla_{y} \log p(y \mid x )\right] = & \int p(x \mid y) \nabla_{y} \log p(y \mid x ) \mathrm{d} x \notag \\
    = & \int p(x \mid y) \nabla_{y} \log \frac{ p(x \mid y )p(y)}{p(x)} \mathrm{d} x\notag \\
    = & \int p(x \mid y) \nabla_{y} \left[ \log p(x \mid y ) + \log p(y) - \log p(x) \right] \mathrm{d} x\notag \\
    = & \int p(x \mid y) \nabla_{y} \log p(x \mid y )\mathrm{d} x + \int p(x \mid y) \nabla_{y} \log p(y) \mathrm{d} x\notag \\
    = & \int p(x \mid y) \nabla_{y} \log p(x \mid y )\mathrm{d} x + \nabla_{y} \log p(y).
\end{align*}
Next, we prove that $\int p(x \mid y) \nabla_{y} \log p(x \mid y )\mathrm{d} x = 0$.
\begin{align*}
    \int p(x \mid y) \nabla_{y} \log p(x \mid y )\mathrm{d} x = & \int p(x \mid y) \frac{1}{p(x \mid y )} \nabla_{y} p(x \mid y )\mathrm{d} x \notag \\
    = & \int \nabla_{y} p(x \mid y )\mathrm{d} x \notag \\
    = & \nabla_{y} \int p(x \mid y )\mathrm{d} x \notag \\
    = & \nabla_{y} 1 = 0.
\end{align*}
Thus, \cref{eq:score_function_estimation_key_part} is proved.
\end{proof}

\subsection{Proof of Equation \cref{eq:minimul_ce_loss}}
\label{sec:proof_classification_noise_label}
\begin{proof}
When cross-entropy loss $\mathrm{CE}\left({a}, {b} \right)$ for two discrete distributions is computed, two inputs, ${a}$ and ${b}$, must satisfy that $\sum_{i=1}^{n} a_i = \sum_{i=1}^{n} b_i = 1$, and for any $i$, $a_i \geq 0, b_i \geq 0$. Otherwise cross-entropy loss cannot measure the closeness of two discrete distributions. Rigorously, when at least one input does not satisfy the condition, we can define $\mathrm{CE}\left( a, b \right) = +\infty$. In addition, we define $0 \log 0 = 0$ since $\lim_{x\rightarrow 0} x \log x = 0$ accroding to L'Hospital's rule. Therefore, $\arg \min_{{z}} \mathbb{E}_{x\mid y} \left[ \mathrm{CE}({z}, {x})\right]$ can be rewritten as the the following optimization problem:
\begin{equation}
\begin{split}
\label{eq:optimization_ce_loss}
&\arg \min_{{z}} \mathbb{E}_{x\mid y} \left[ \mathrm{CE}({z}, {x} ) \right] \\
\text{s.t} & \sum_{i=1}^{n} z_i = 1, \\
&z_i \geq 0, i = 1, ..., n.
\end{split}
\end{equation}
We denote the $n$-dimension one-hot vector whose components are all $0$ but $1$ at the position $i$ as ${x}^{(i)}$. Suppose ${z}$ satisfies the constraints. Since $p\left({x} = {x}^{(i)} \mid y\right) = q_{y,i}$, we have that
\begin{align}
\label{eq:ce_loss_computation}
    \mathbb{E}_{x \mid y} \left[ \mathrm{CE} \left({z}, {x} \right) \right] & = \sum_{i=1}^n q_{y,i} \sum_{j=1}^{n} - x^{(i)}_{j} \log z_{j} = - \sum_{i=1}^{n} q_{y,i} \log z_{i}.
\end{align}
Bring \cref{eq:ce_loss_computation} to the optimization problem \cref{eq:optimization_ce_loss} and solve it by the Lagrange multiplier method. We derive that
\begin{equation*}
    \arg \min_{{z}} \mathbb{E}_{x\mid y} \left[ \mathrm{CE}({z}, {x} ) \right] = {q}_{y}.
\end{equation*}

\end{proof}

\section{More details of Experiments}

\subsection{Details for Image Classification with Noisy Labels}
\label{sec:details_mnist}

The CNN network architecture and concrete parameters are listed in \cref{tab:mnist_network_parameter}. The digit image is scaled to range of $[0, 1]$ by dividing $255$. The specific values of $\alpha$ and $\beta$ for different noise level are listed in \cref{tab:noise_type}.

\begin{table}[ht]
\caption{Parameters of CNN}
\label{tab:mnist_network_parameter}
\begin{center}
\begin{small}
\begin{sc}
\begin{tabular}{l c}
\toprule
        Layer & Parameters \\
        \midrule
        conv & $6$ channels, stride is $5$ \\
        relu & None \\
        max pooling & None \\
        conv & $16$ channels, stride is $5$\\
        relu & None \\
        max pooling & None \\
        flatten & None \\
        linear & output dimension is $128$ \\
        relu & None \\
        linear & output dimension is $64$ \\
        relu & None \\
        linear & output dimension is $10$ \\
        
\bottomrule
\end{tabular}
\end{sc}
\end{small}
\end{center}
\end{table}

\begin{table}[ht]
\caption{The values of $\alpha$ and $\beta$ at different noise levels and corresponding correct label ratio for three types noise labels.}
\label{tab:noise_type}
\begin{center}
\begin{small}
\begin{sc}
\begin{tabular}{l r r r}
\toprule
        Noise Type & Uniform & Bias & Generated \\
        \midrule
        $\alpha=0.8913, \beta=0.9$ & $0.8919$ & $0.8920$ & $0.8913$ \\
        $\alpha=0.7121, \beta=0.7$ & $0.7122$ & $0.7122$ & $0.7121$ \\ 
        $\alpha=0.5254, \beta=0.5$ & $0.5262$ & $0.5264$ & $0.5254$ \\ 
        $\alpha=0.4302, \beta=0.4$ & $0.4274$ & $0.4275$ & $0.4302$ \\ 
        $\alpha=0.3318, \beta=0.3$ & $0.3300$ & $0.3326$ & $0.3318$ \\ 
\bottomrule
\end{tabular}
\end{sc}
\end{small}
\end{center}
\end{table}

\subsection{Details for Pixel-wise Variance Estimation of Single Image Super-resolution}
\label{sec:details_sr}

\cref{tab:examples_of_statistics} lists some examples of statistics that can be represented by $\mathbb{E}_{x \mid y} \left[ g(x, y) \right]$.

Our code of this experiment was based on Dhariwal et al.'s code\footnote{https://github.com/openai/guided-diffusion} \citep{dhariwal2021diffusion}. We use the default parameters provided by their code except that we only used $4$ times down-sampling. We also utilized mixed precision to accelerate computing. To avoid $f_{\text{var}}$ outputs values less than $0$, we replaced those values by $0$.

\begin{table}[t]
\caption{Examples of statistics represented by $\mathbb{E}_{x \mid y} \left[ g(x, y) \right]$.}
\label{tab:examples_of_statistics}
\begin{center}
\begin{small}
\begin{sc}
\begin{tabular}{lc}
\toprule
        Statistics & $g(x, y)$ \\
        \midrule
        Mean & $x$ \\
        Covariance matrix & $\left(x - \mathbb{E}_{x \mid y}\left[ x \right]\right) \left(x - \mathbb{E}_{x \mid y}\left[ x \right]\right)^{T}$ \\
        Pixel-wise variance & $\left( x - \mathbb{E}_{x \mid y}\left[ x \right] \right)^2$ \\
        Pixel-wise skewness & $\frac{\left( x - \mathbb{E}_{x \mid y}\left[ x \right] \right)^{3}}{\mathbb{E}_{x \mid y}\left[ \left( x - \mathbb{E}_{x \mid y}\left[ x \right] \right)^2 \right]^{3/2}}$ \\
\bottomrule
\end{tabular}
\end{sc}
\end{small}
\end{center}
\end{table}

\section{More Results of Experiments}

\subsection{Image Classification with Noisy Labels}
\label{sec:more_results_mnist}

\begin{figure}[t]
\begin{center}
    \subfigure[$\beta=0.4$, digit of $0$]{
    \centering{\includegraphics[width=0.22\columnwidth]{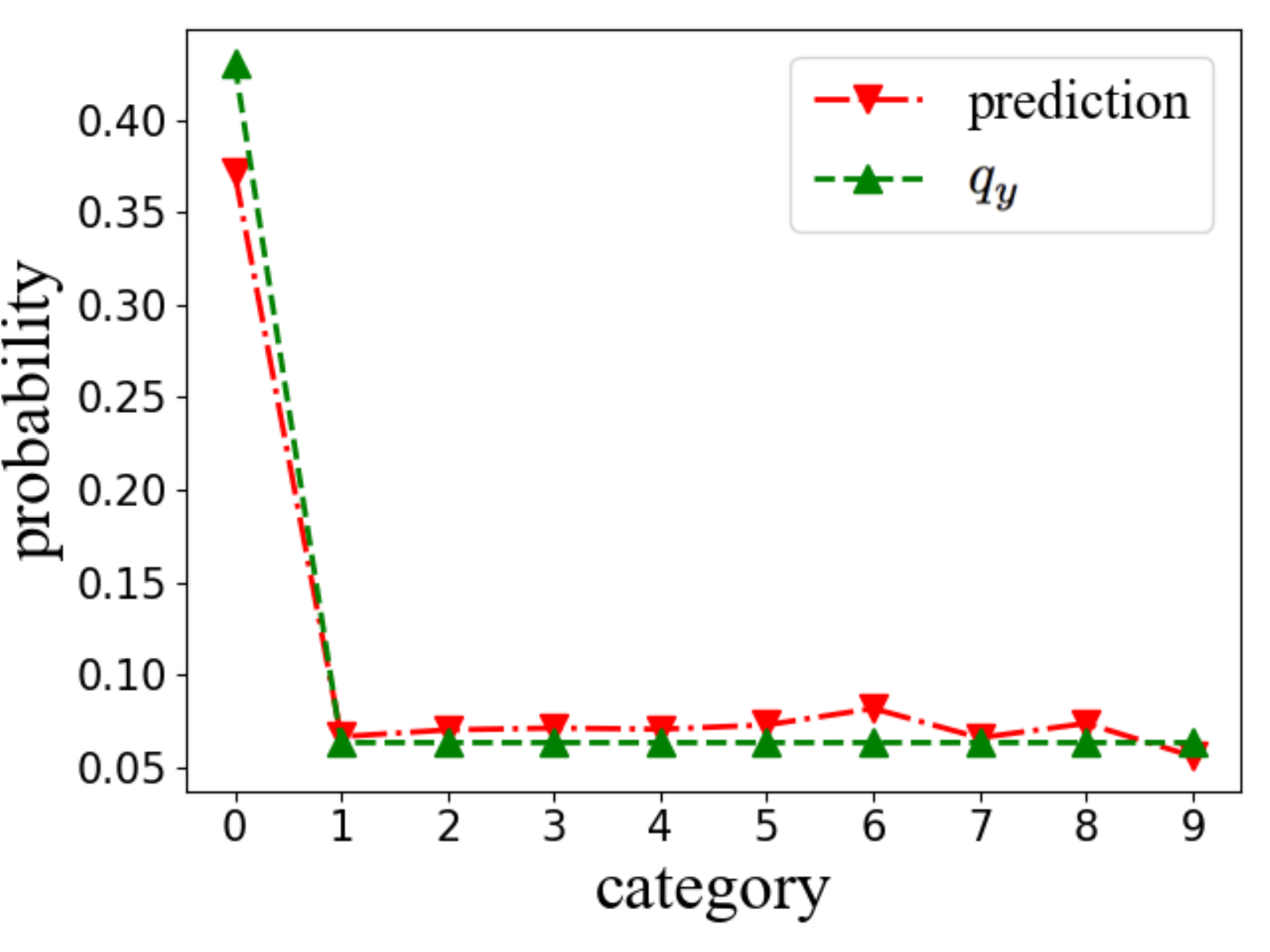}}}
    \subfigure[$\beta=0.4$, digit of $1$]{
    \centering{\includegraphics[width=0.22\columnwidth]{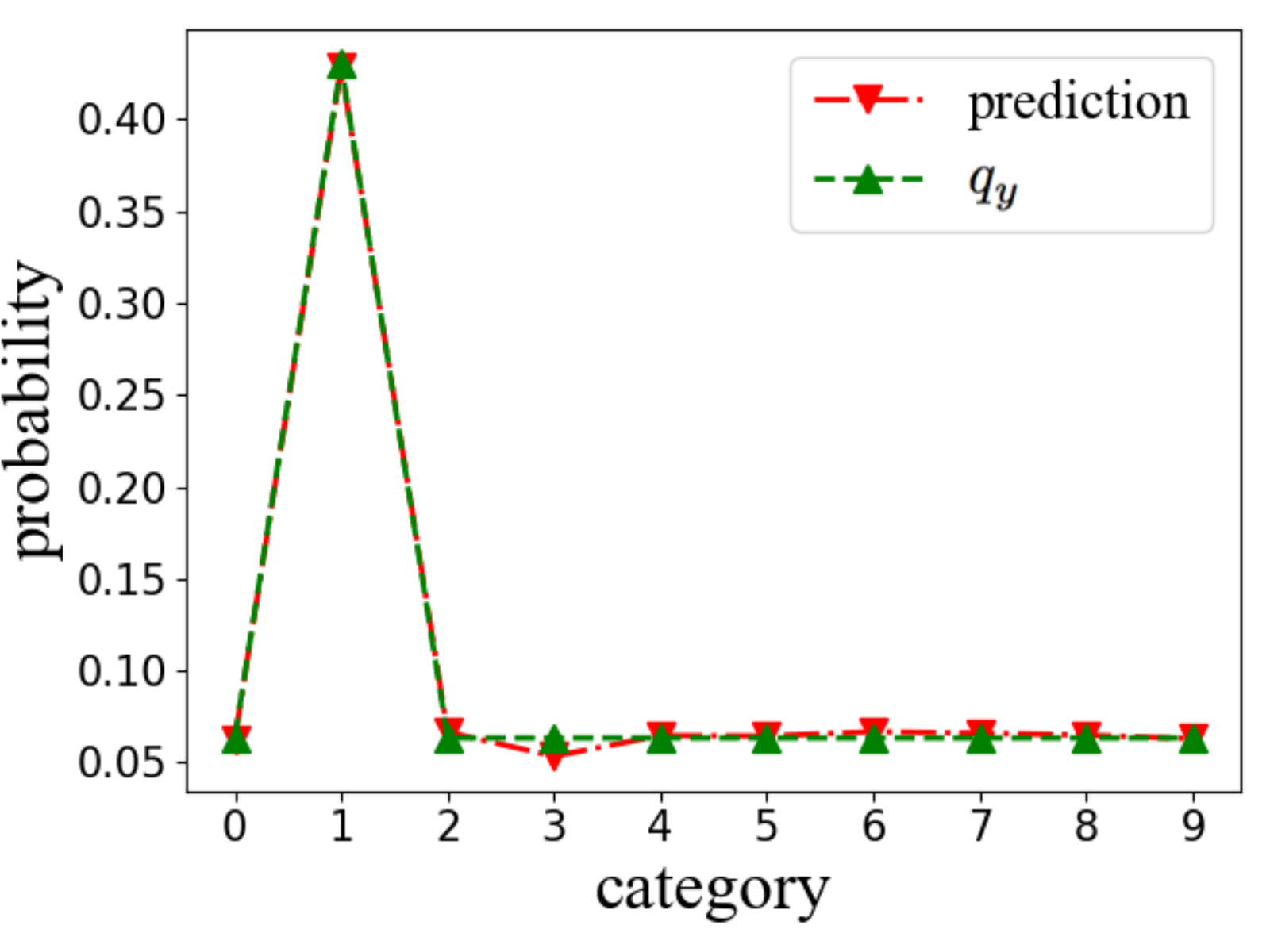}}}
    \subfigure[$\beta=0.4$, digit of $2$]{
    \centering{\includegraphics[width=0.22\columnwidth]{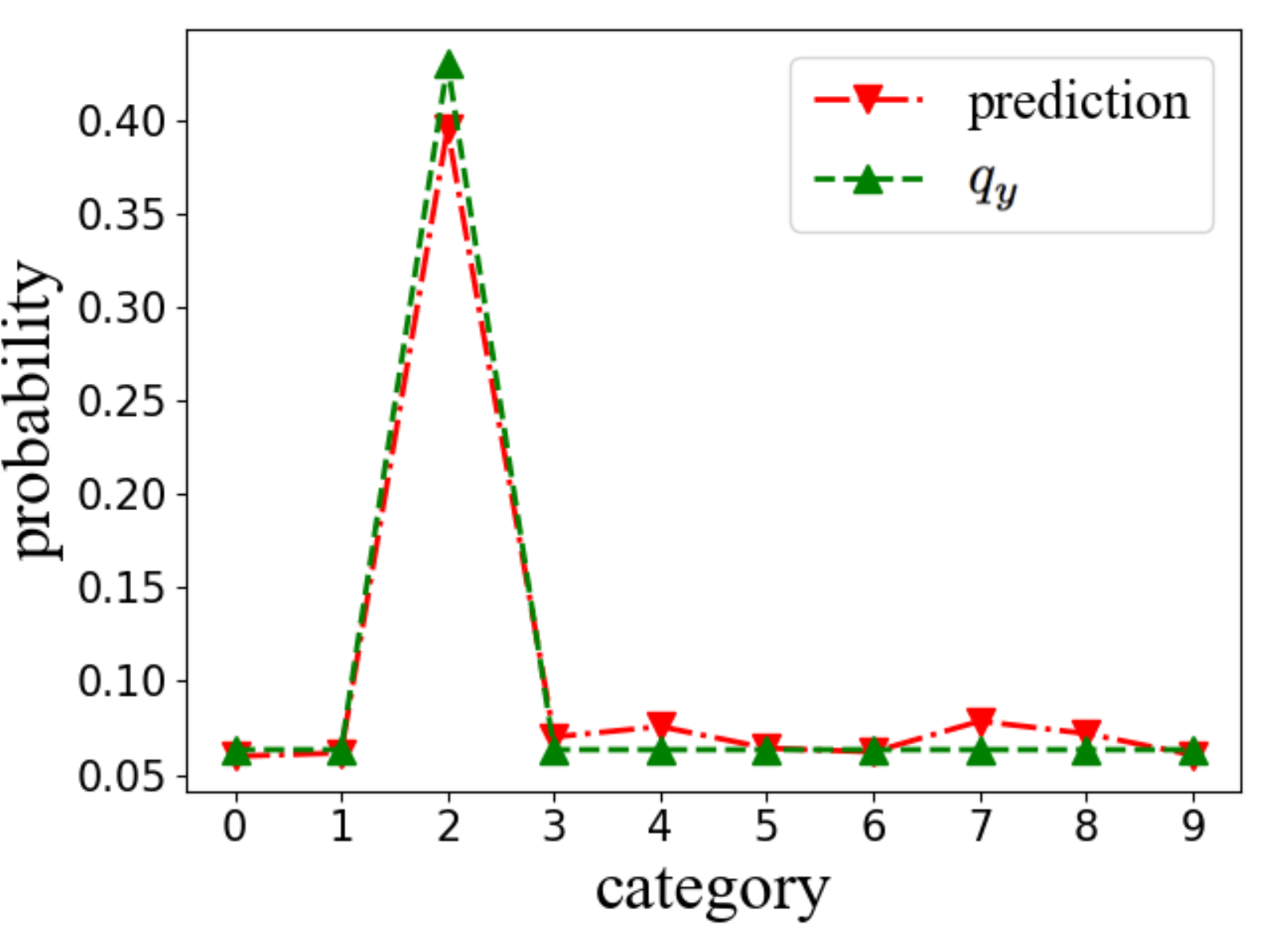}}}
    \subfigure[$\beta=0.4$, digit of $3$]{
    \centering{\includegraphics[width=0.22\columnwidth]{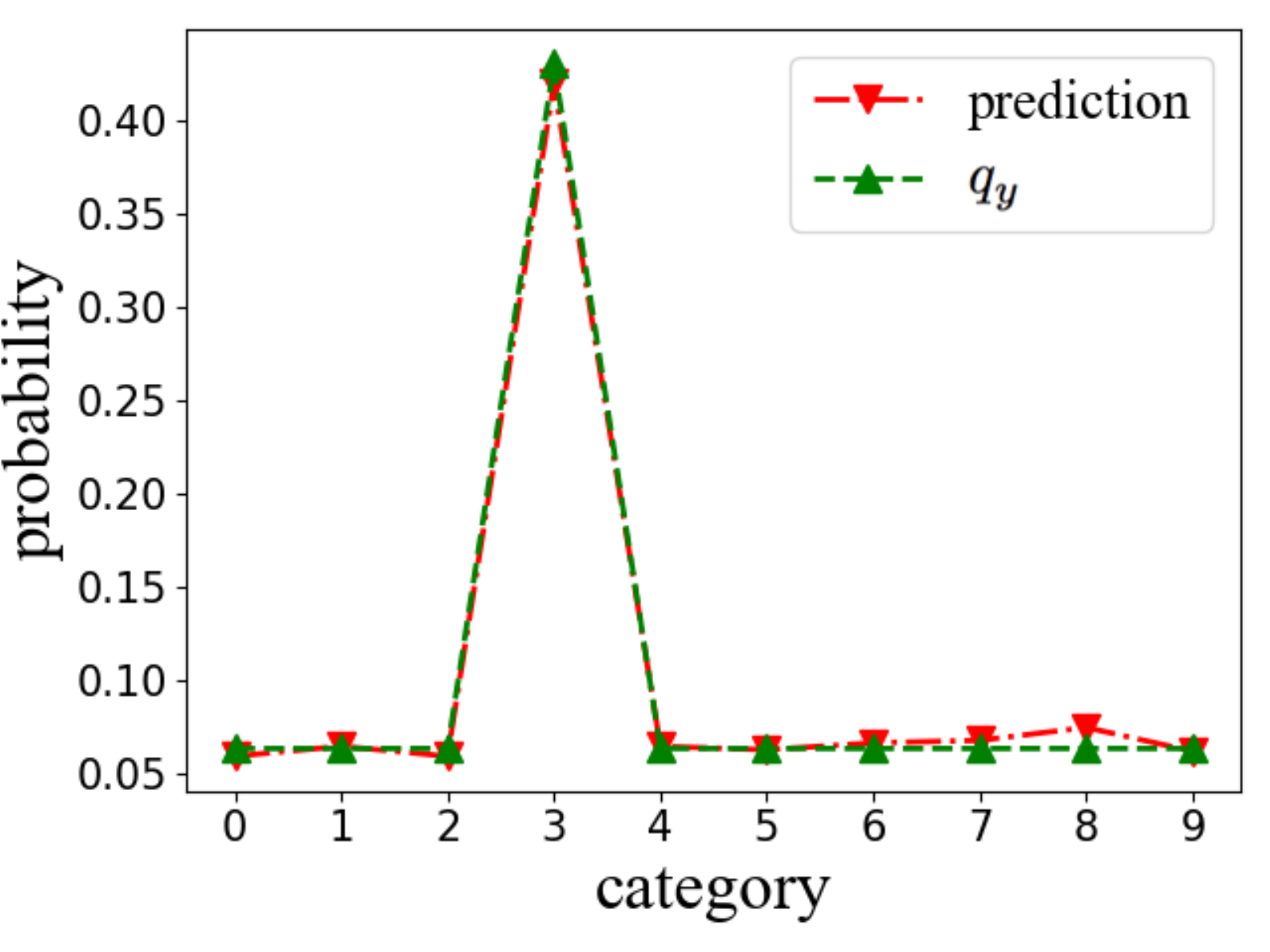}}}
    
    \subfigure[$\beta=0.4$, digit of $4$]{
    \centering{\includegraphics[width=0.22\columnwidth]{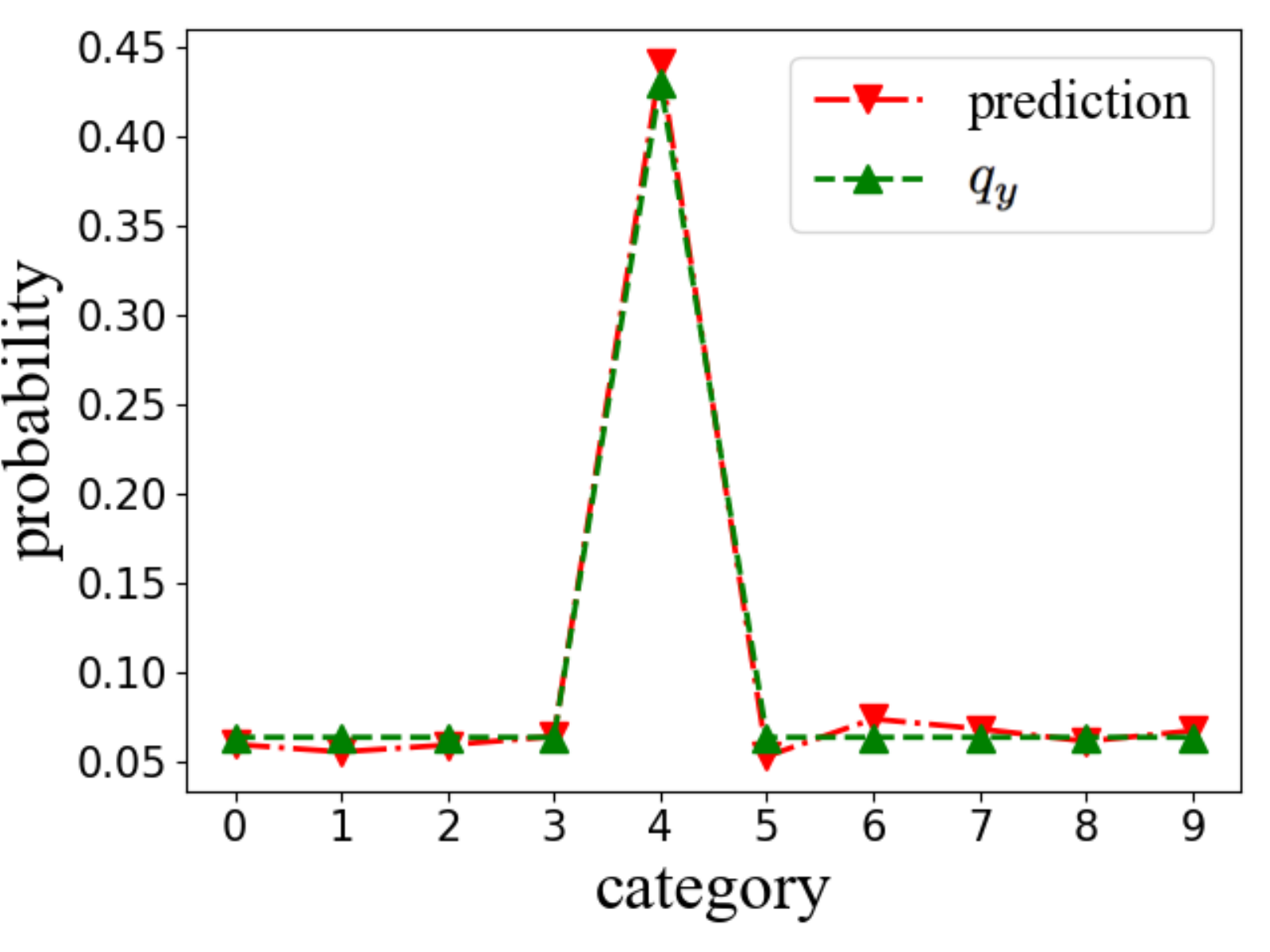}}}
    \subfigure[$\beta=0.4$, digit of $5$]{
    \centering{\includegraphics[width=0.22\columnwidth]{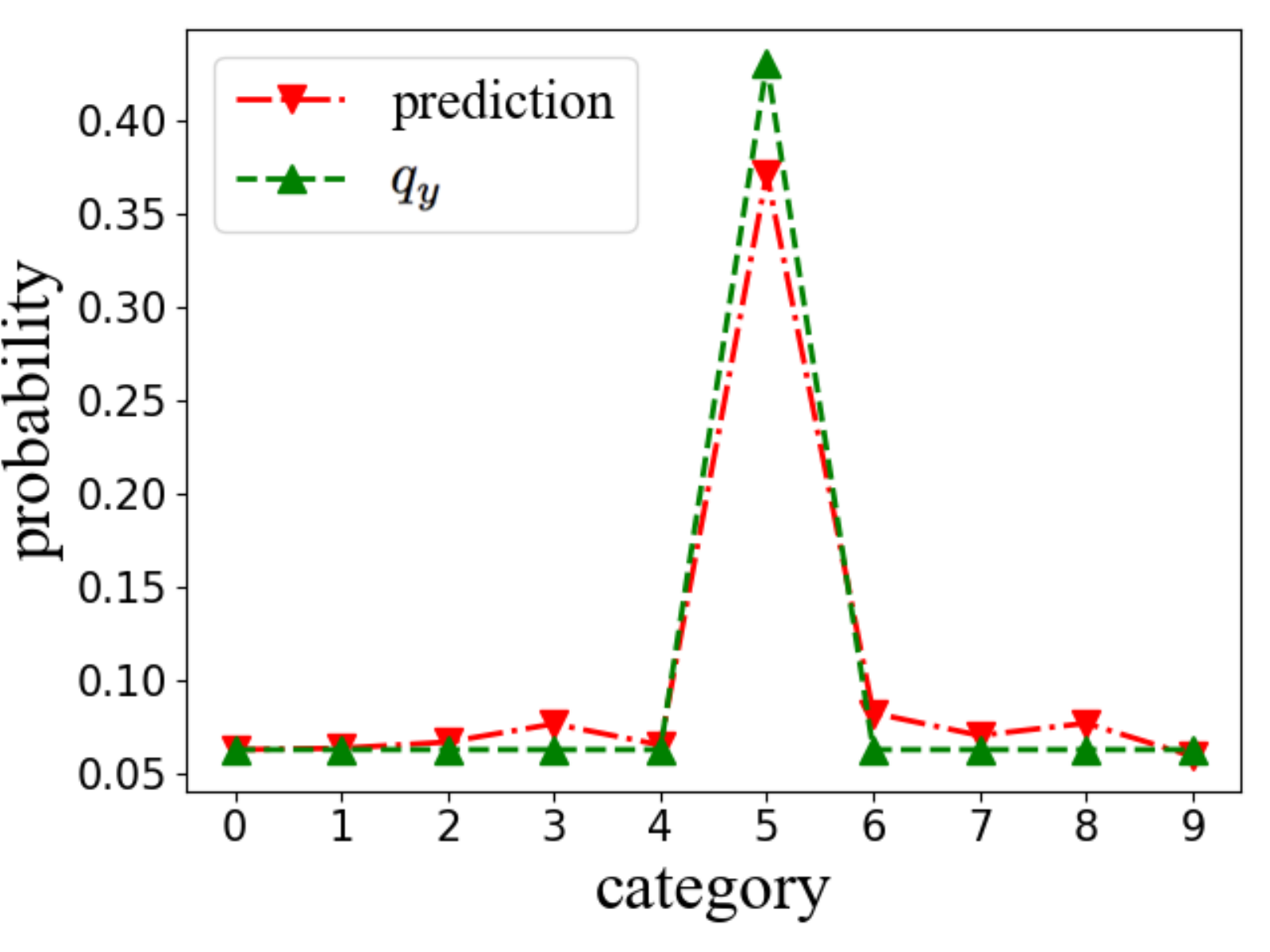}}}
    \subfigure[$\beta=0.4$, digit of $6$]{
    \centering{\includegraphics[width=0.22\columnwidth]{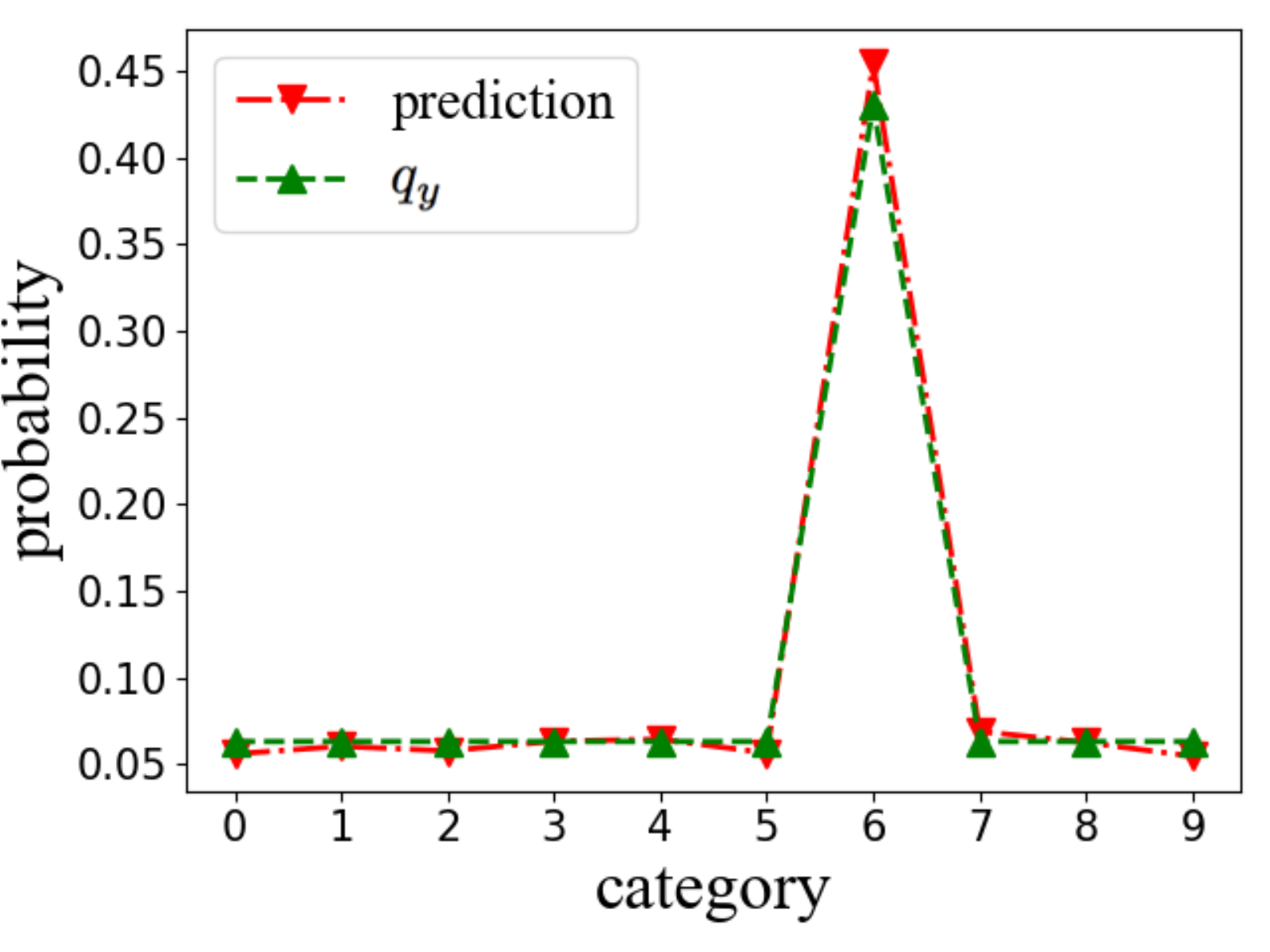}}}
    \subfigure[$\beta=0.4$, digit of $7$]{
    \centering{\includegraphics[width=0.22\columnwidth]{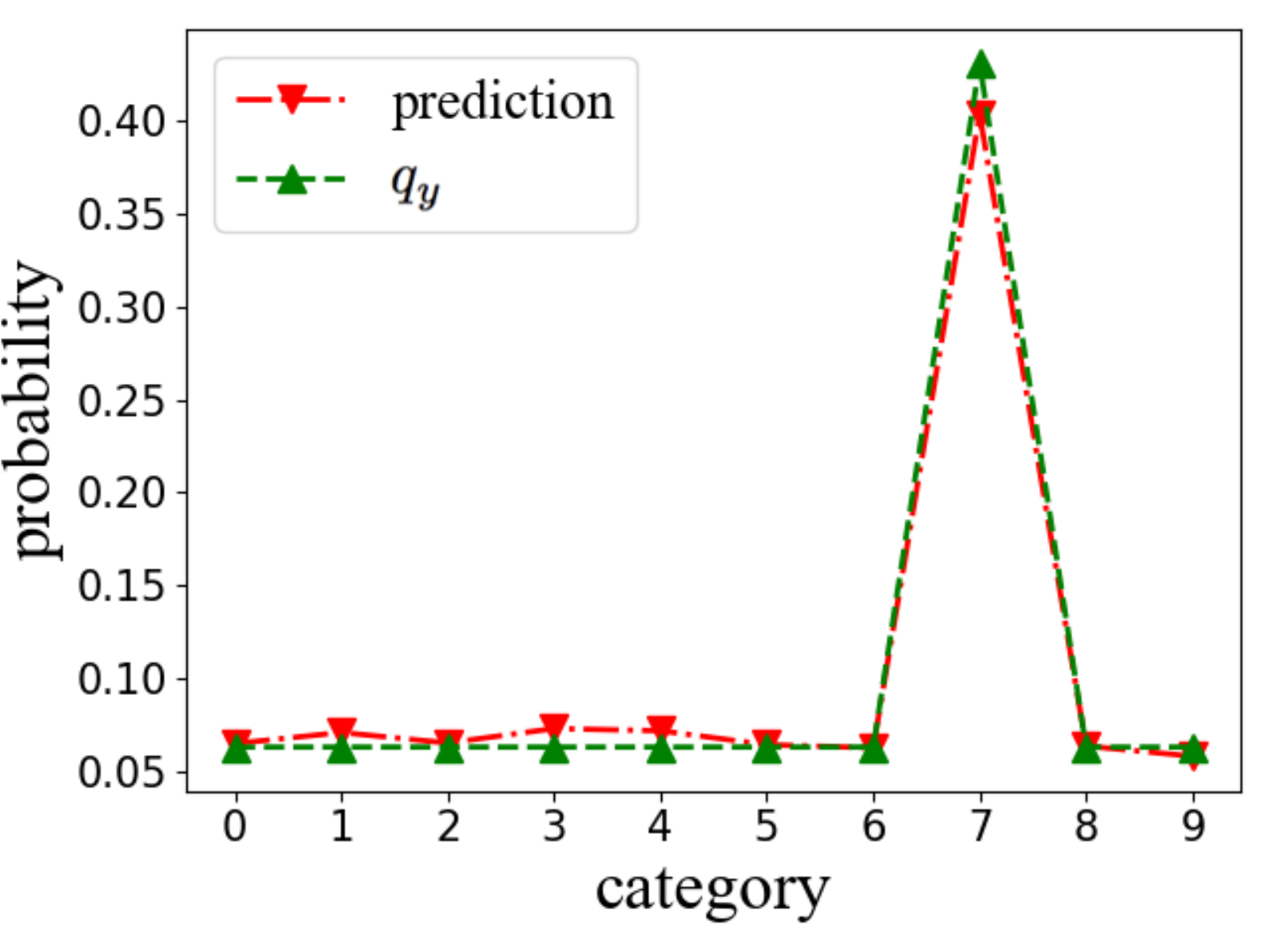}}}
    
    \subfigure[$\beta=0.4$, digit of $8$]{
    \centering{\includegraphics[width=0.22\columnwidth]{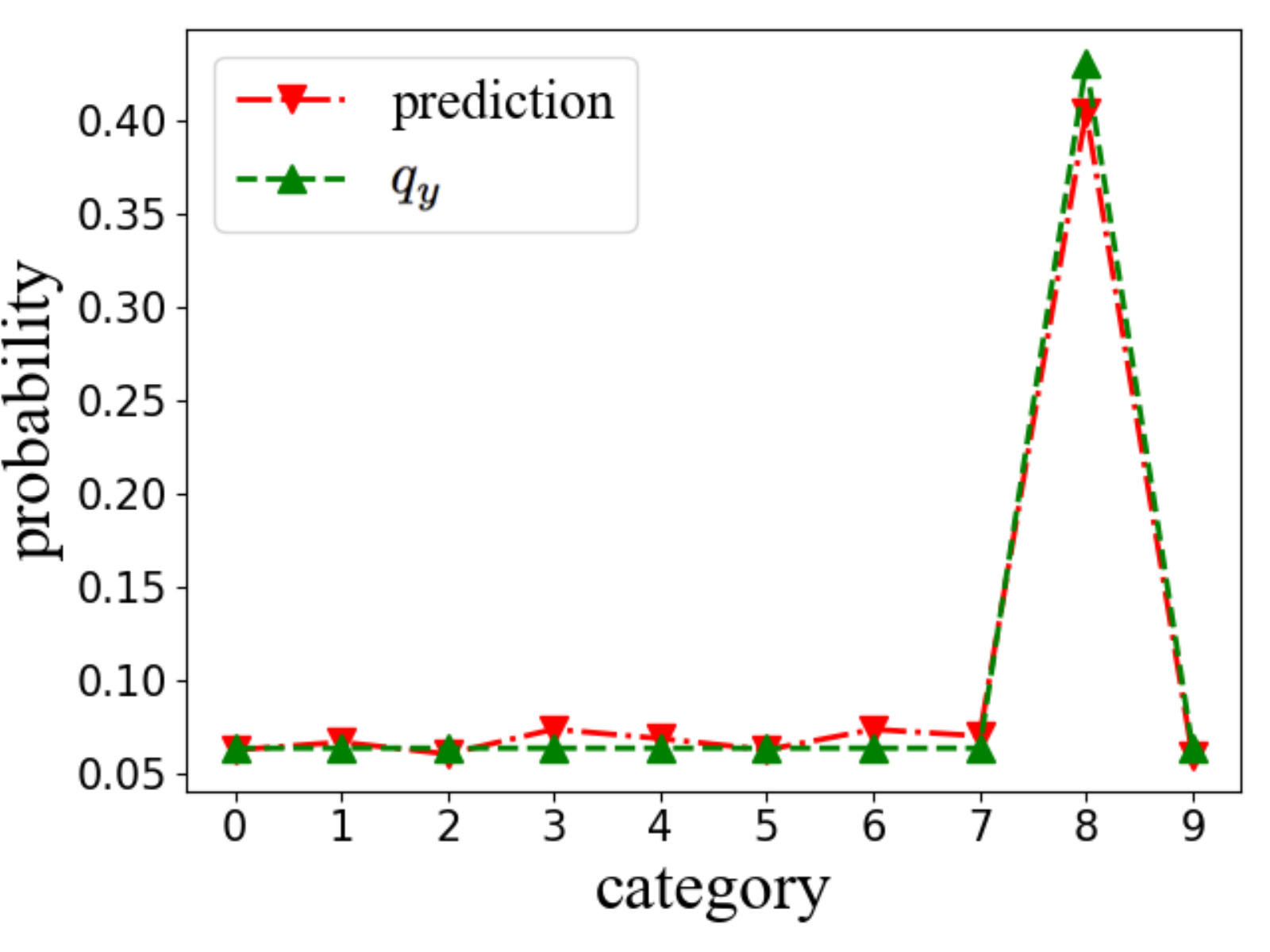}}}
    \subfigure[$\beta=0.4$, digit of $9$]{
    \centering{\includegraphics[width=0.22\columnwidth]{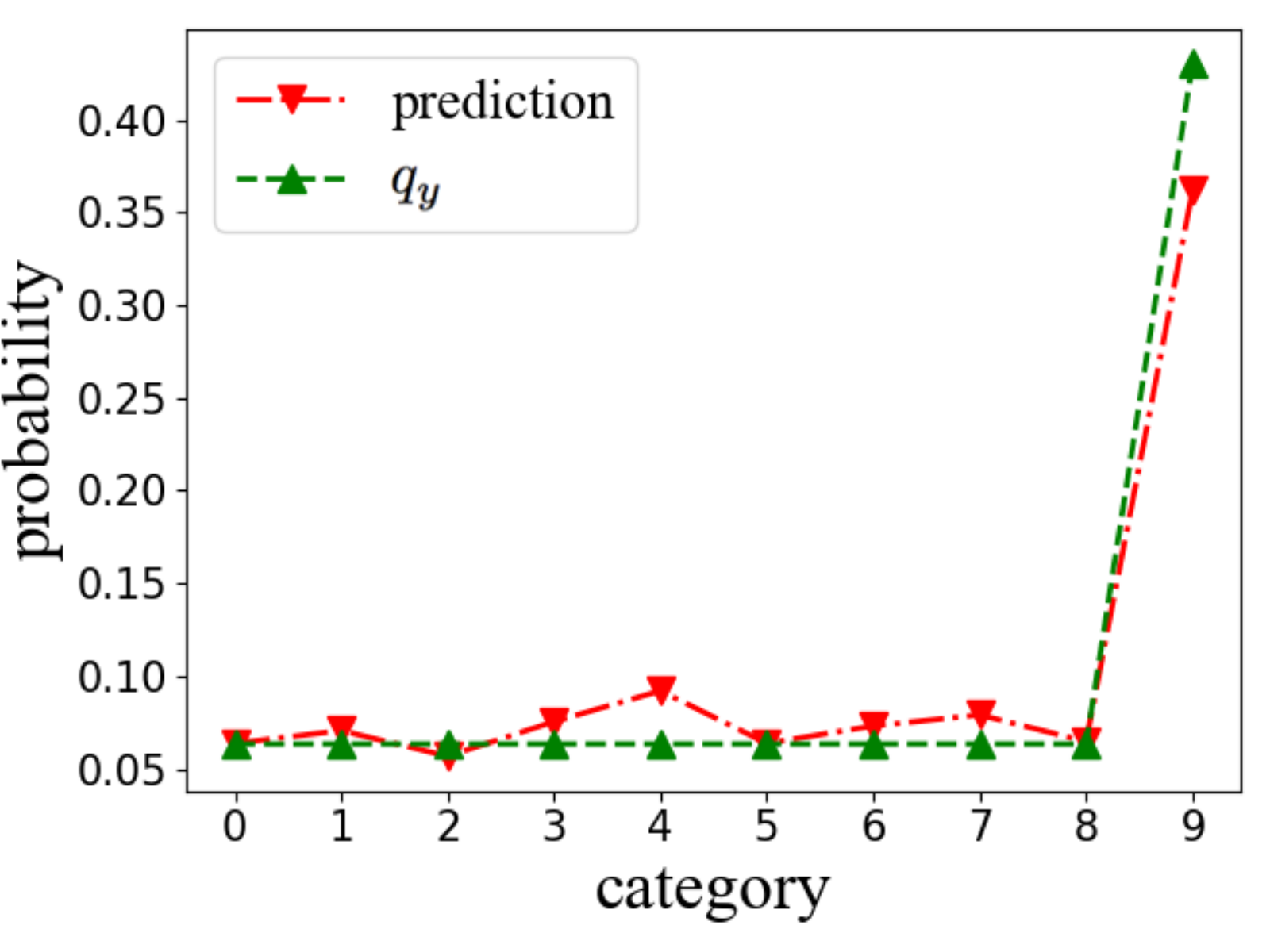}}}
    \subfigure[$\beta=0.4$, instance]{
    \centering{\includegraphics[width=0.22\columnwidth]{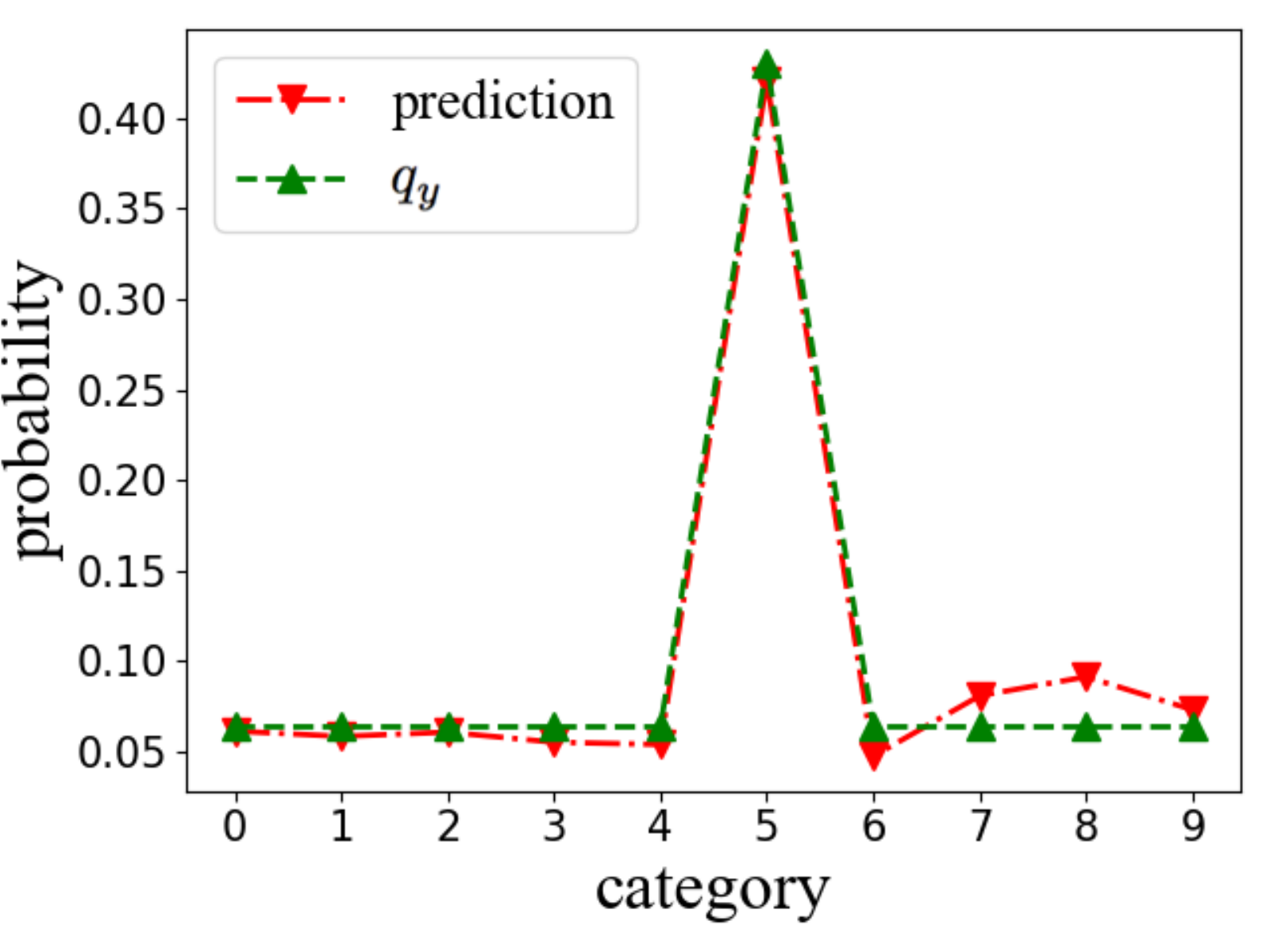}}}
    \subfigure[$\beta=0.4$, instance]{
    \centering{\includegraphics[width=0.22\columnwidth]{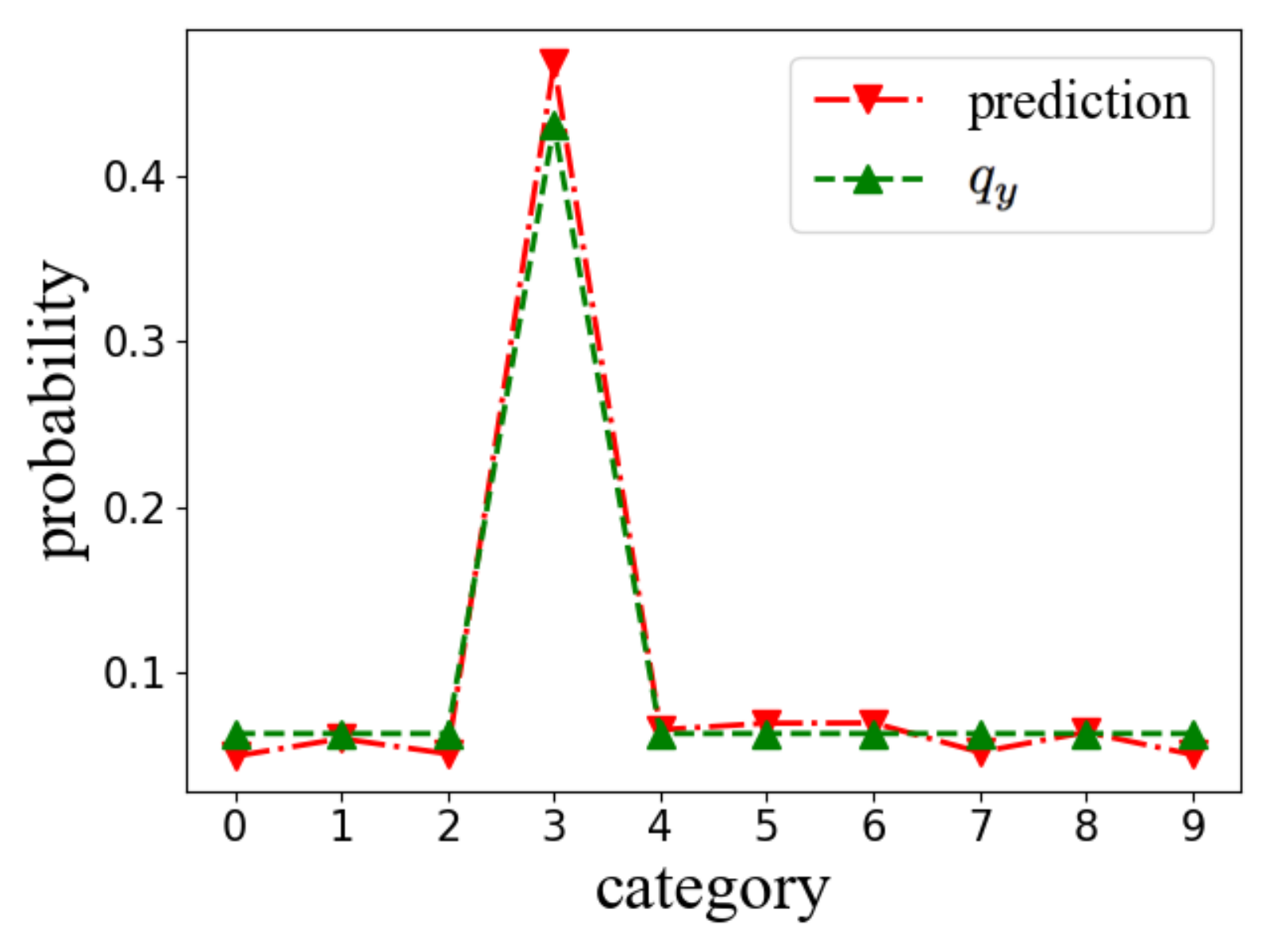}}}
    
    \subfigure[$\beta=0.8$, digit of $0$]{
    \centering{\includegraphics[width=0.22\columnwidth]{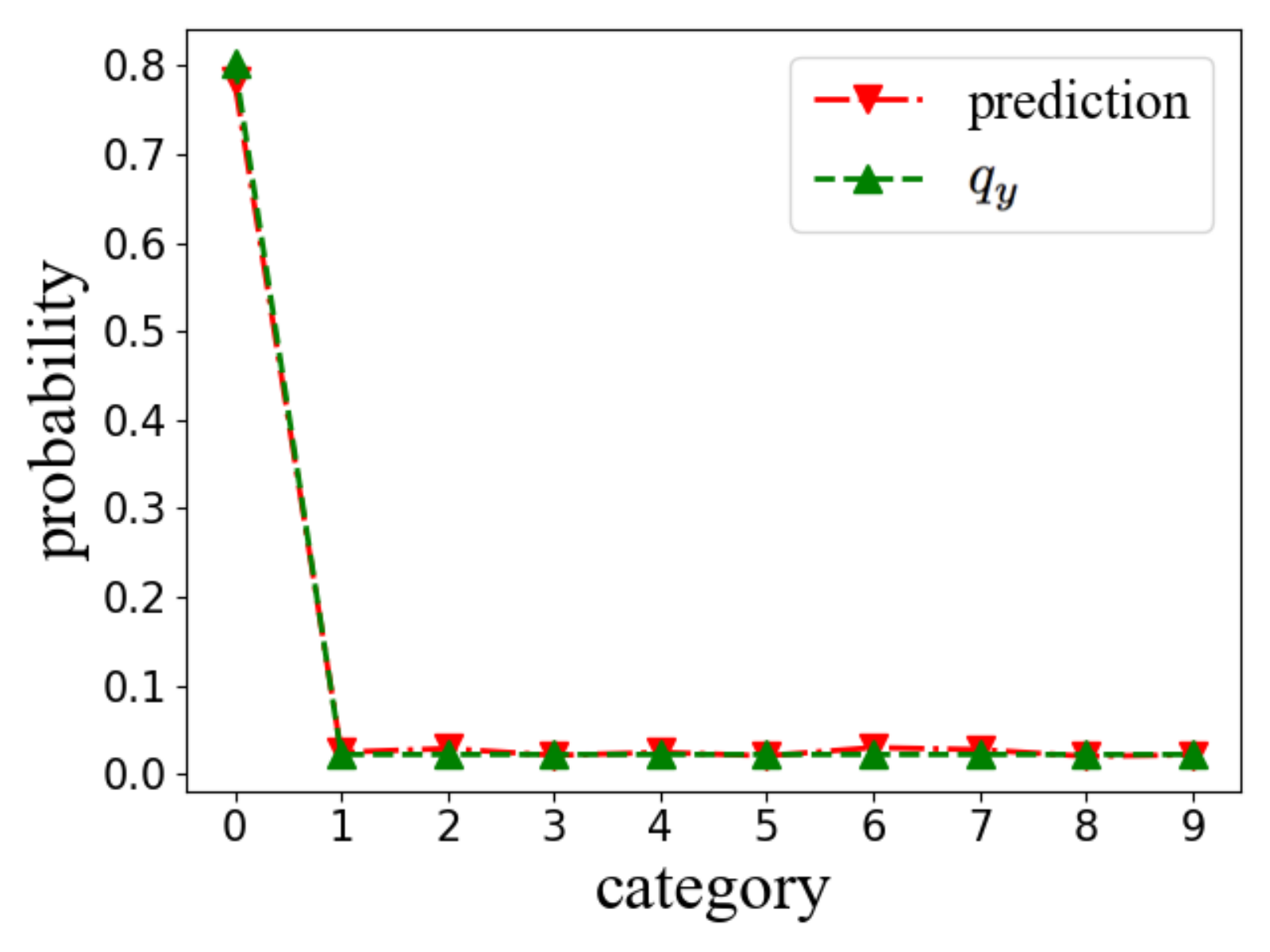}}}
    \subfigure[$\beta=0.8$, digit of $1$]{
    \centering{\includegraphics[width=0.22\columnwidth]{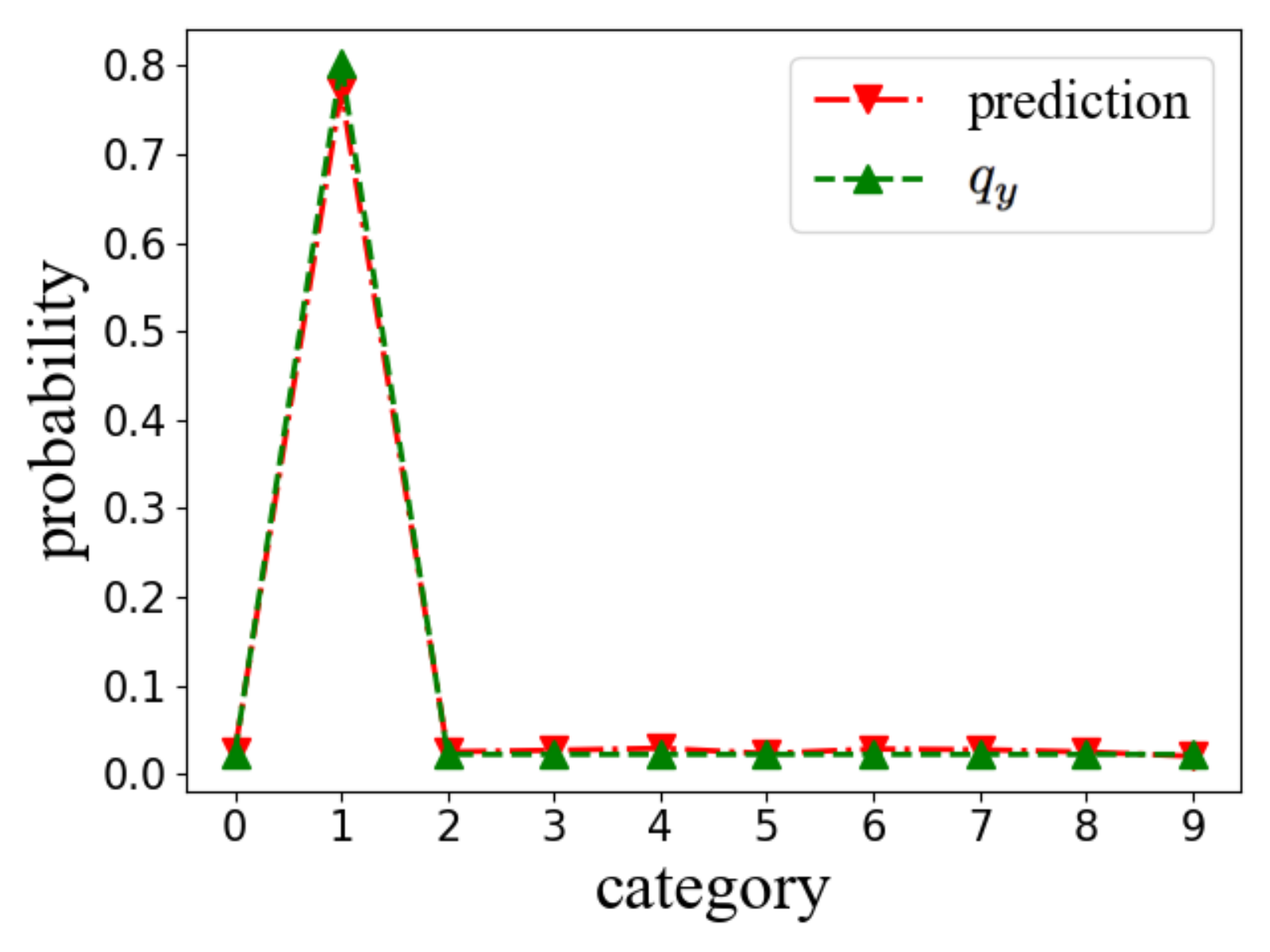}}}
    \subfigure[$\beta=0.8$, digit of $2$]{
    \centering{\includegraphics[width=0.22\columnwidth]{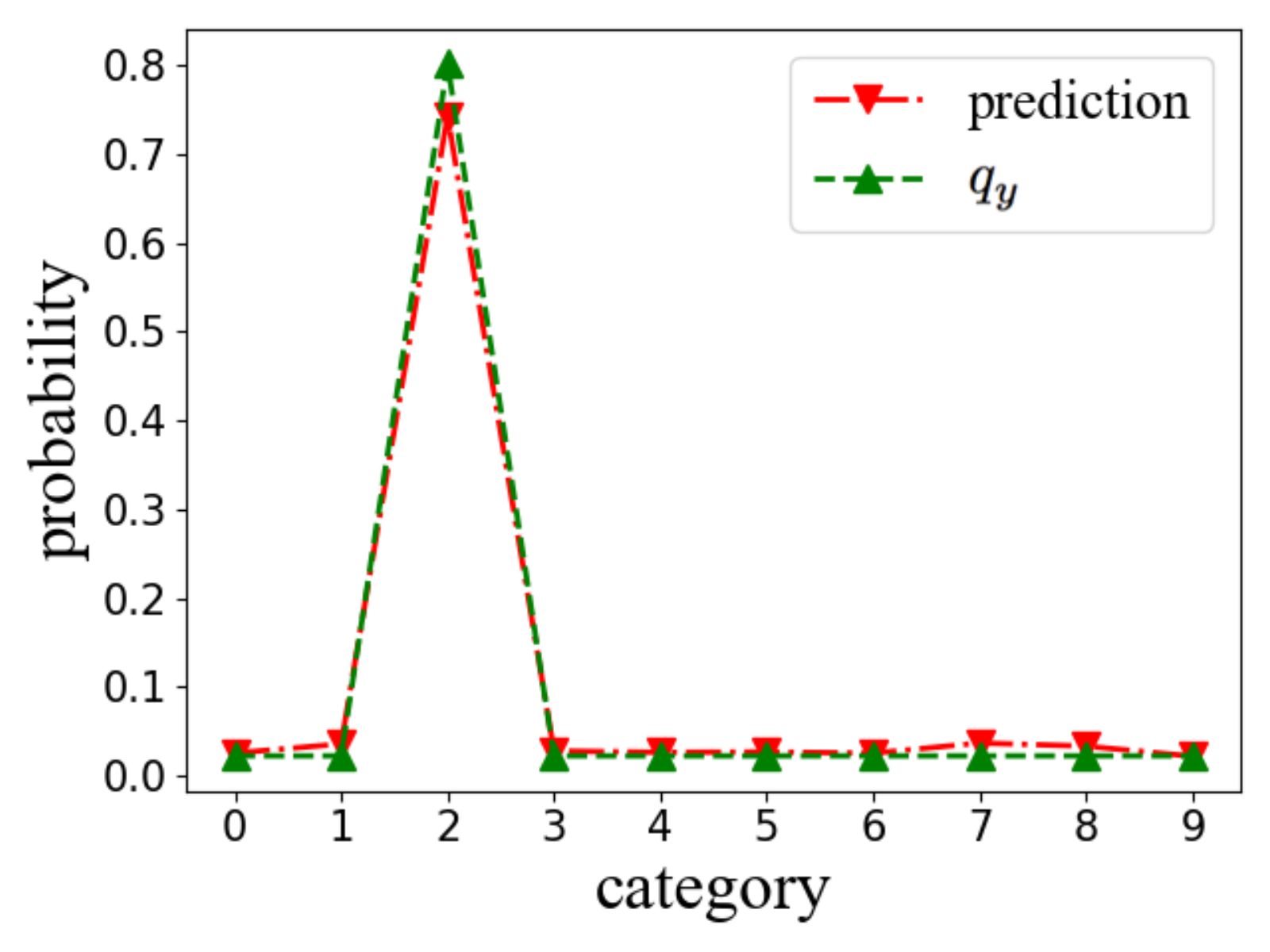}}}
    \subfigure[$\beta=0.8$, digit of $3$]{
    \centering{\includegraphics[width=0.22\columnwidth]{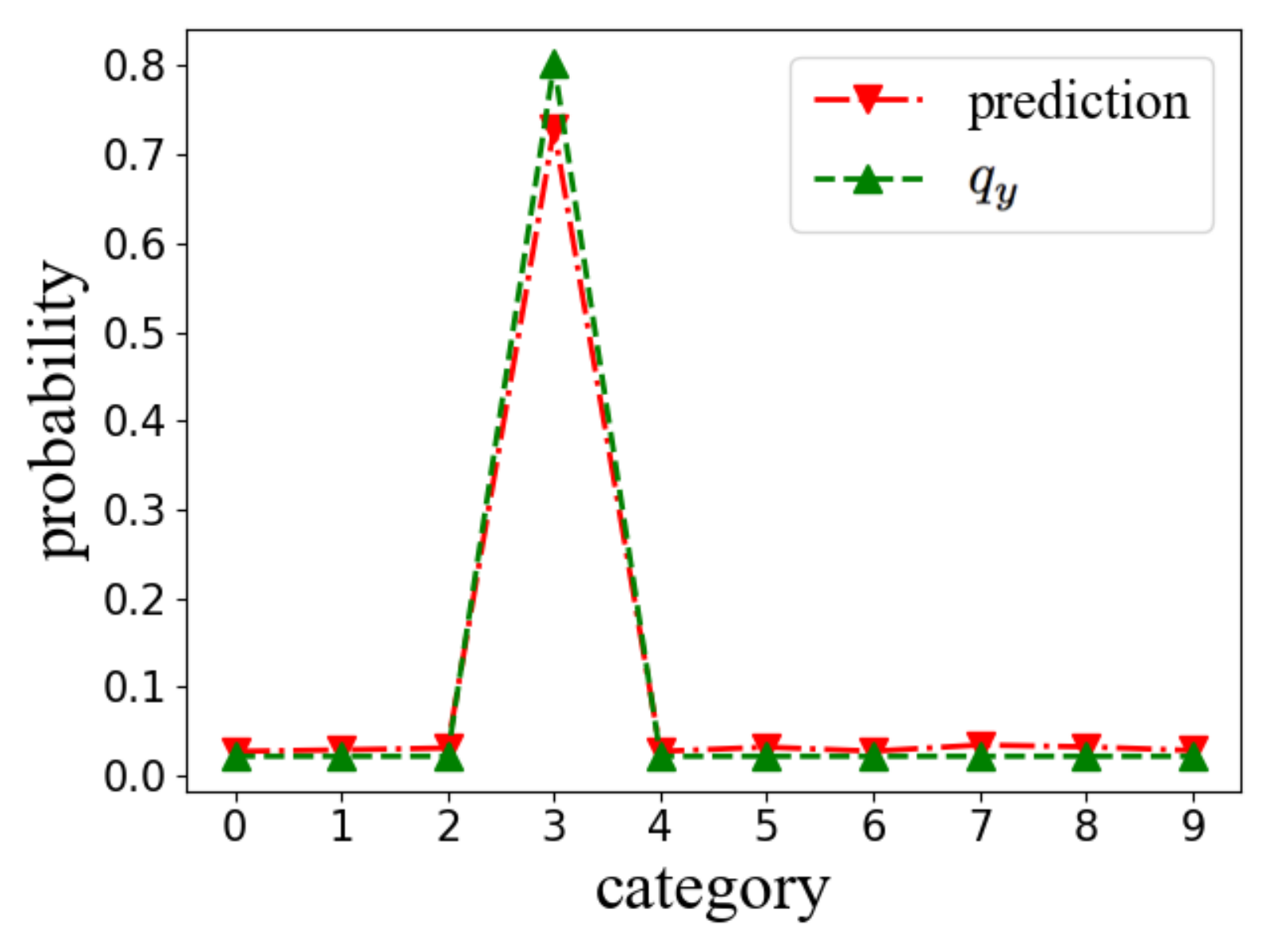}}}
    
    \subfigure[$\beta=0.8$, digit of $4$]{
    \centering{\includegraphics[width=0.22\columnwidth]{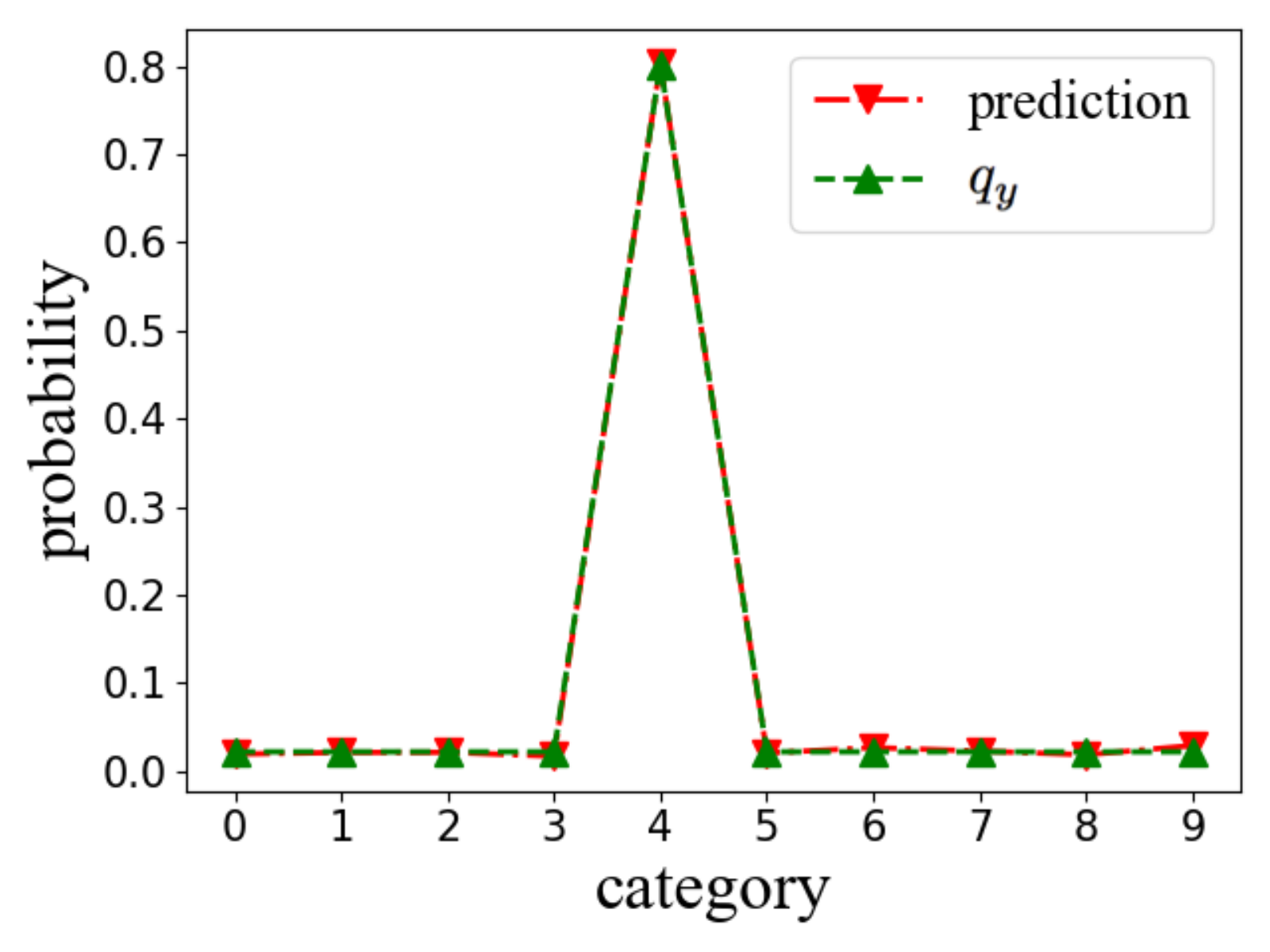}}}
    \subfigure[$\beta=0.8$, digit of $5$]{
    \centering{\includegraphics[width=0.22\columnwidth]{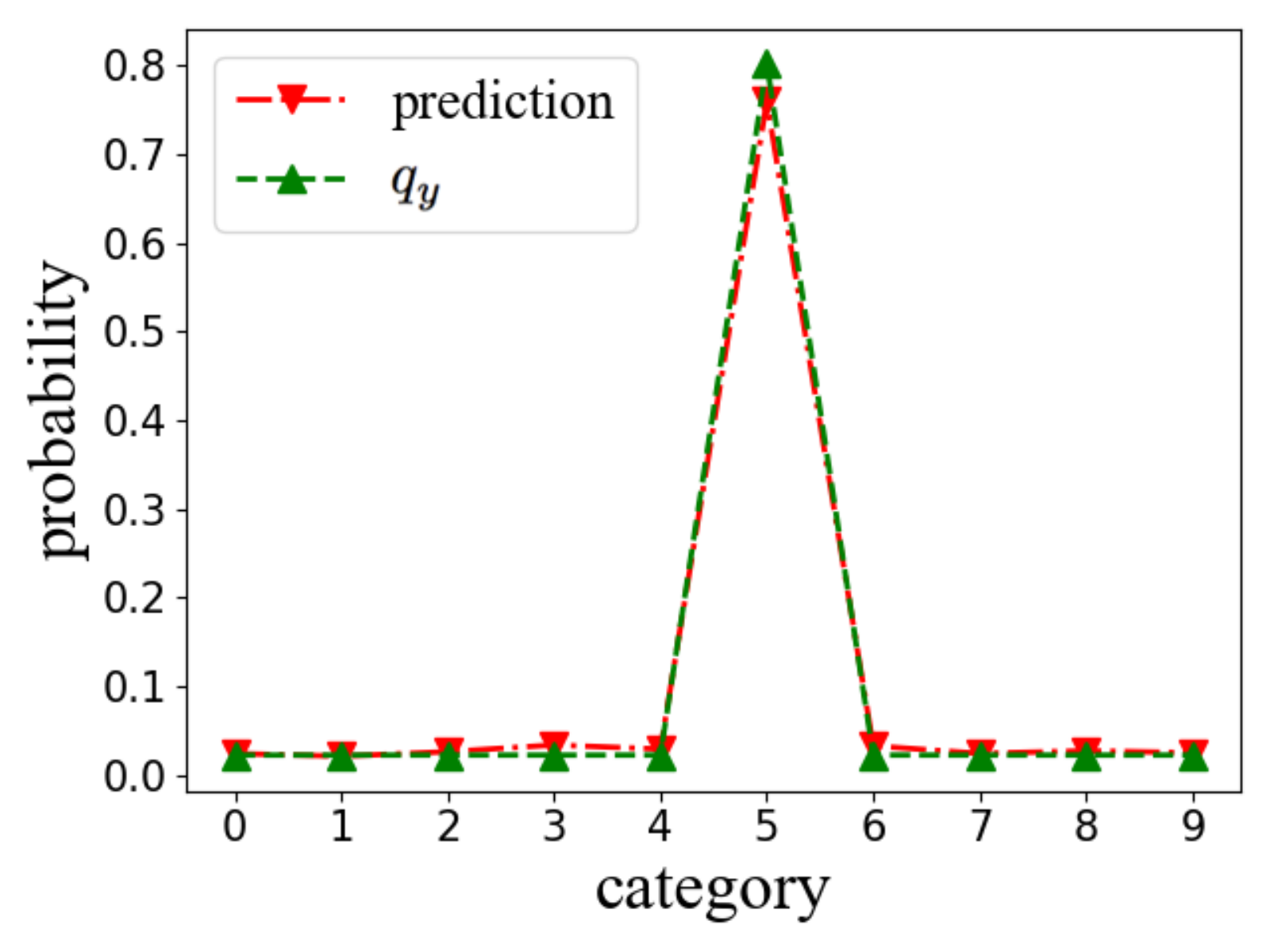}}}
    \subfigure[$\beta=0.8$, digit of $6$]{
    \centering{\includegraphics[width=0.22\columnwidth]{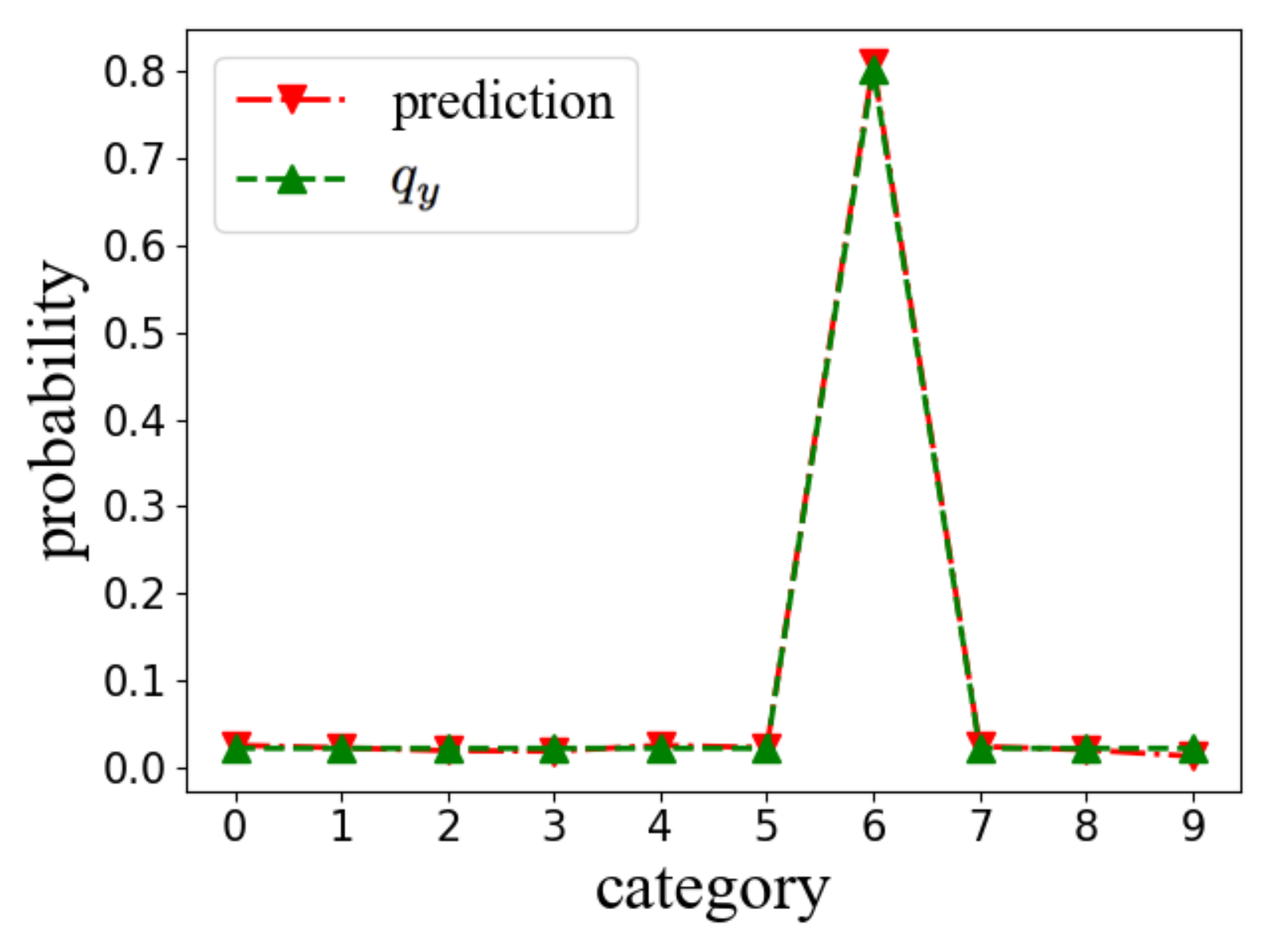}}}
    \subfigure[$\beta=0.8$, digit of $7$]{
    \centering{\includegraphics[width=0.22\columnwidth]{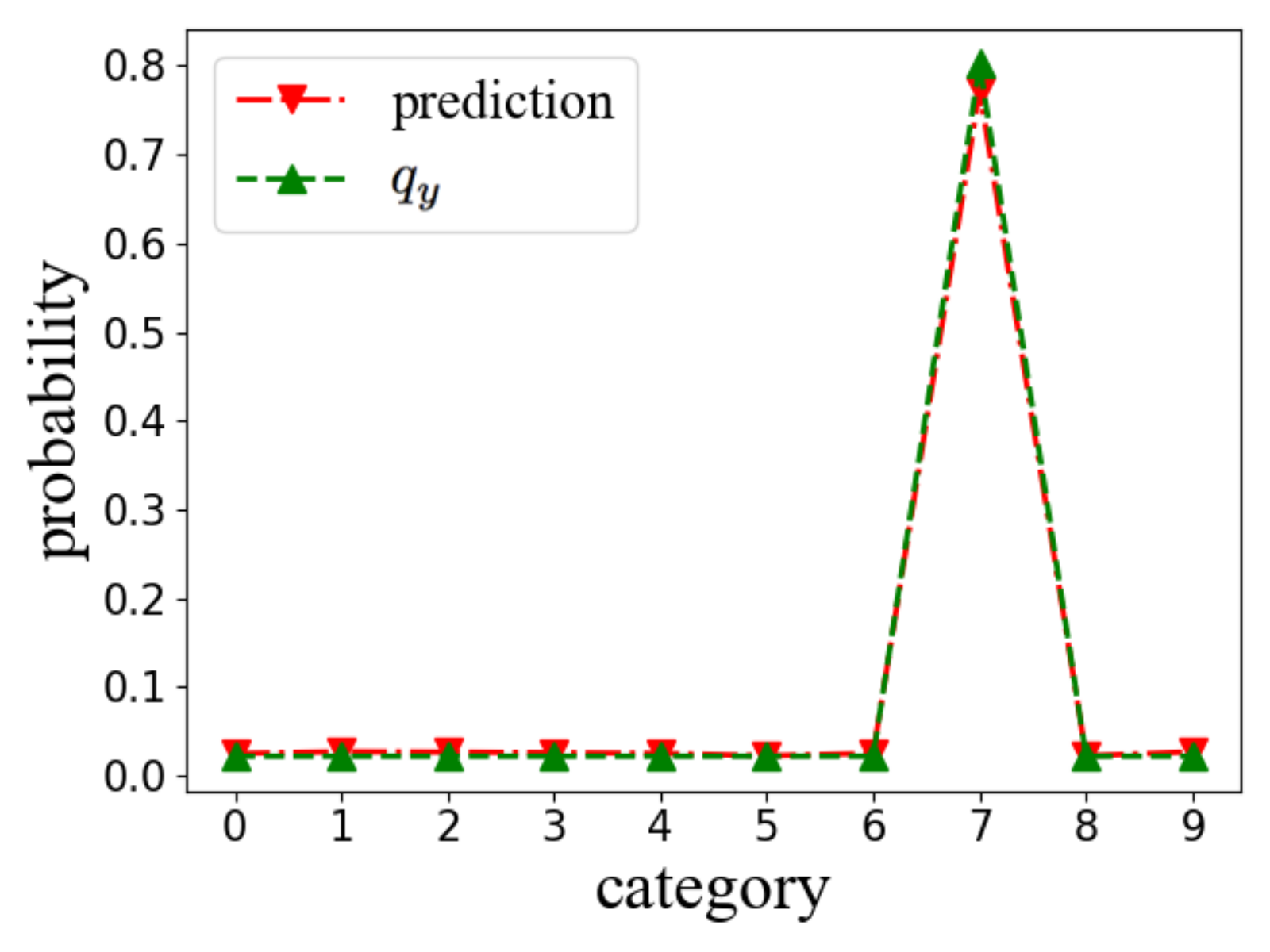}}}
    
    \subfigure[$\beta=0.8$, digit of $8$]{
    \centering{\includegraphics[width=0.22\columnwidth]{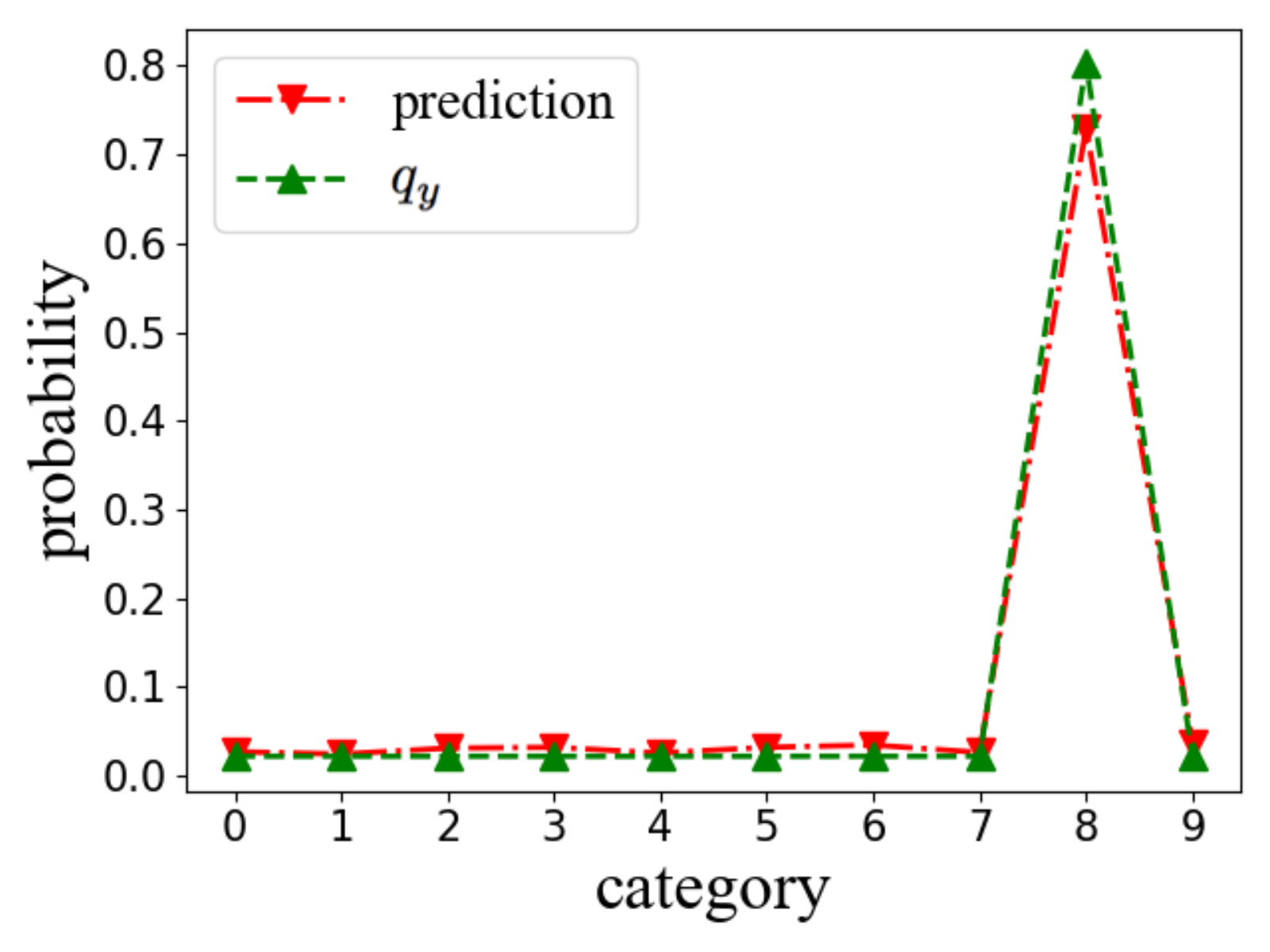}}}
    \subfigure[$\beta=0.8$, digit of $9$]{
    \centering{\includegraphics[width=0.22\columnwidth]{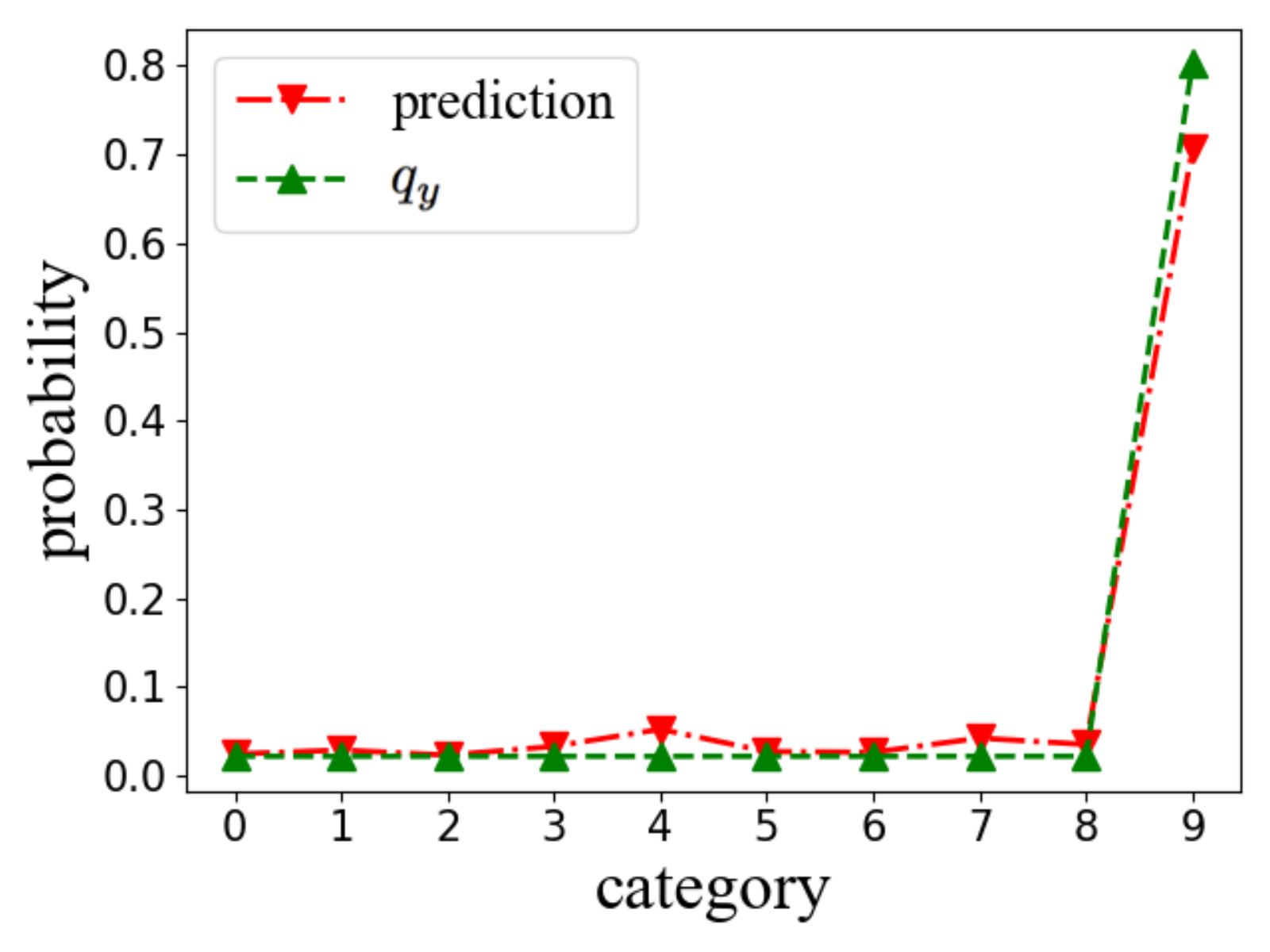}}}
    \subfigure[$\beta=0.8$, instance]{
    \centering{\includegraphics[width=0.22\columnwidth]{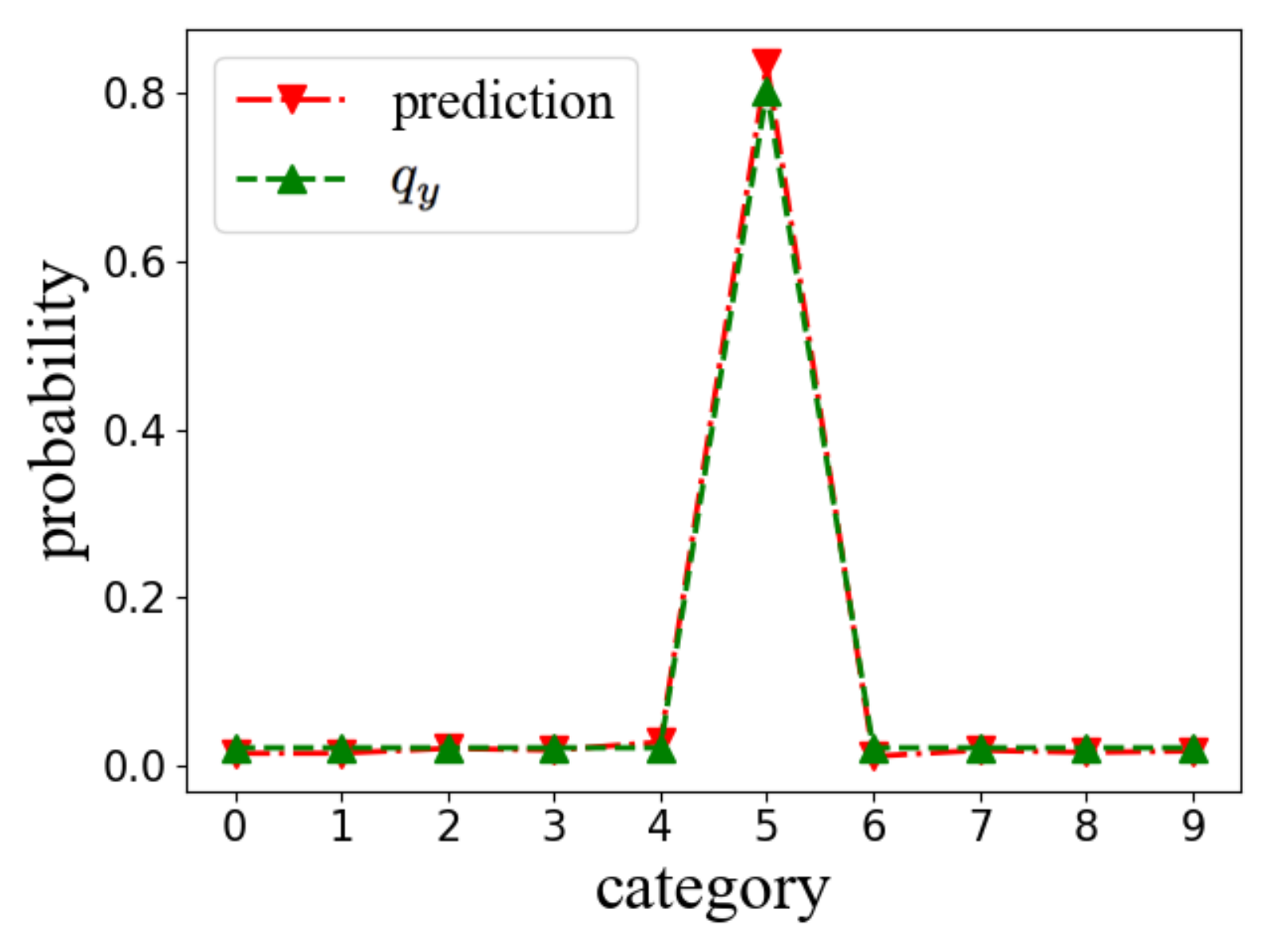}}}
    \subfigure[$\beta=0.8$, instance]{
    \centering{\includegraphics[width=0.22\columnwidth]{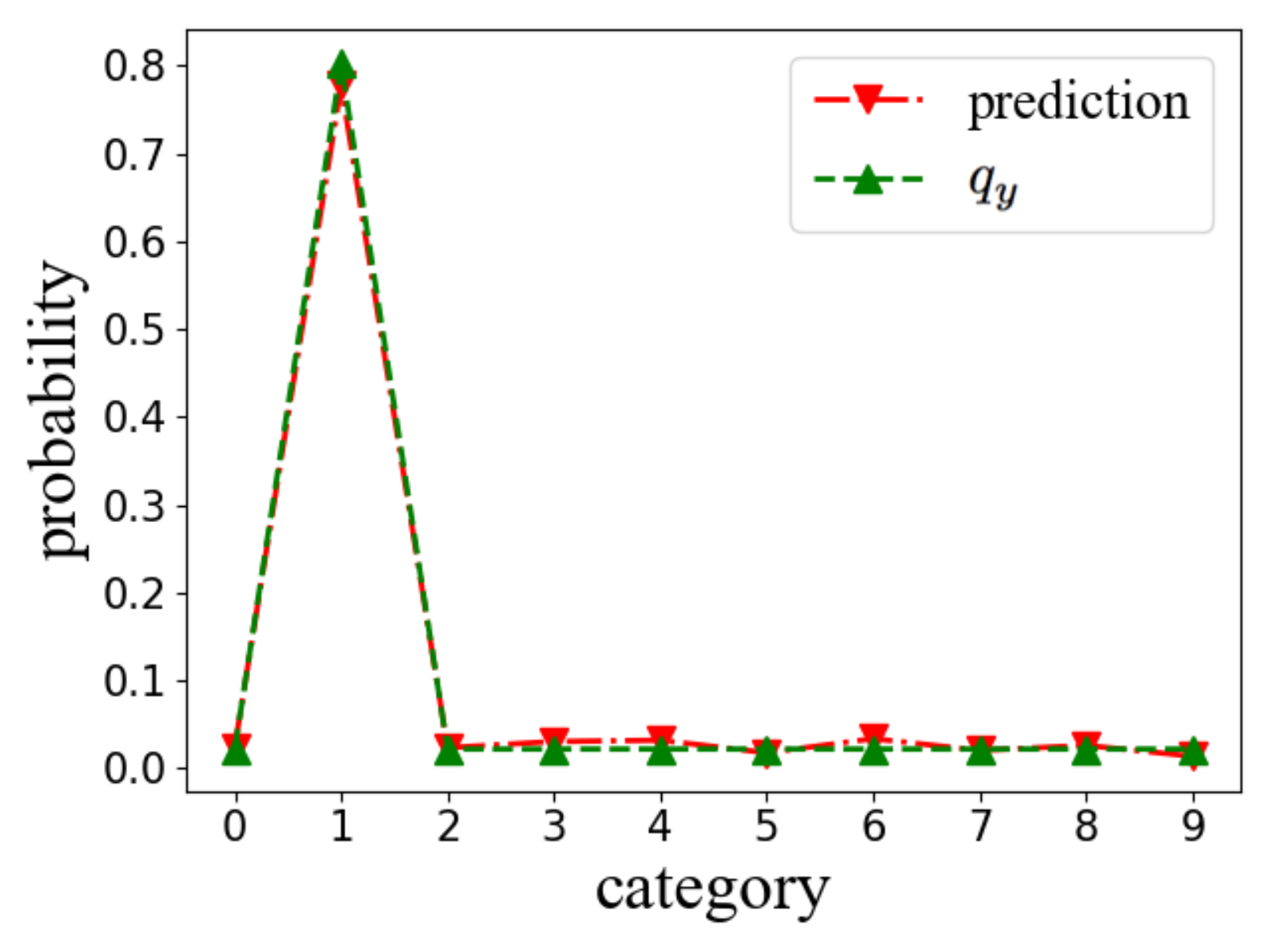}}}

\caption{Uniform noise when $\beta=0.4$ and $\beta=0.8$. (k), (l), (w) and (x): instances of digit $5$. Others are the average of predicted probabilities and $q_y$ for all images of different digits. The red lines represent $f(y; \theta)$ whereas the green lines represent ${q}_{y}$.}
\label{fig:uniform_noise}
\end{center}
\end{figure}

\begin{figure}[t]
\begin{center}
    \subfigure[$\beta=0.4$, digit of $0$]{
    \centering{\includegraphics[width=0.22\columnwidth]{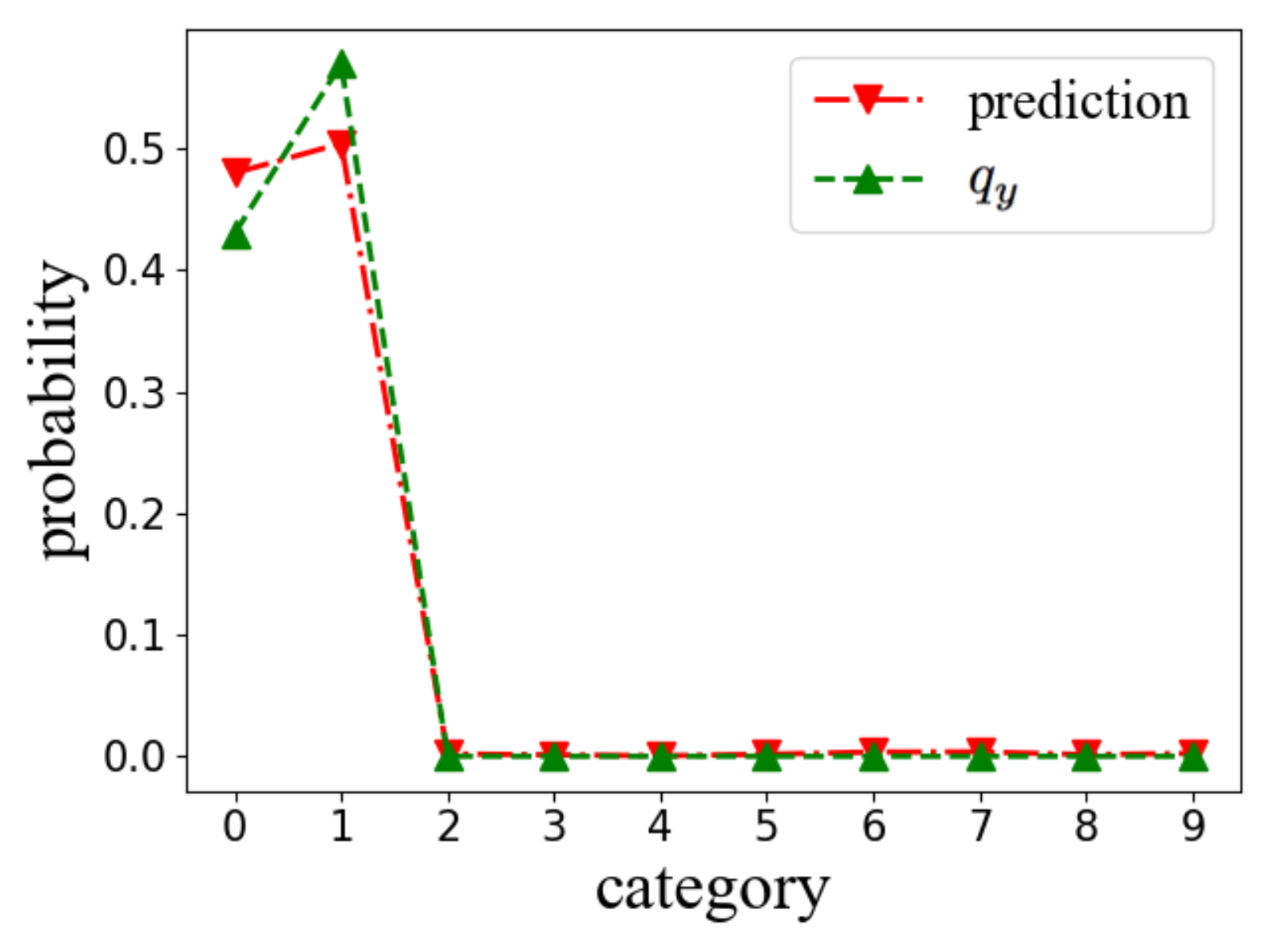}}}
    \subfigure[$\beta=0.4$, digit of $1$]{
    \centering{\includegraphics[width=0.22\columnwidth]{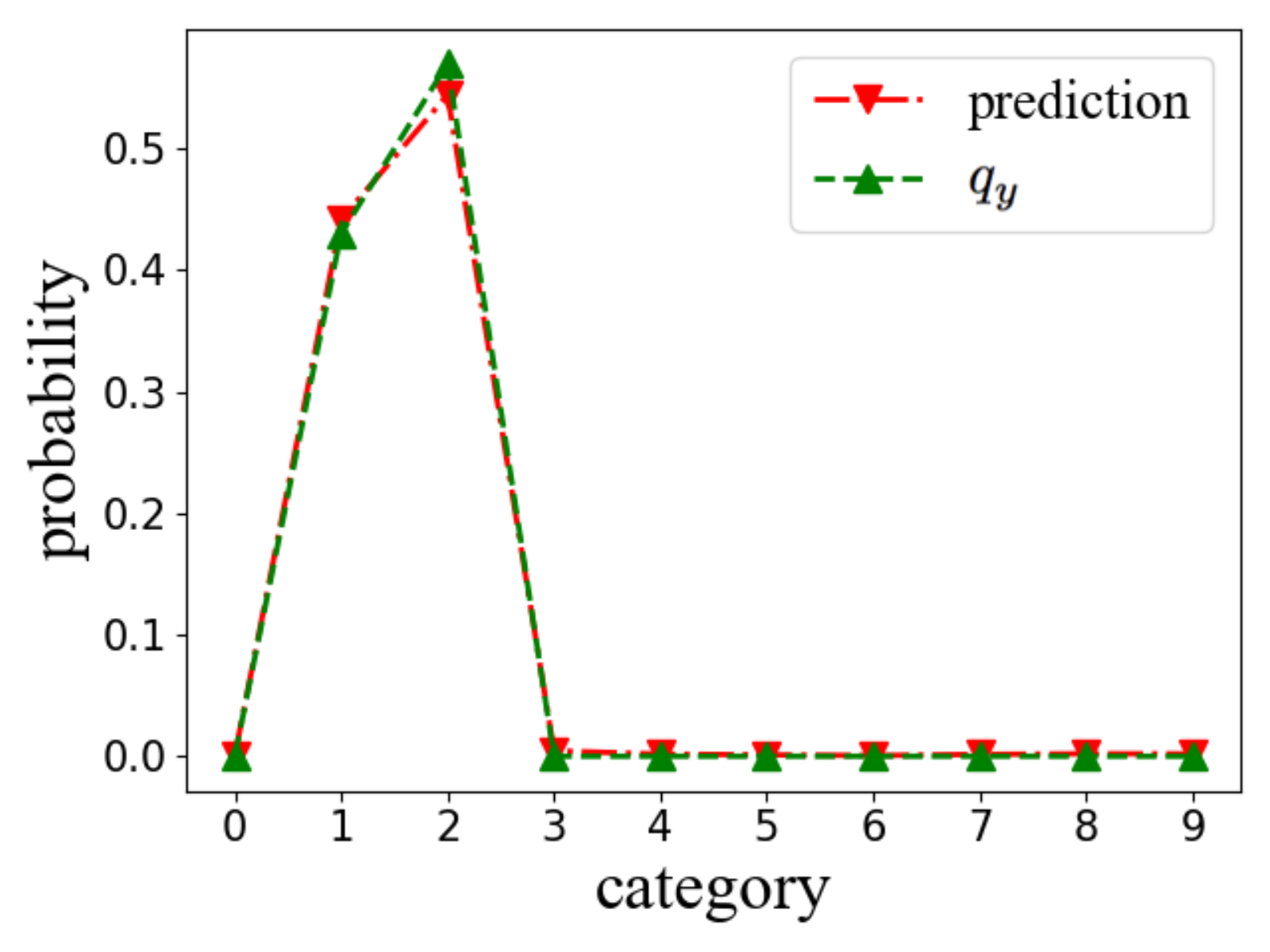}}}
    \subfigure[$\beta=0.4$, digit of $2$]{
    \centering{\includegraphics[width=0.22\columnwidth]{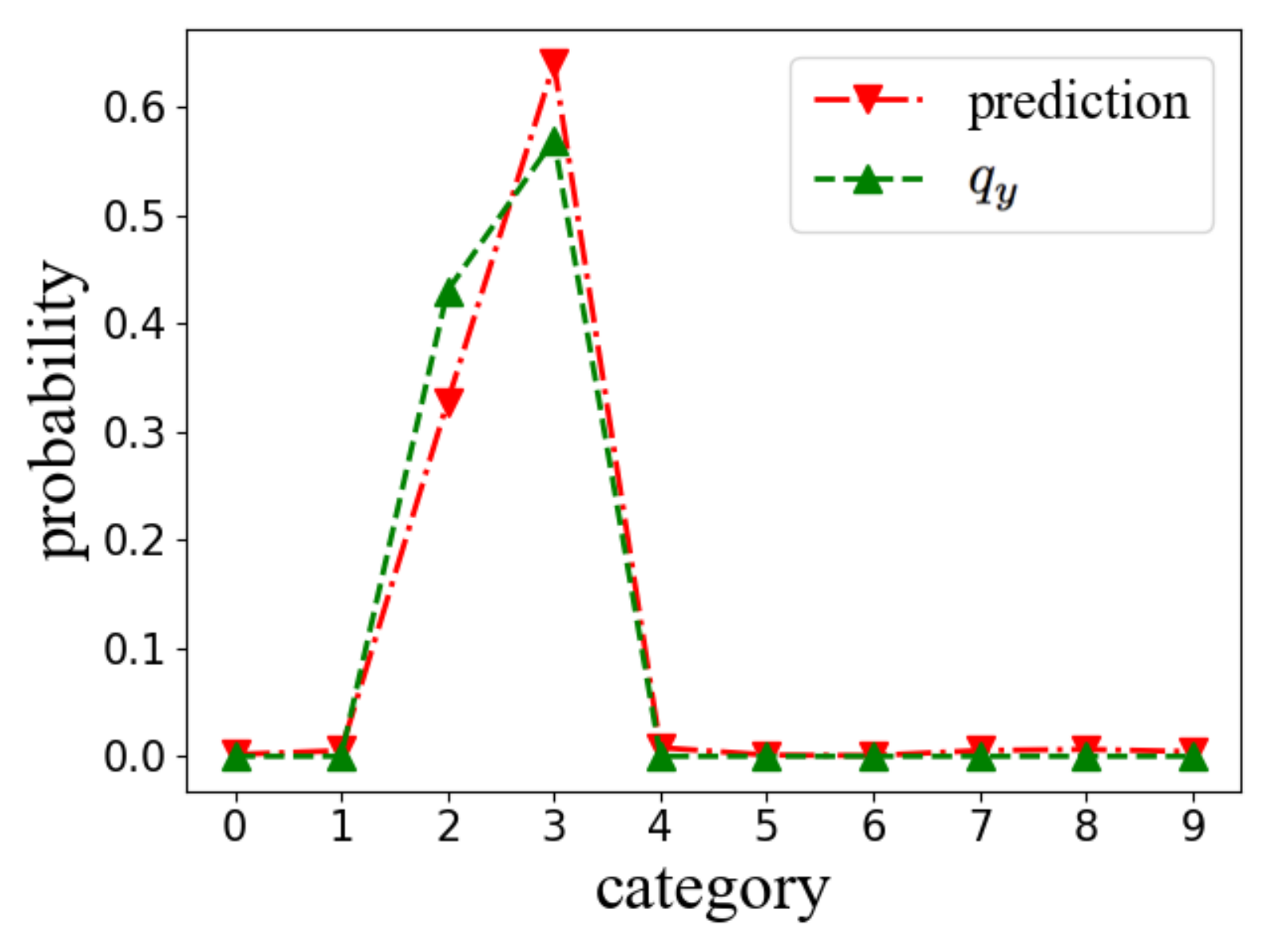}}}
    \subfigure[$\beta=0.4$, digit of $3$]{
    \centering{\includegraphics[width=0.22\columnwidth]{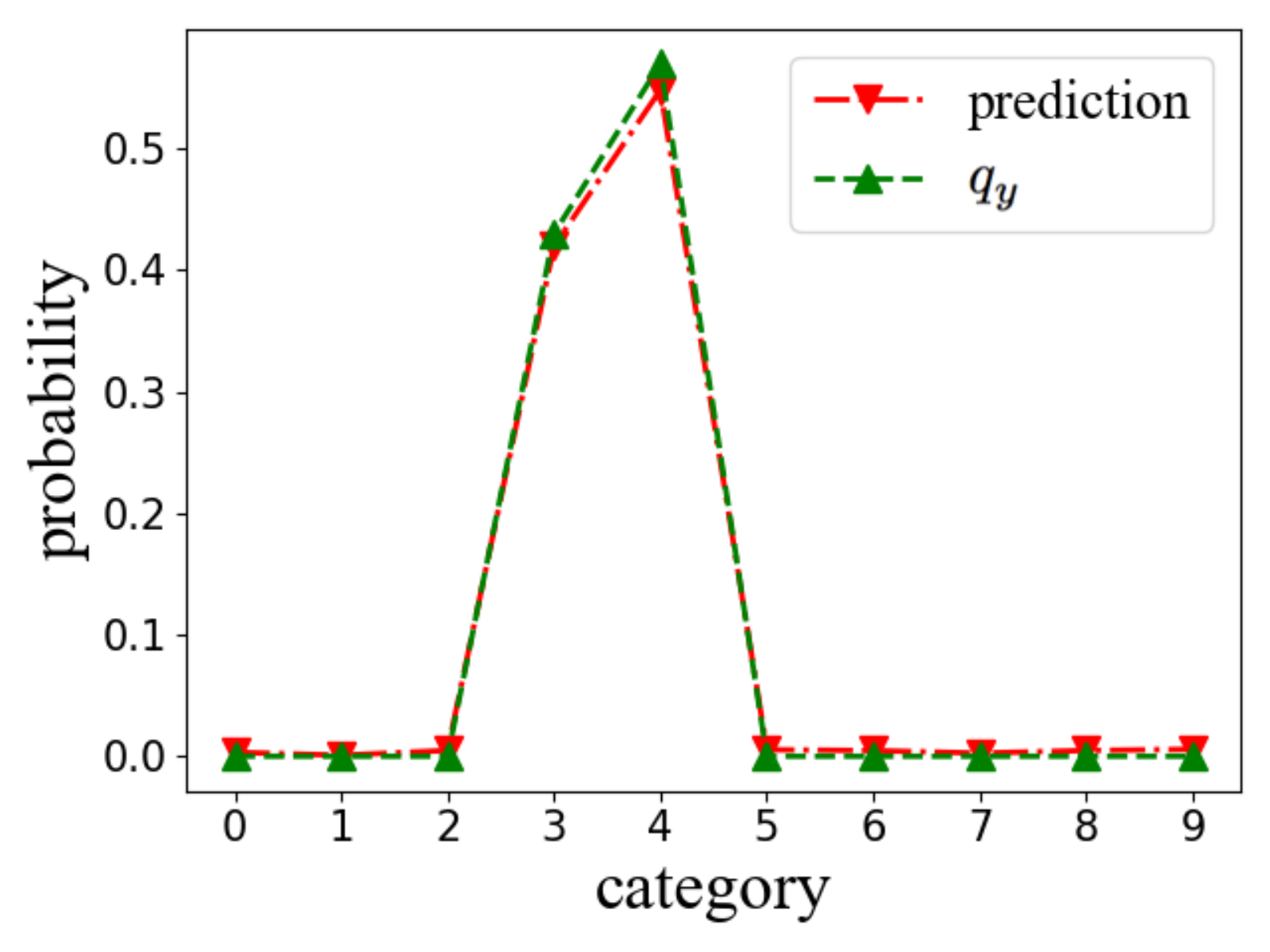}}}
    
    \subfigure[$\beta=0.4$, digit of $4$]{
    \centering{\includegraphics[width=0.22\columnwidth]{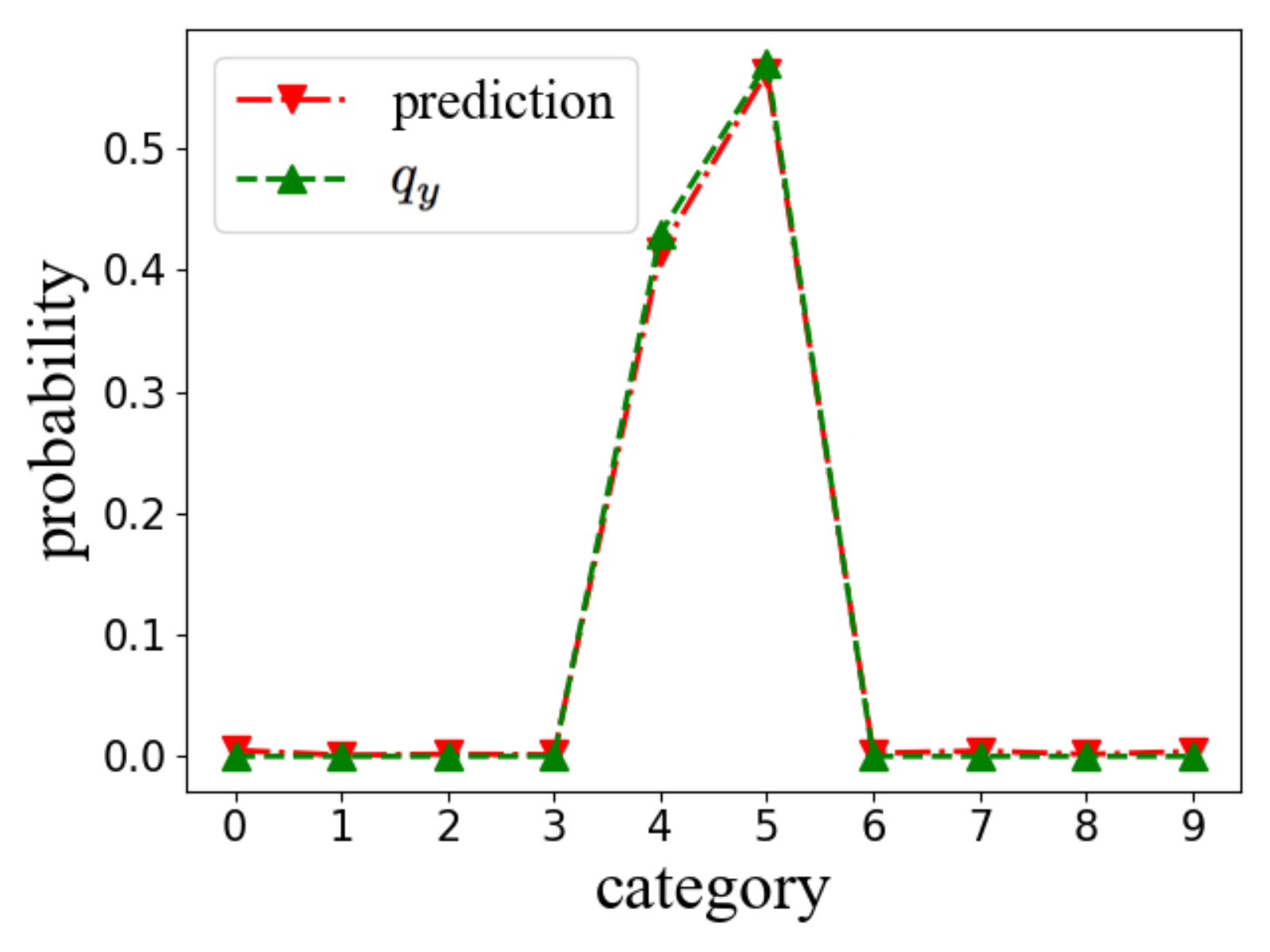}}}
    \subfigure[$\beta=0.4$, digit of $5$]{
    \centering{\includegraphics[width=0.22\columnwidth]{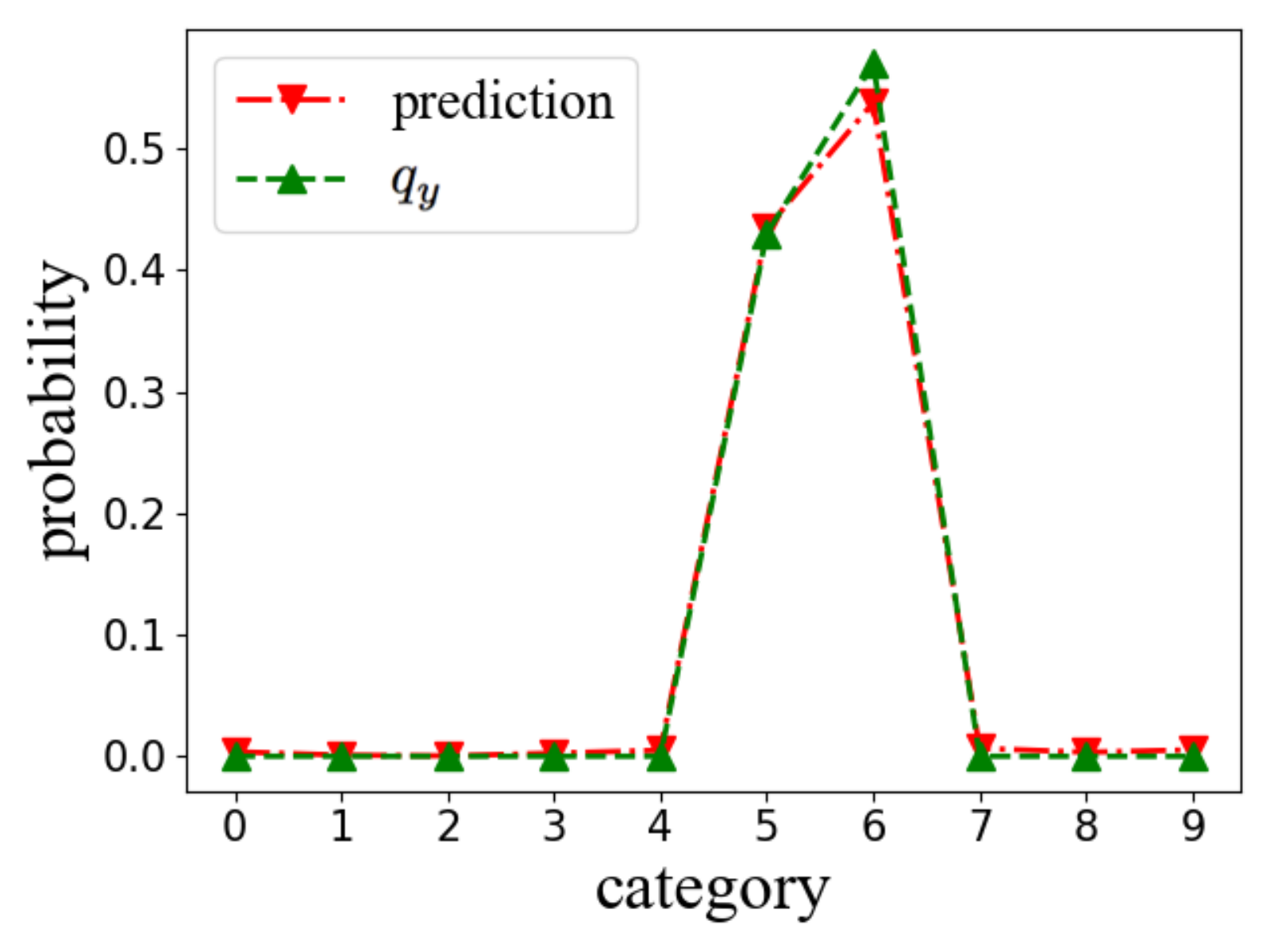}}}
    \subfigure[$\beta=0.4$, digit of $6$]{
    \centering{\includegraphics[width=0.22\columnwidth]{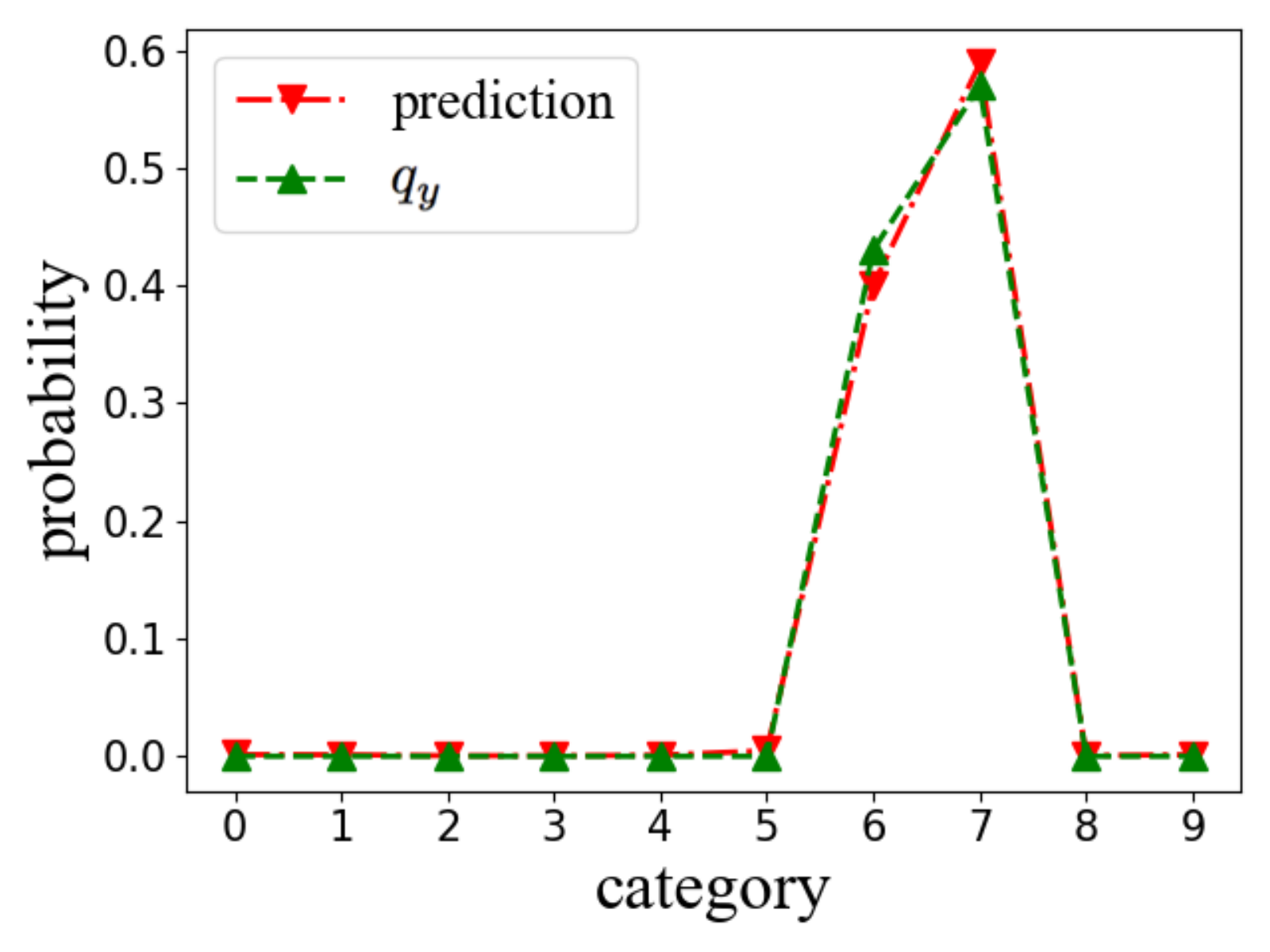}}}
    \subfigure[$\beta=0.4$, digit of $7$]{
    \centering{\includegraphics[width=0.22\columnwidth]{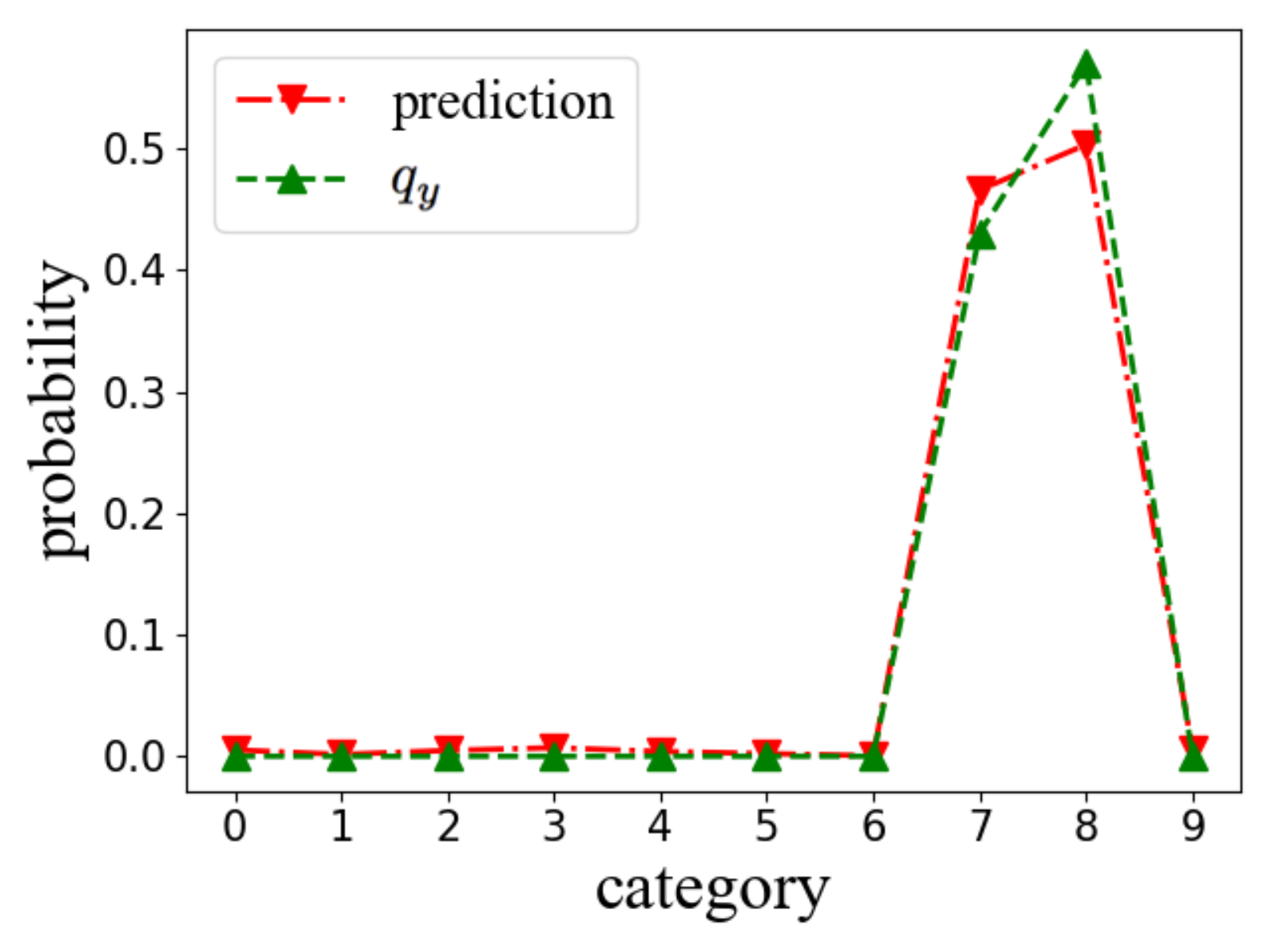}}}
    
    \subfigure[$\beta=0.4$, digit of $8$]{
    \centering{\includegraphics[width=0.22\columnwidth]{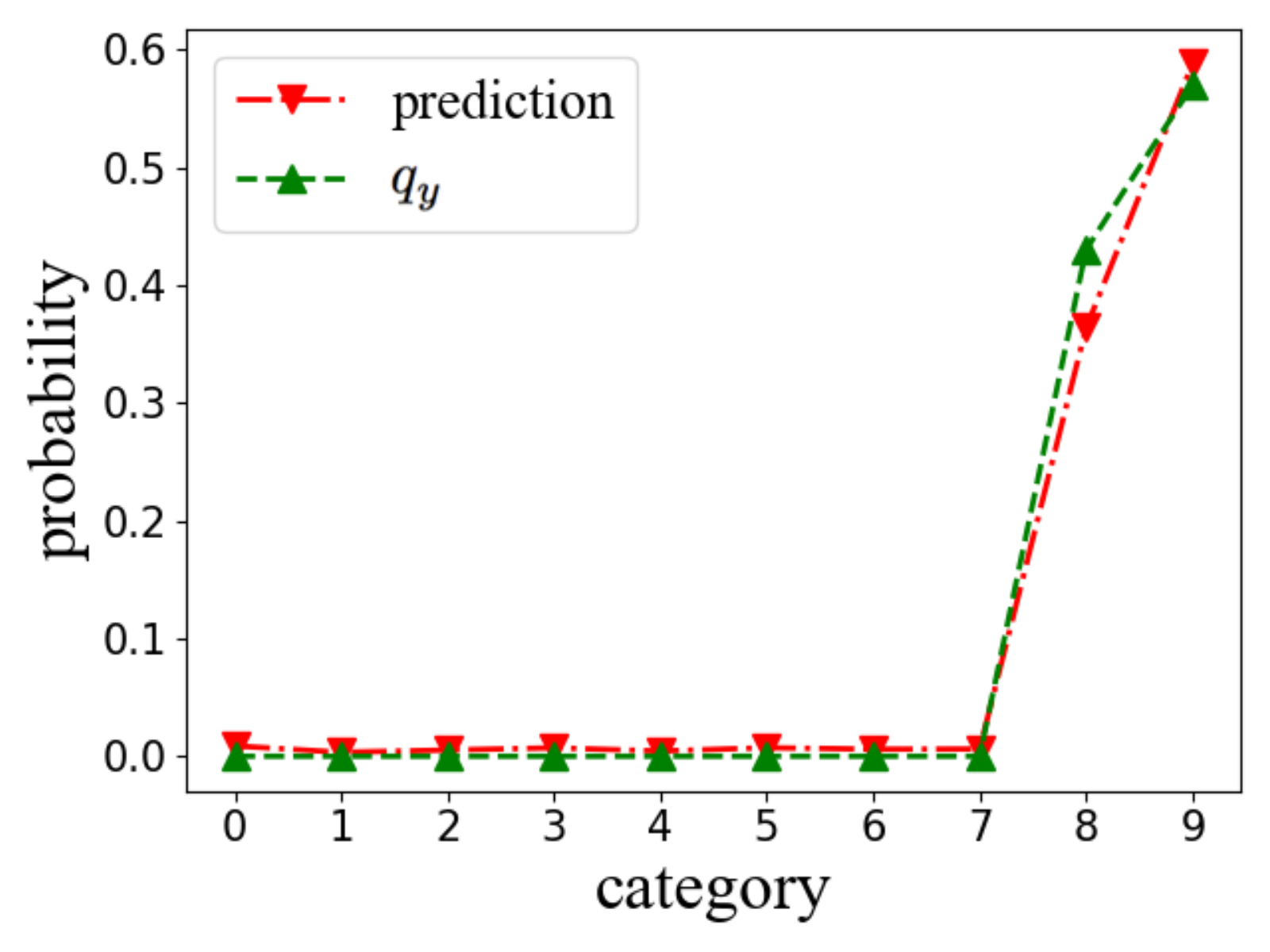}}}
    \subfigure[$\beta=0.4$, digit of $9$]{
    \centering{\includegraphics[width=0.22\columnwidth]{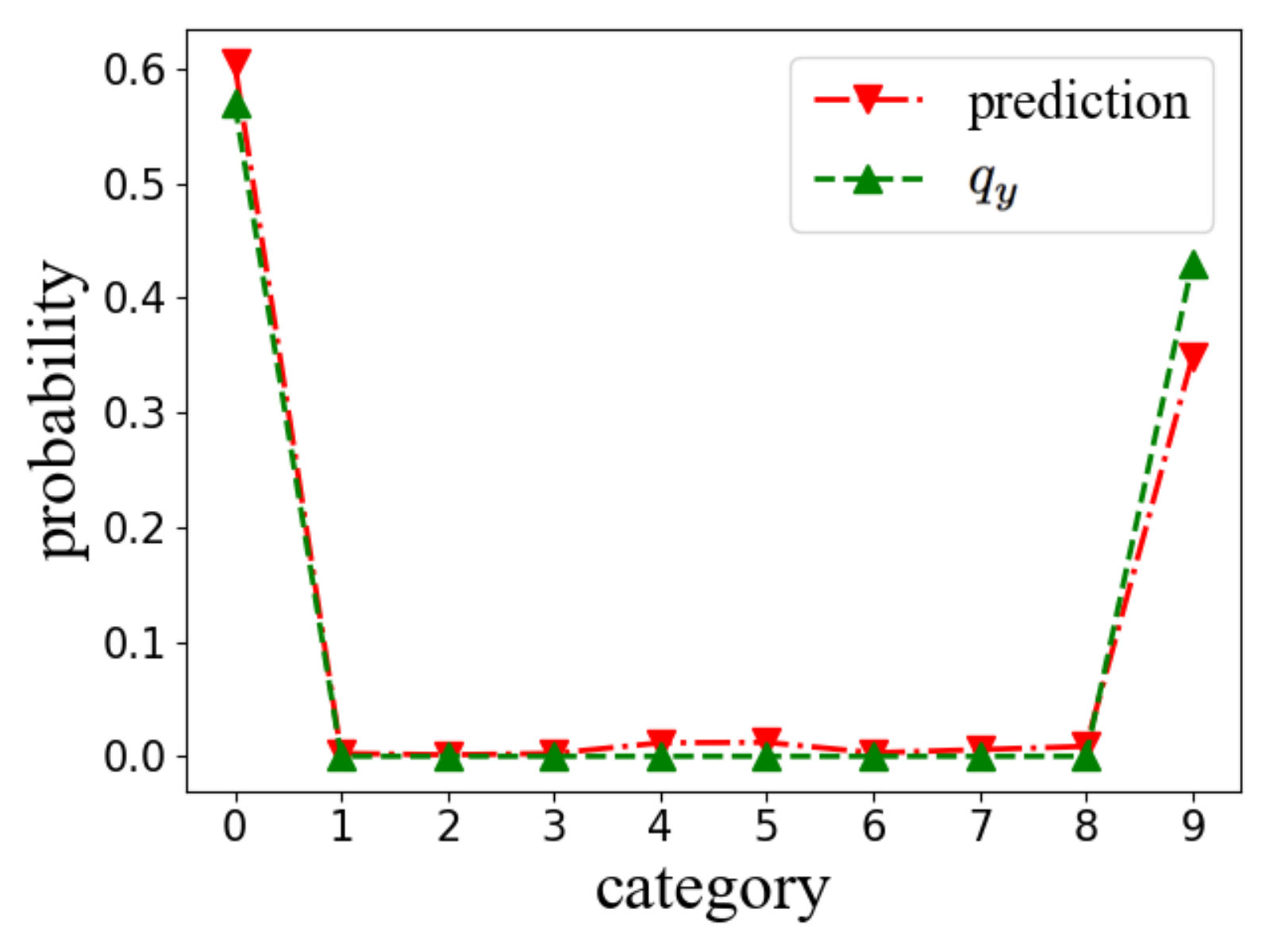}}}
    \subfigure[$\beta=0.4$, instance]{
    \centering{\includegraphics[width=0.22\columnwidth]{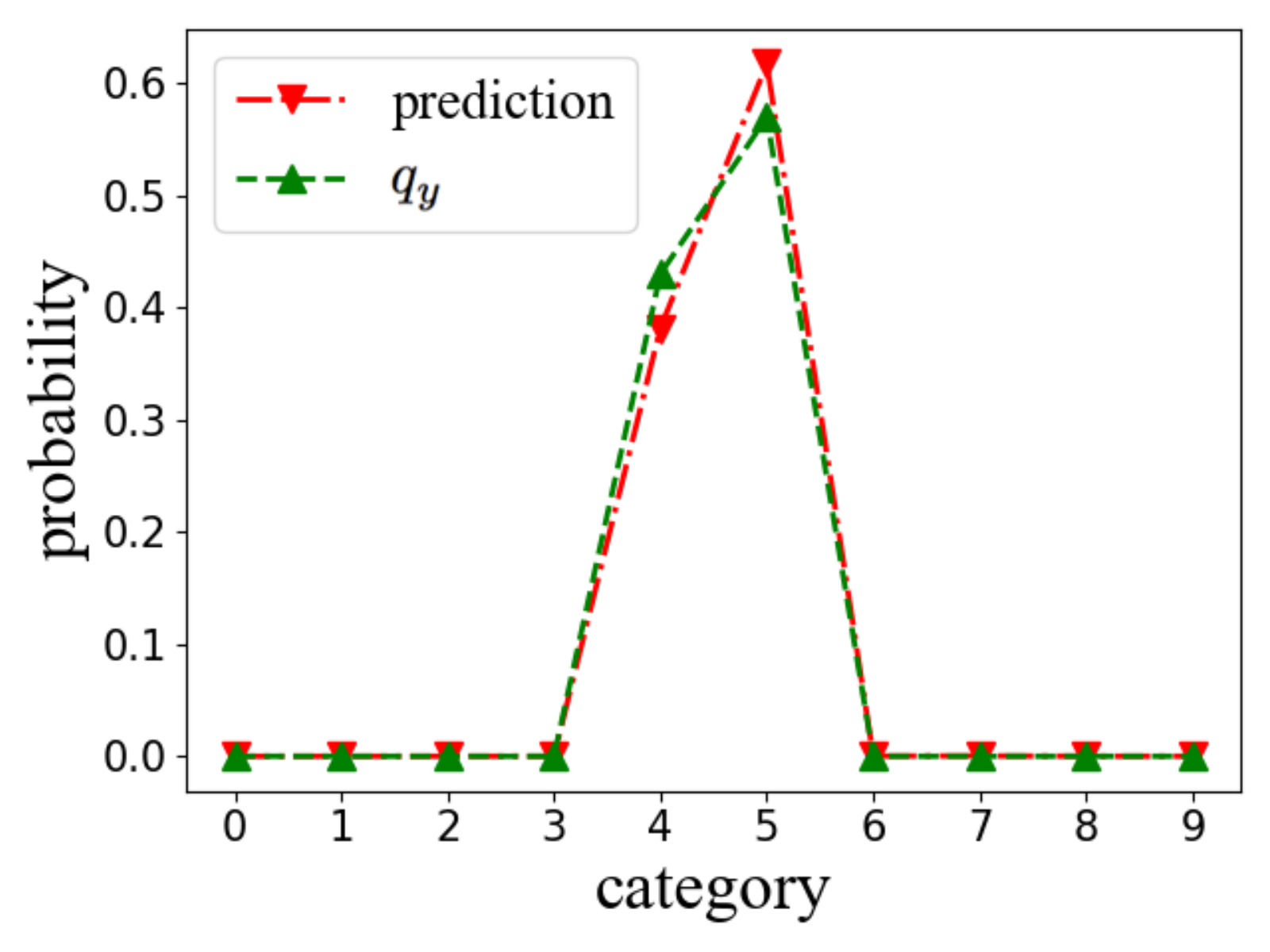}}}
    \subfigure[$\beta=0.4$, instance]{
    \centering{\includegraphics[width=0.22\columnwidth]{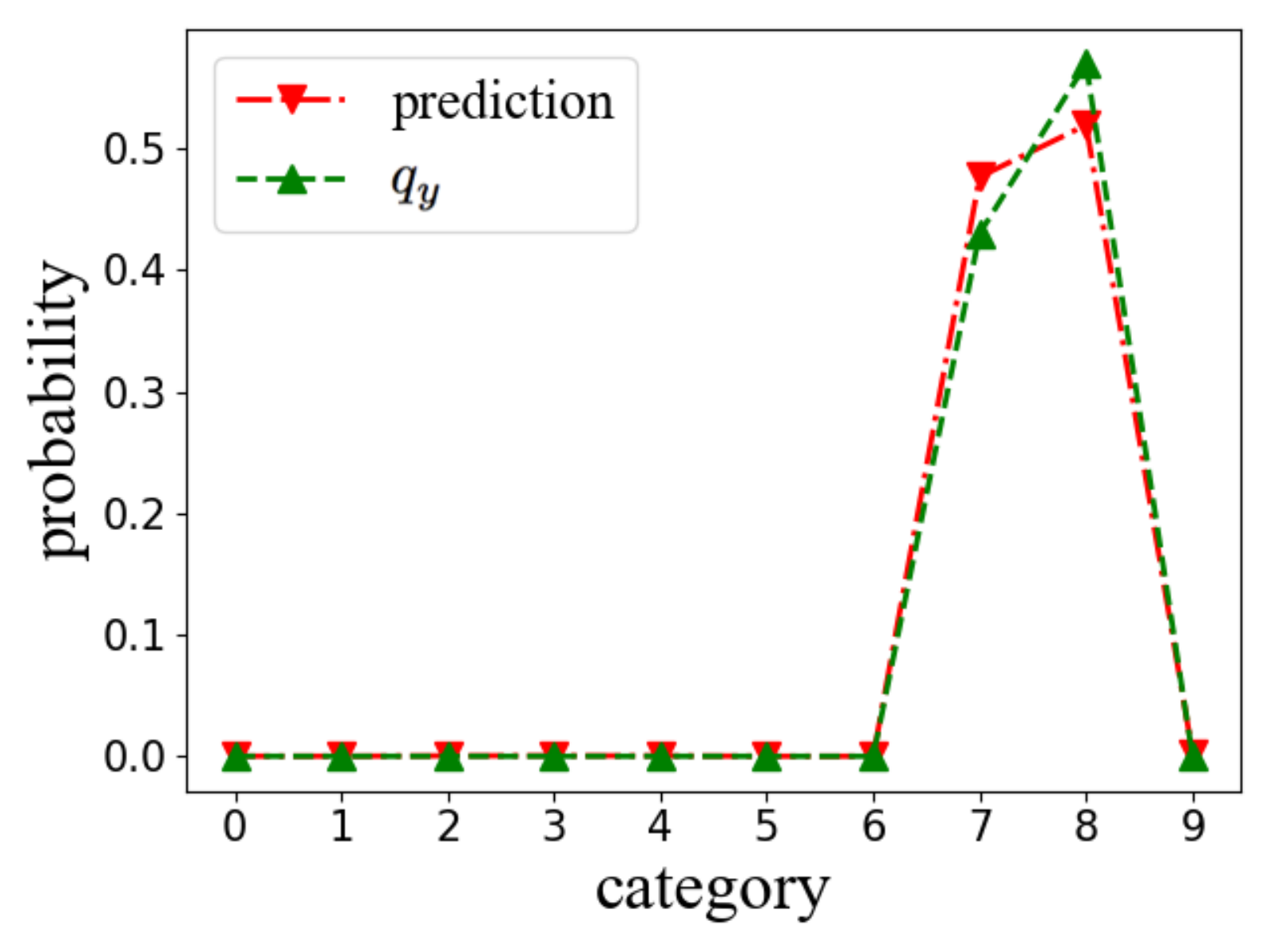}}}
    
    \subfigure[$\beta=0.8$, digit of $0$]{
    \centering{\includegraphics[width=0.22\columnwidth]{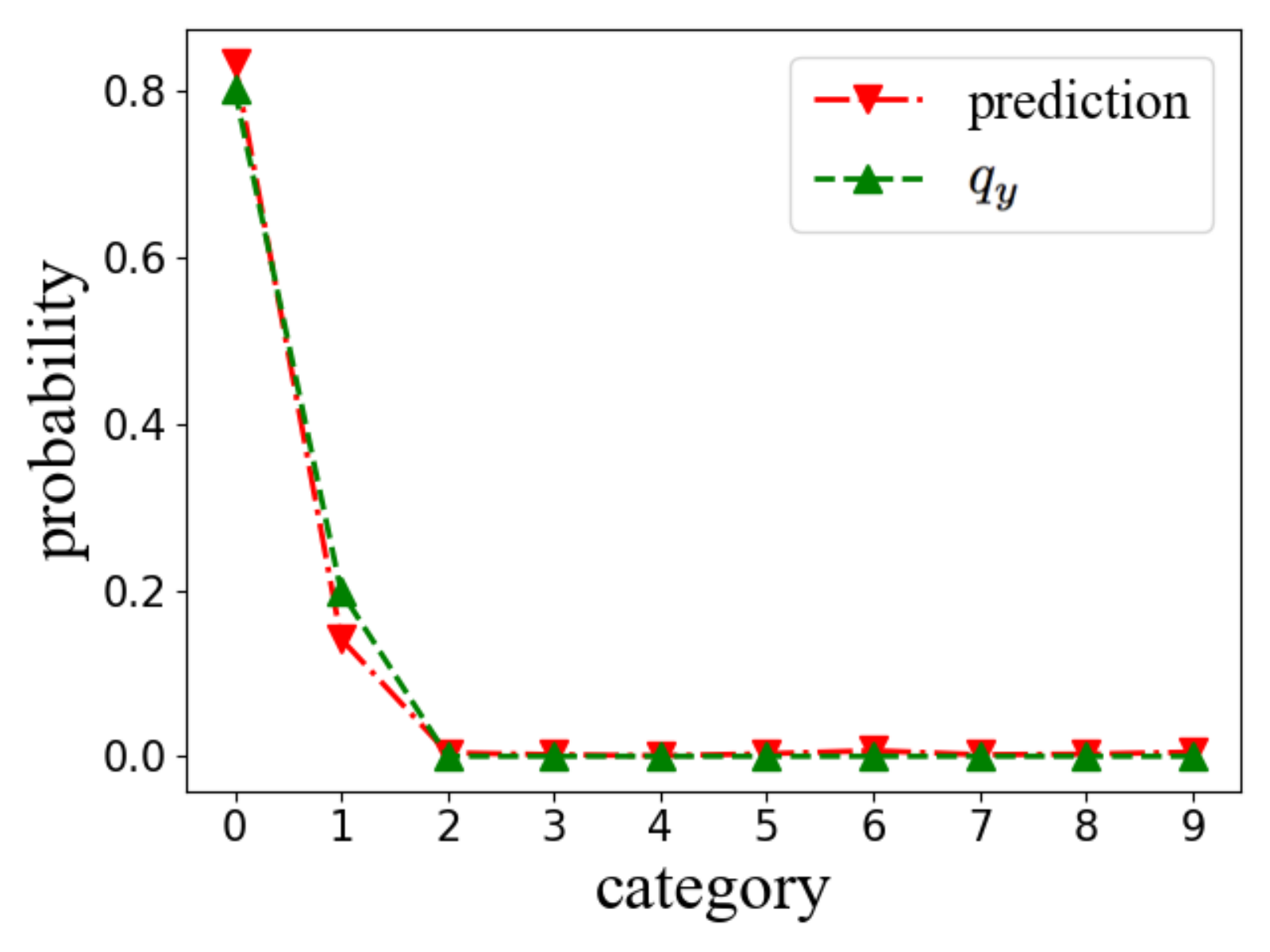}}}
    \subfigure[$\beta=0.8$, digit of $1$]{
    \centering{\includegraphics[width=0.22\columnwidth]{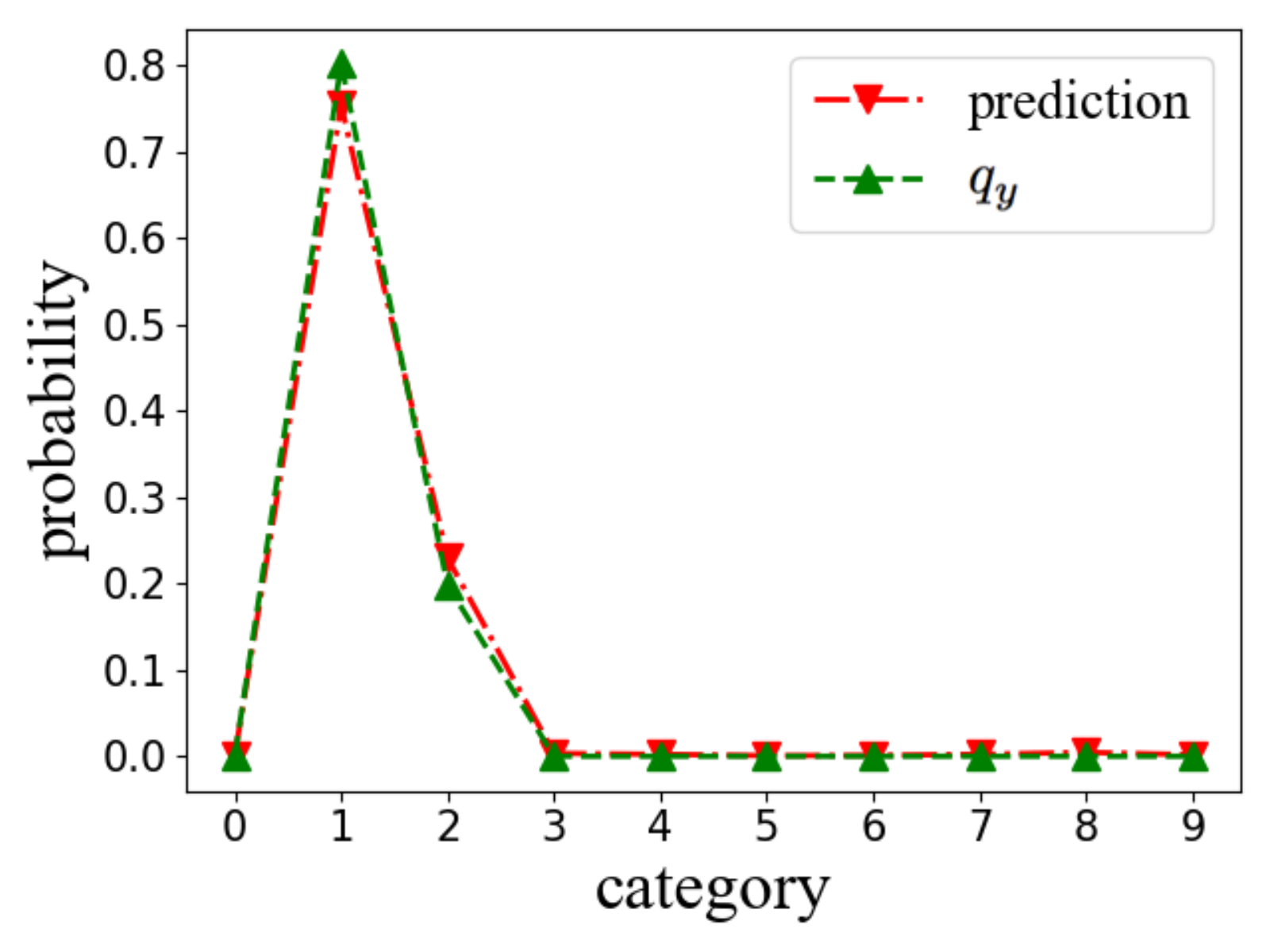}}}
    \subfigure[$\beta=0.8$, digit of $2$]{
    \centering{\includegraphics[width=0.22\columnwidth]{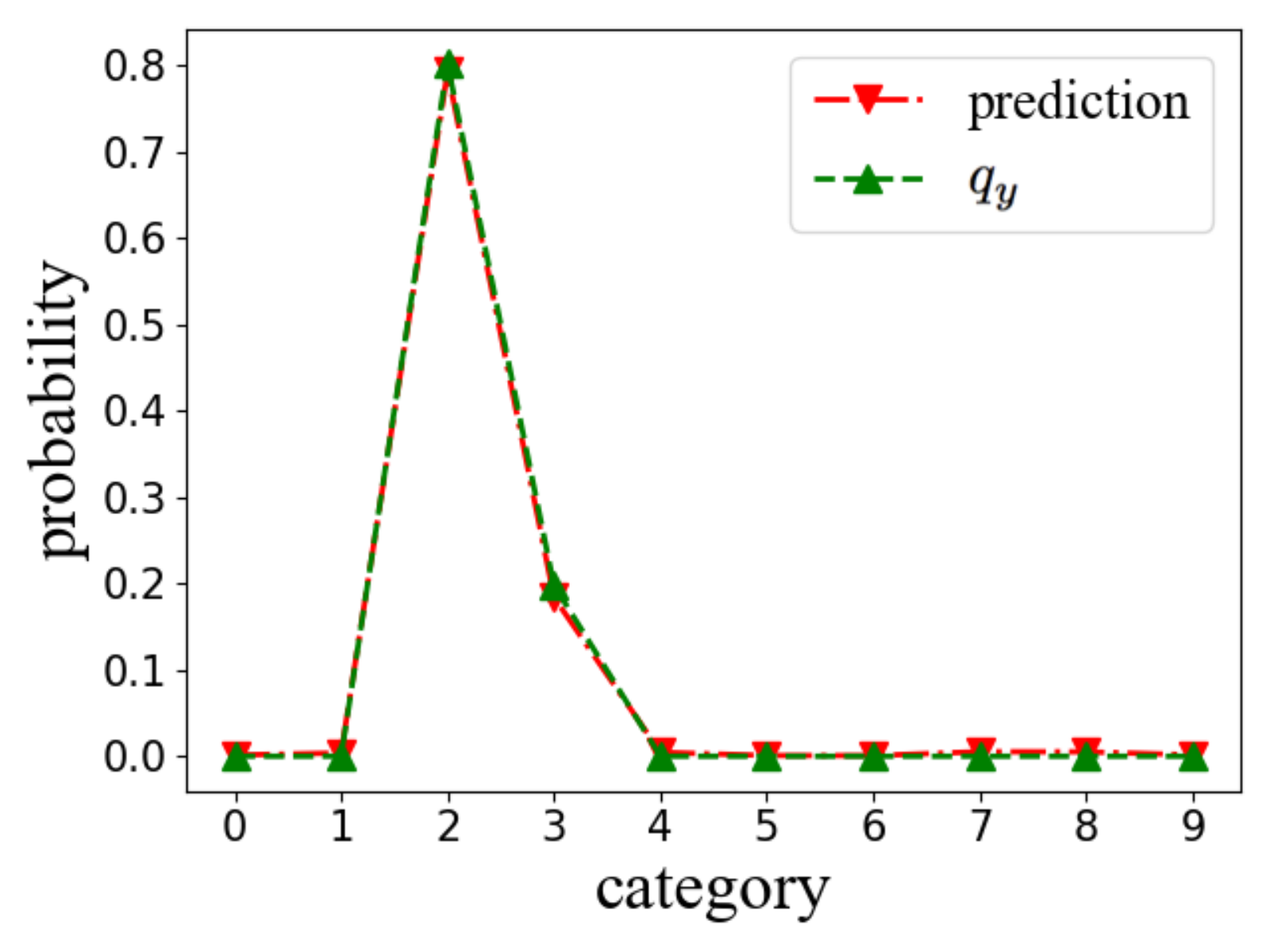}}}
    \subfigure[$\beta=0.8$, digit of $3$]{
    \centering{\includegraphics[width=0.22\columnwidth]{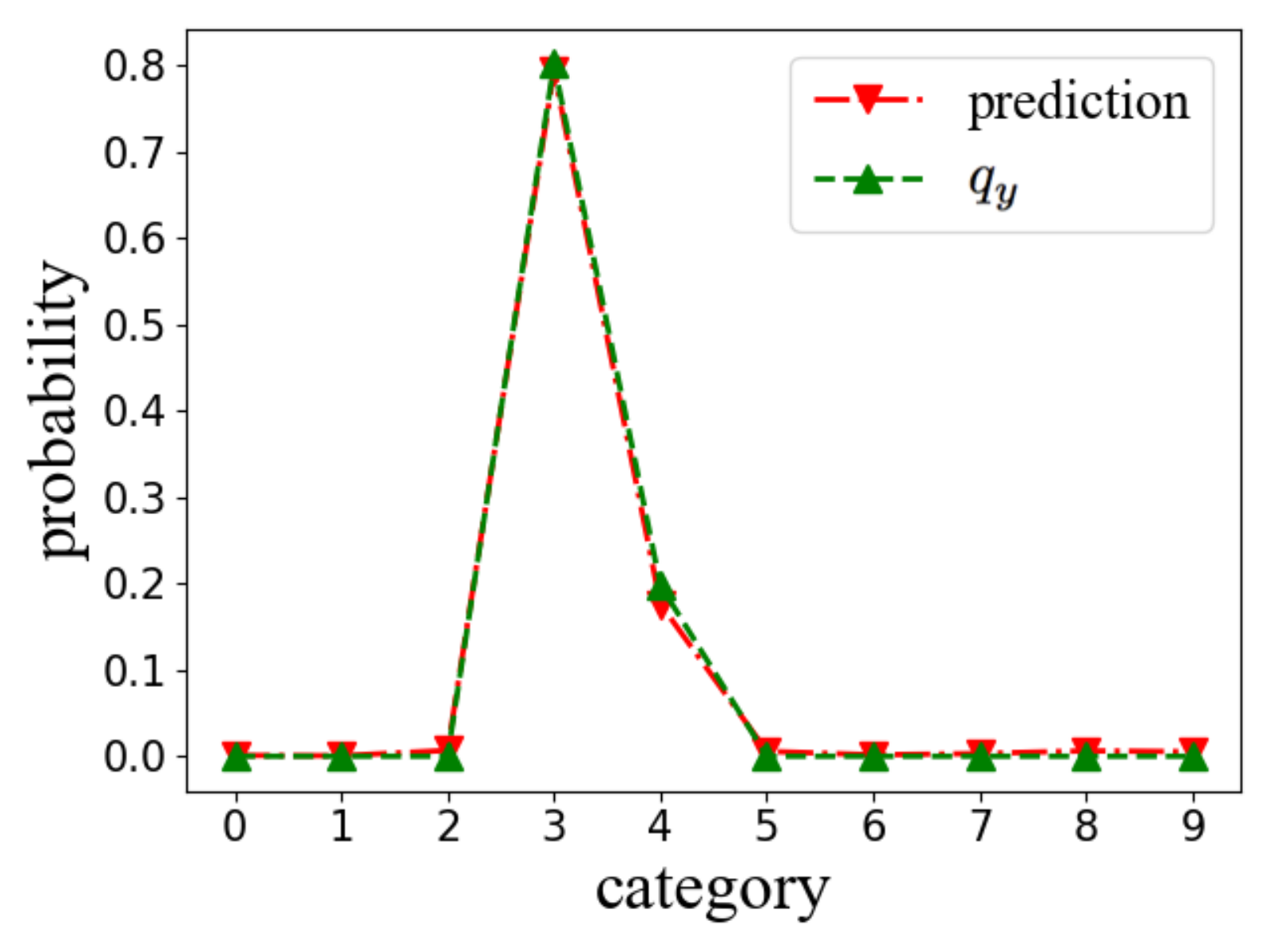}}}
    
    \subfigure[$\beta=0.8$, digit of $4$]{
    \centering{\includegraphics[width=0.22\columnwidth]{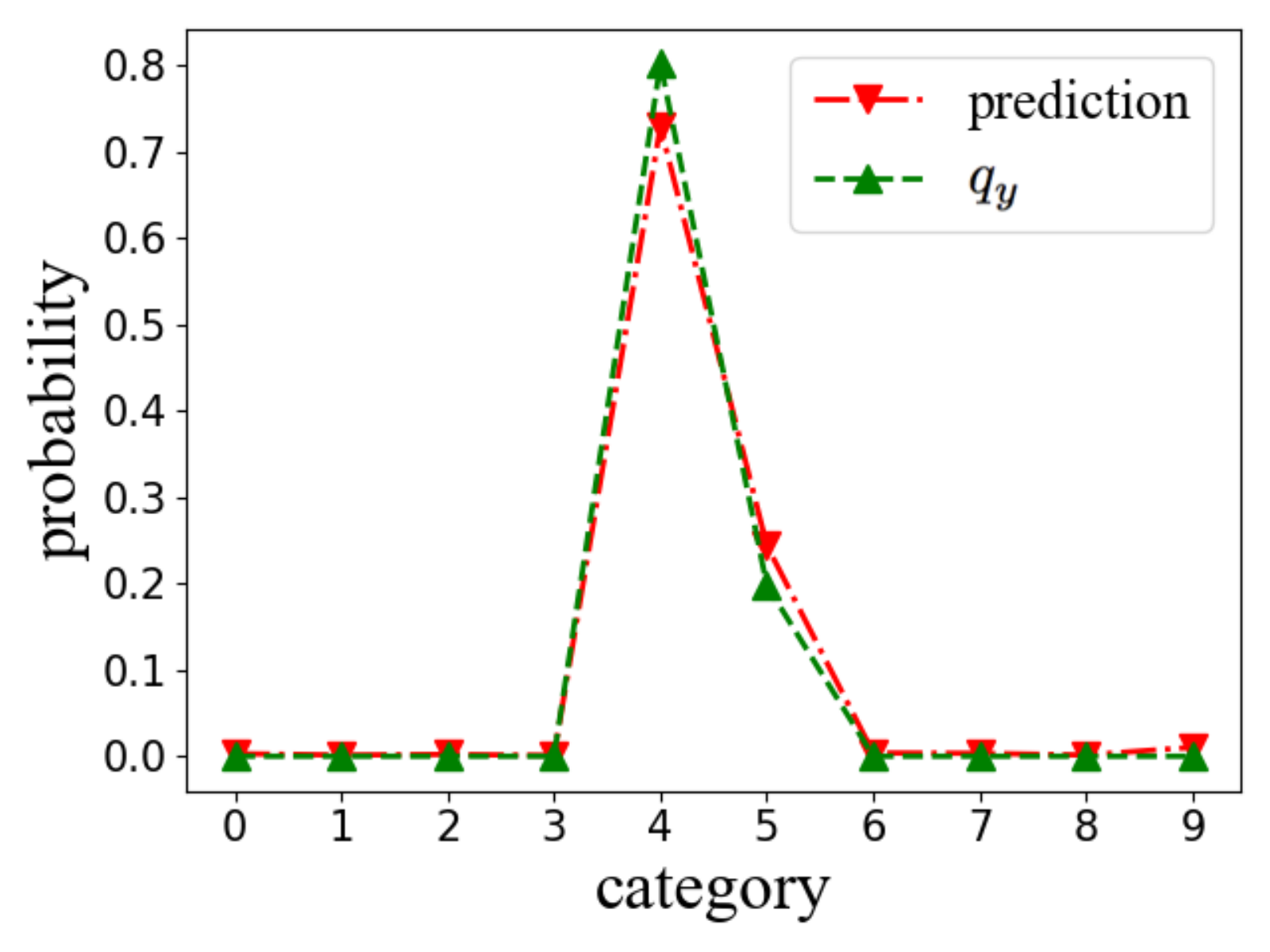}}}
    \subfigure[$\beta=0.8$, digit of $5$]{
    \centering{\includegraphics[width=0.22\columnwidth]{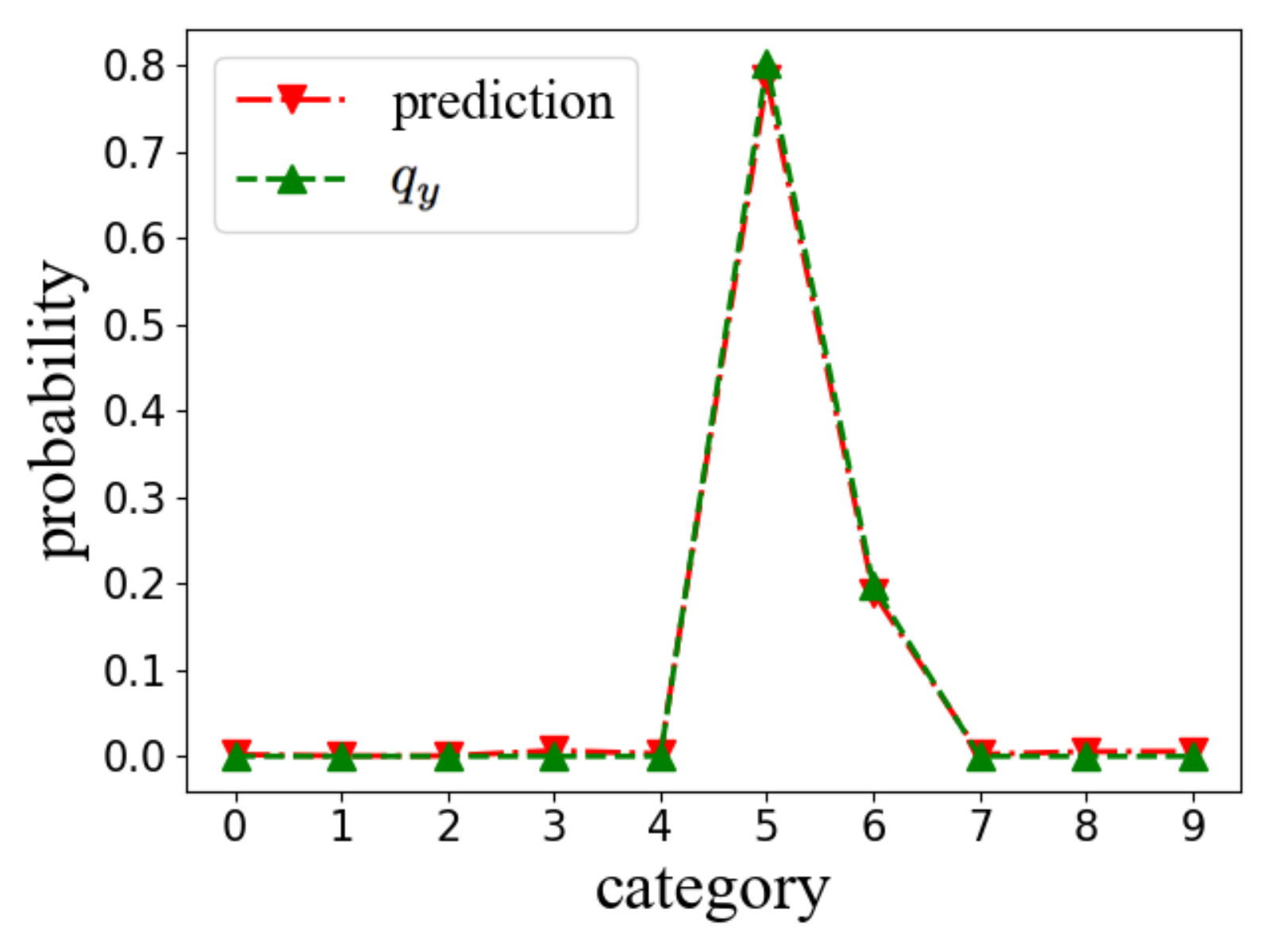}}}
    \subfigure[$\beta=0.8$, digit of $6$]{
    \centering{\includegraphics[width=0.22\columnwidth]{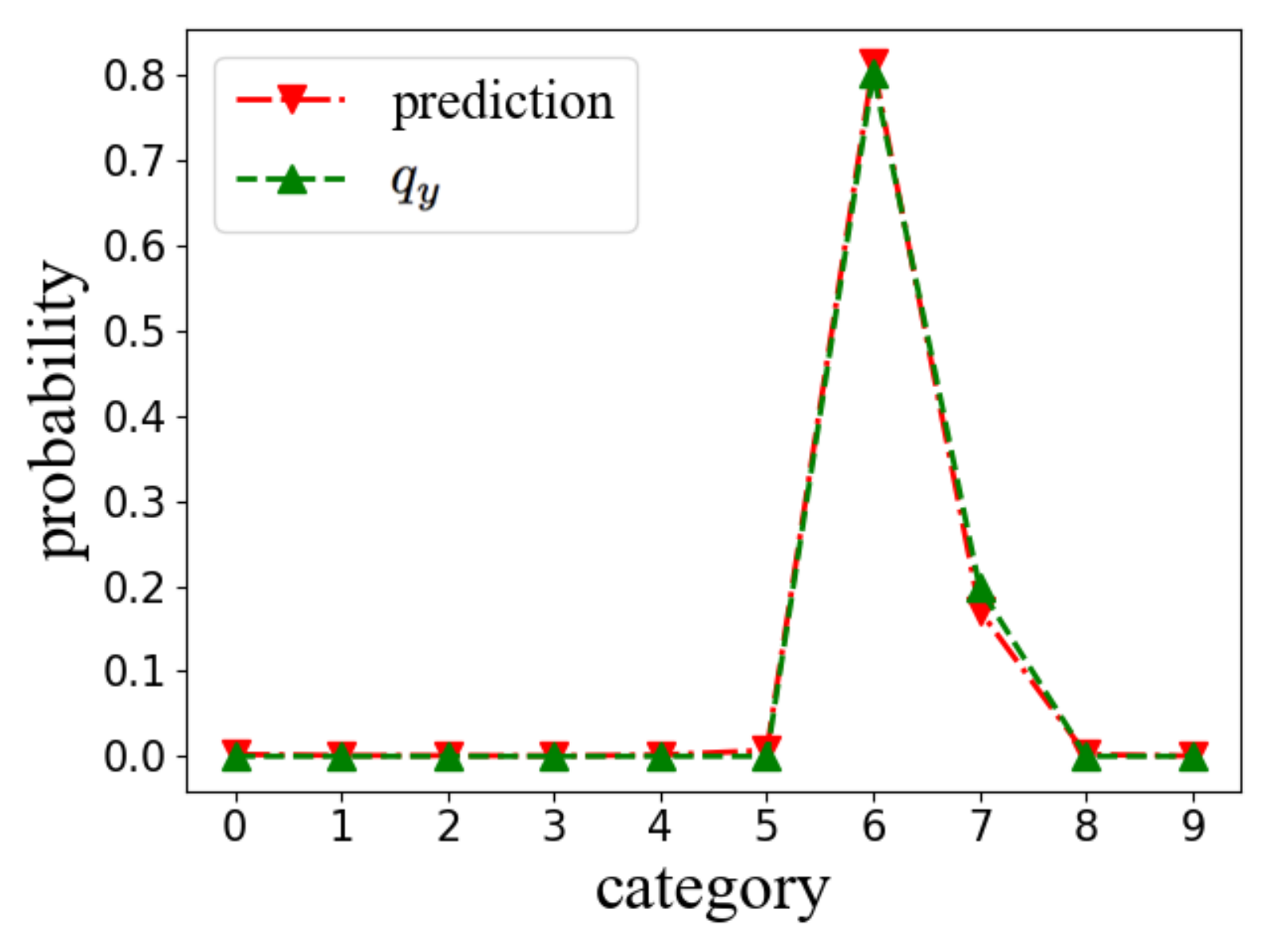}}}
    \subfigure[$\beta=0.8$, digit of $7$]{
    \centering{\includegraphics[width=0.22\columnwidth]{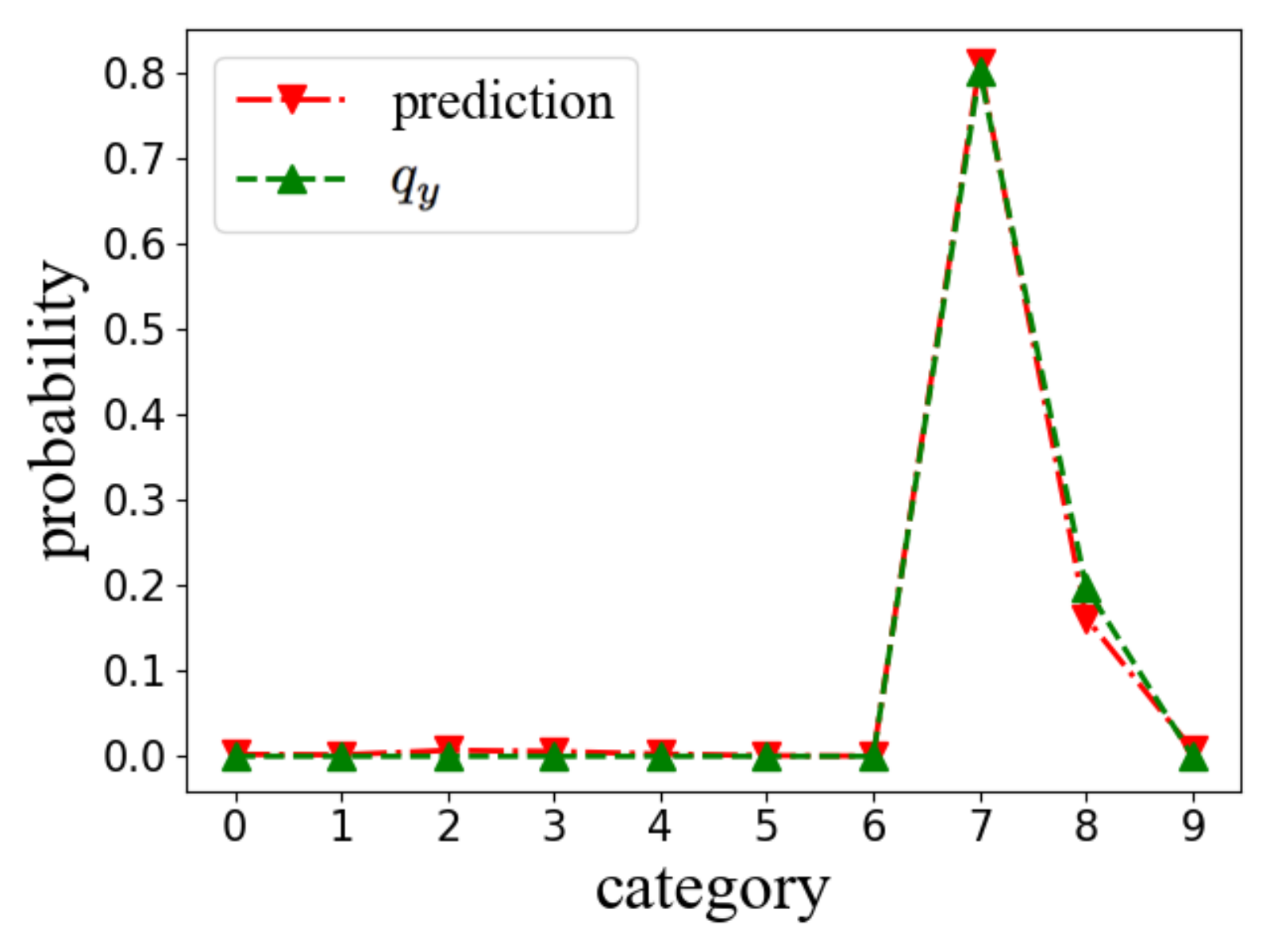}}}
    
    \subfigure[$\beta=0.8$, digit of $8$]{
    \centering{\includegraphics[width=0.22\columnwidth]{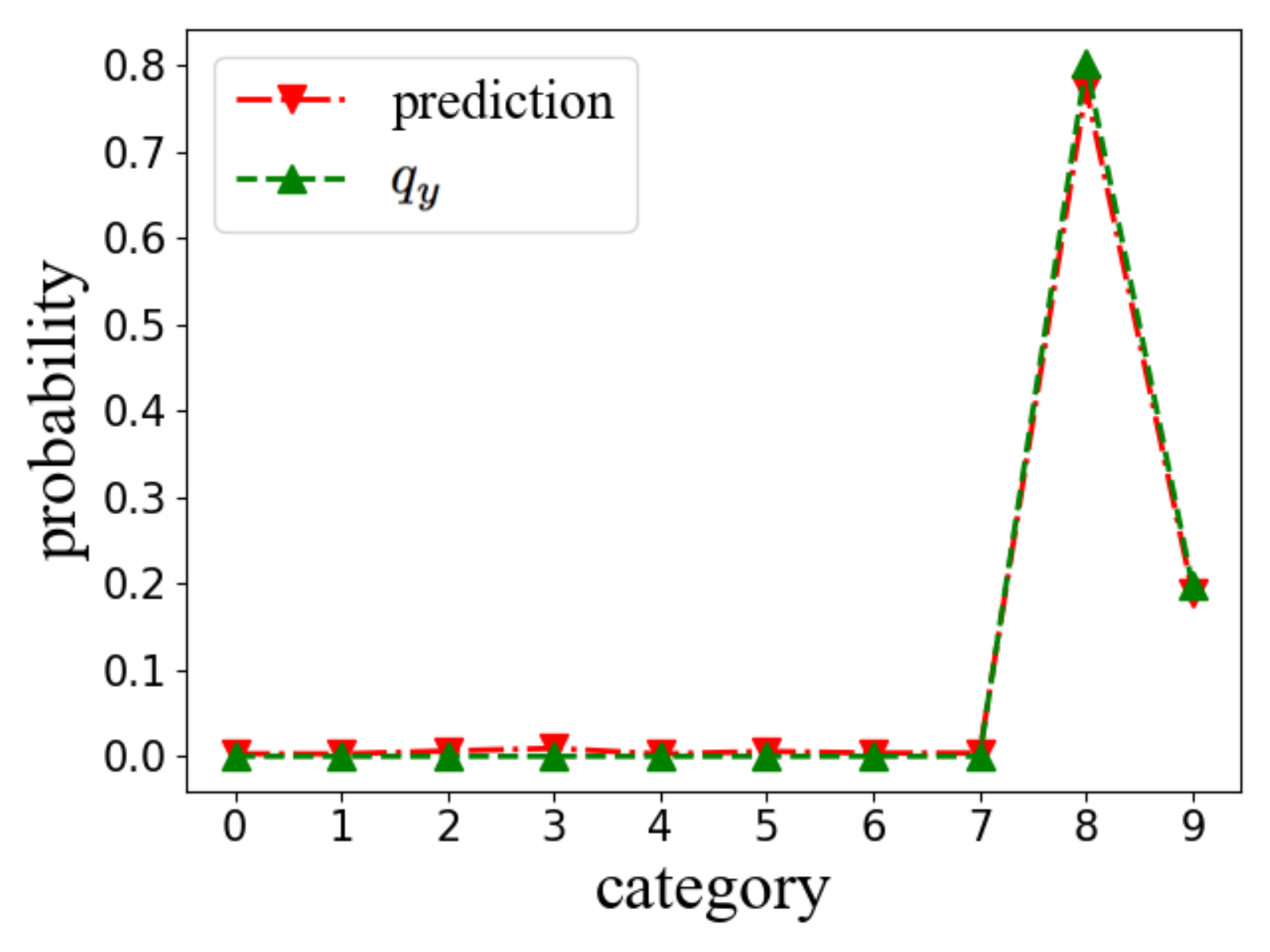}}}
    \subfigure[$\beta=0.8$, digit of $9$]{
    \centering{\includegraphics[width=0.22\columnwidth]{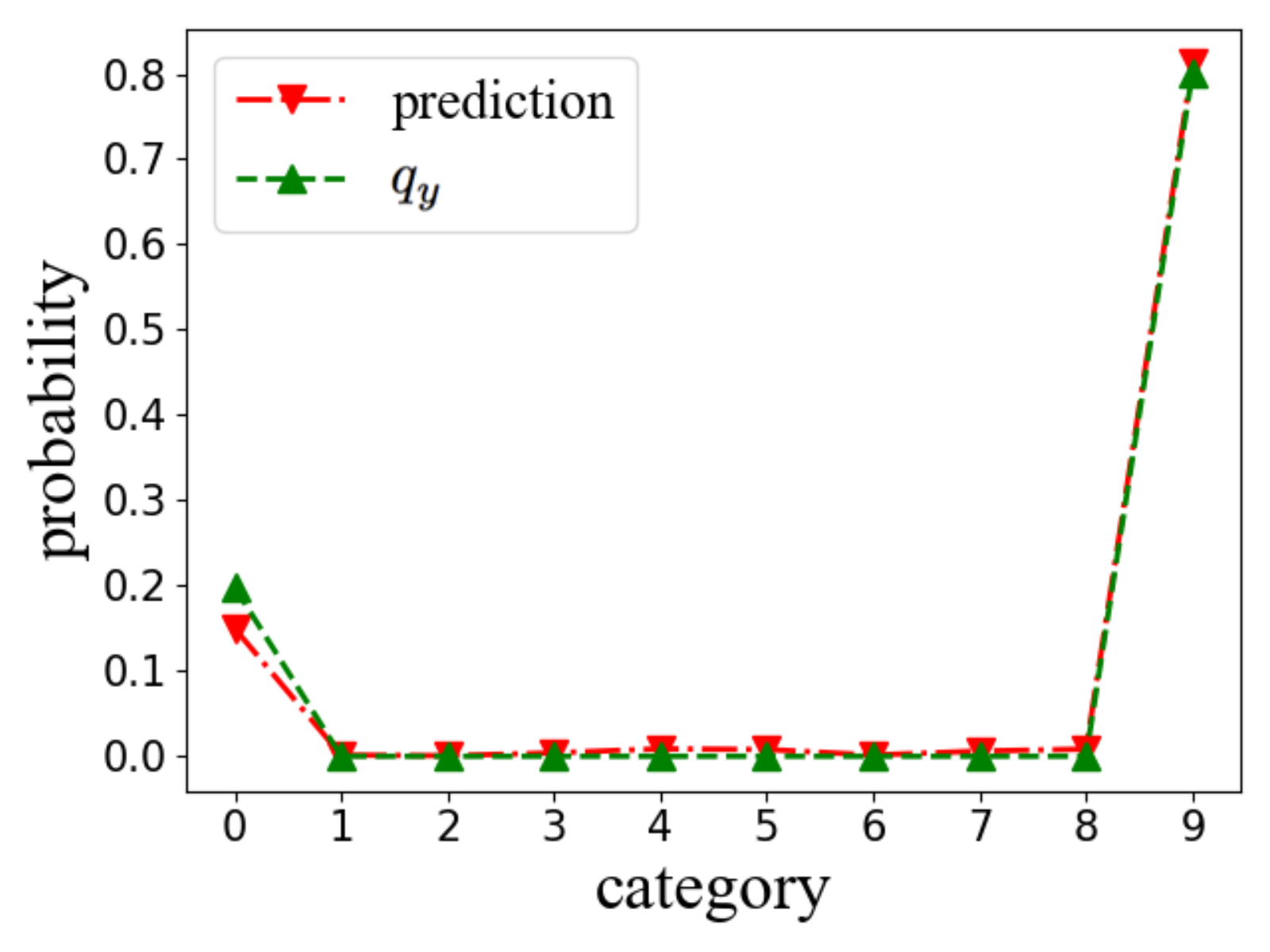}}}
    \subfigure[$\beta=0.8$, instance]{
    \centering{\includegraphics[width=0.22\columnwidth]{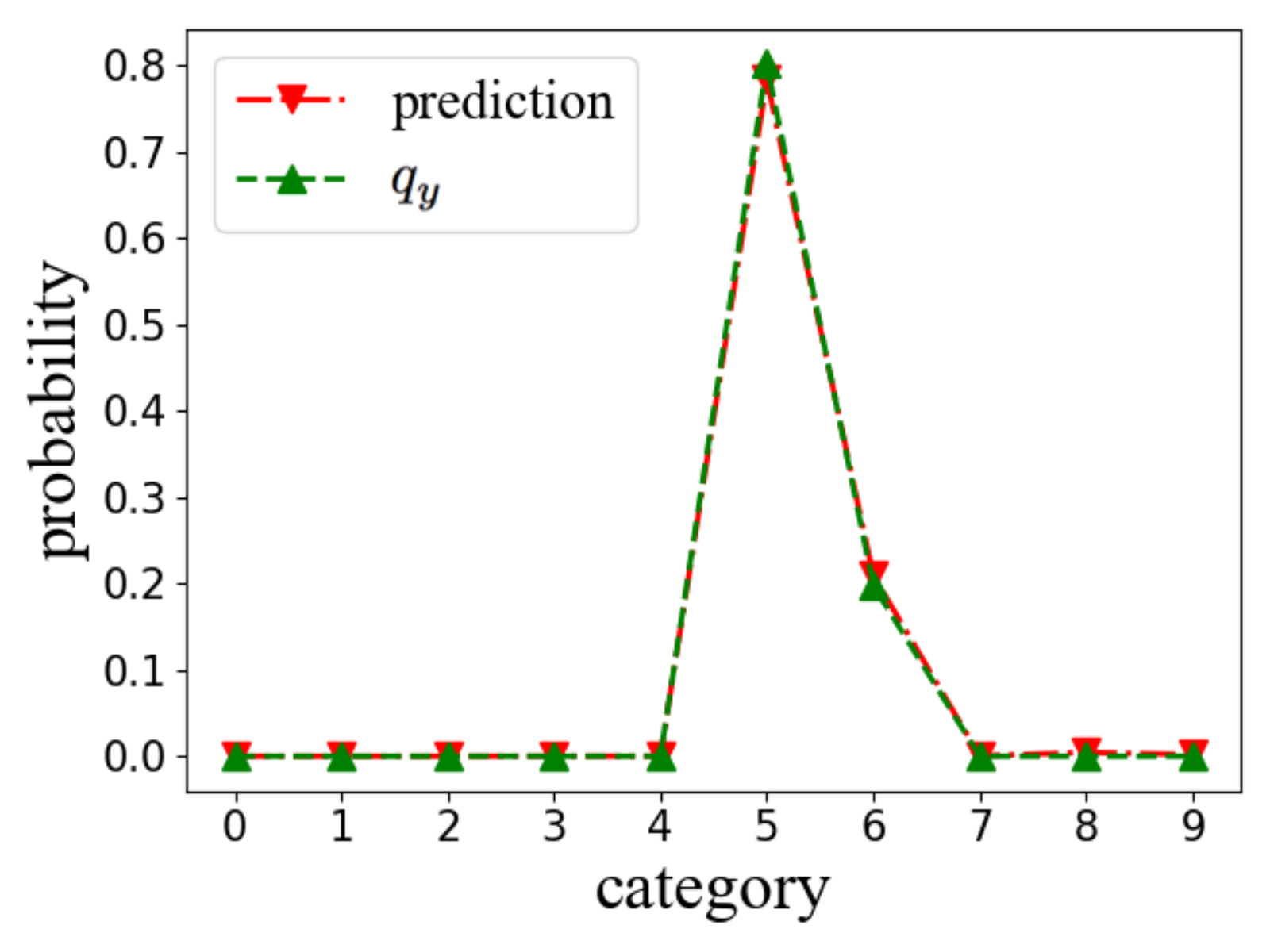}}}
    \subfigure[$\beta=0.8$, instance]{
    \centering{\includegraphics[width=0.22\columnwidth]{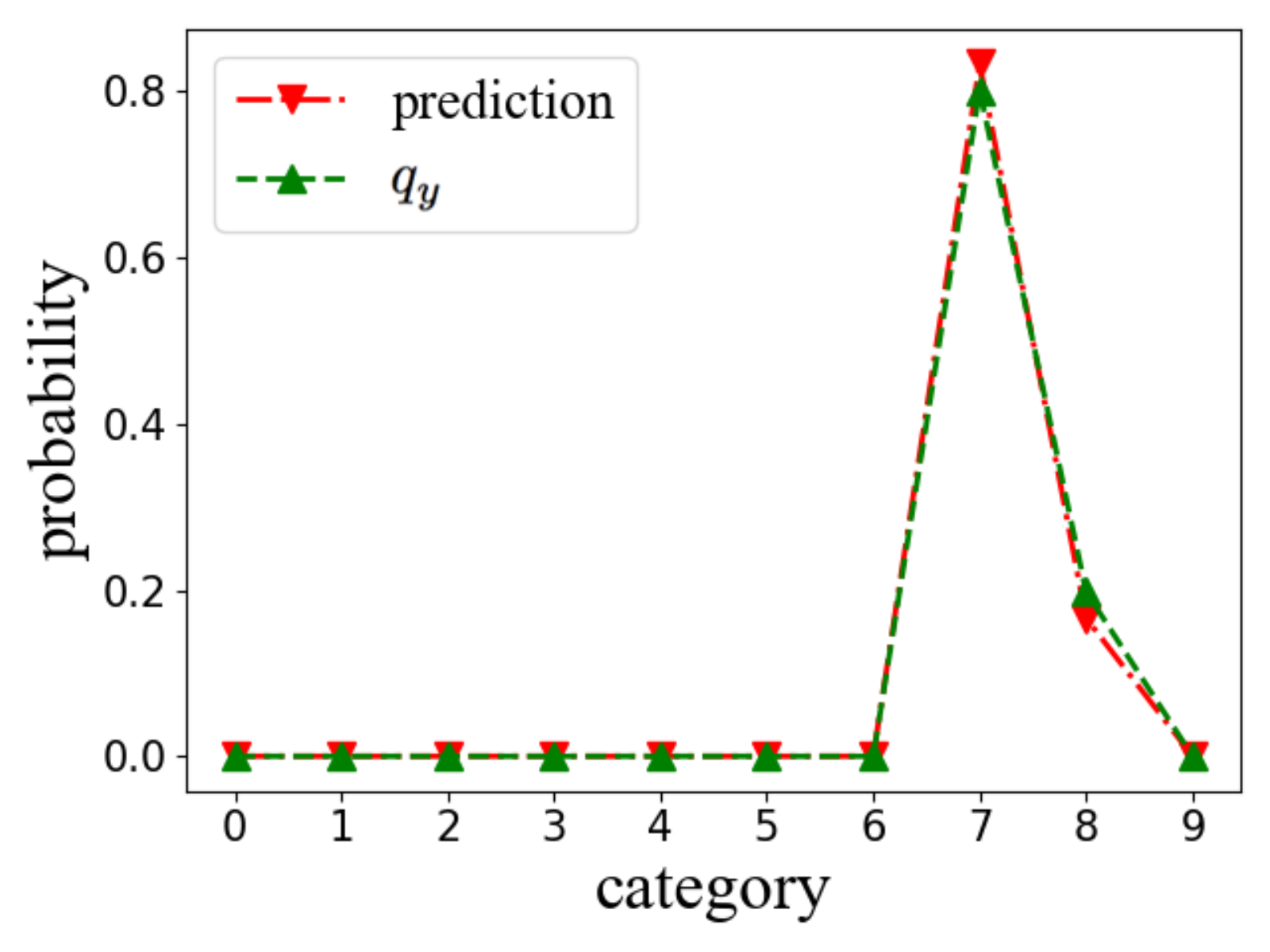}}}

\caption{Biased noise when $\beta=0.4$ and $\beta=0.8$. (k), (l), (w) and (x): instances of digit $5$. Others are the average of predicted probabilities and $q_y$ for all images of different digits. The red lines represent $f(y; \theta)$ whereas the green lines represent ${q}_{y}$.}
\label{fig:biased_noise}
\end{center}
\end{figure}

\begin{figure}[t]
\begin{center}
    \subfigure[$\beta=0.4$, digit of $0$]{
    \centering{\includegraphics[width=0.22\columnwidth]{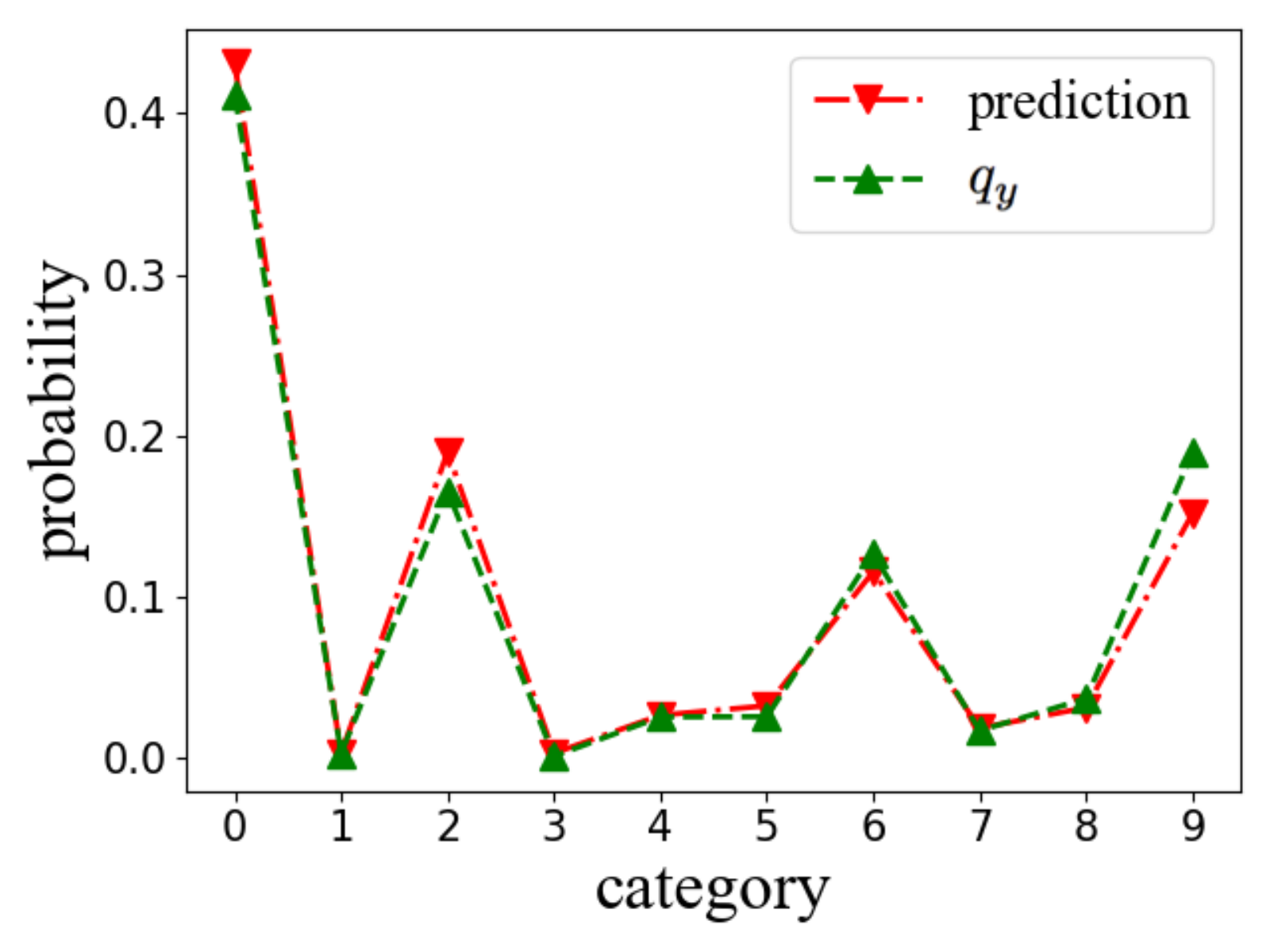}}}
    \subfigure[$\beta=0.4$, digit of $1$]{
    \centering{\includegraphics[width=0.22\columnwidth]{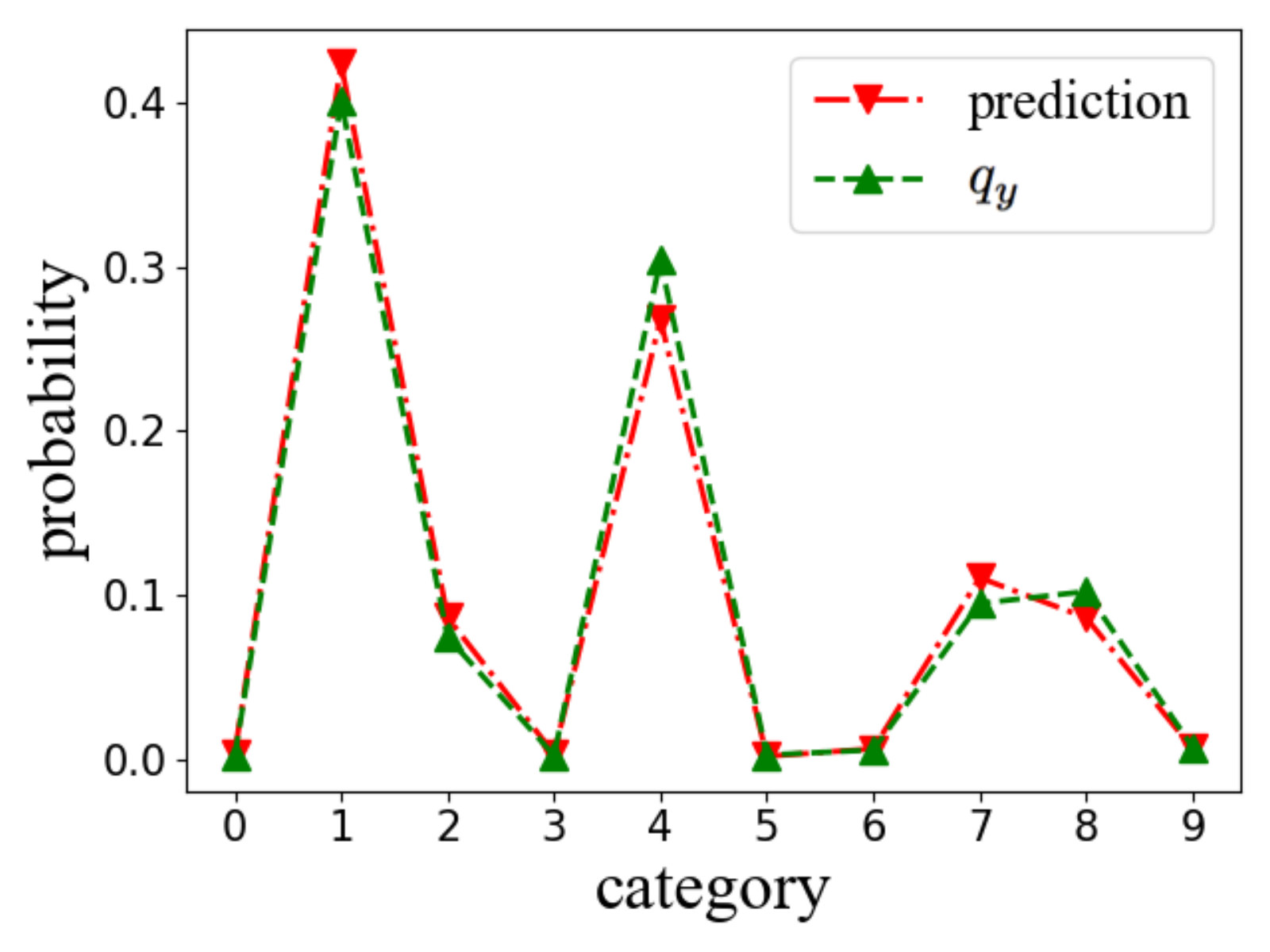}}}
    \subfigure[$\beta=0.4$, digit of $2$]{
    \centering{\includegraphics[width=0.22\columnwidth]{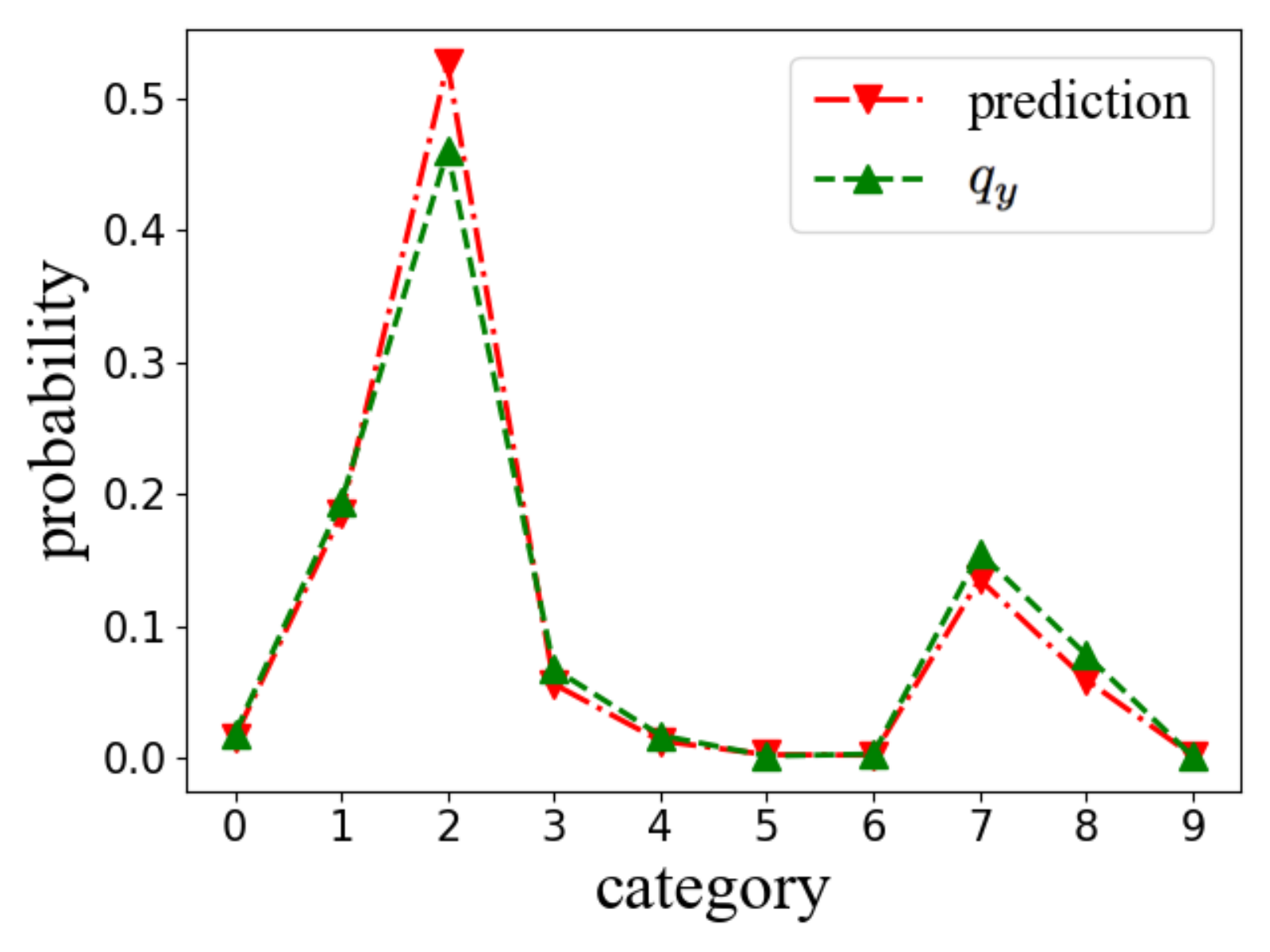}}}
    \subfigure[$\beta=0.4$, digit of $3$]{
    \centering{\includegraphics[width=0.22\columnwidth]{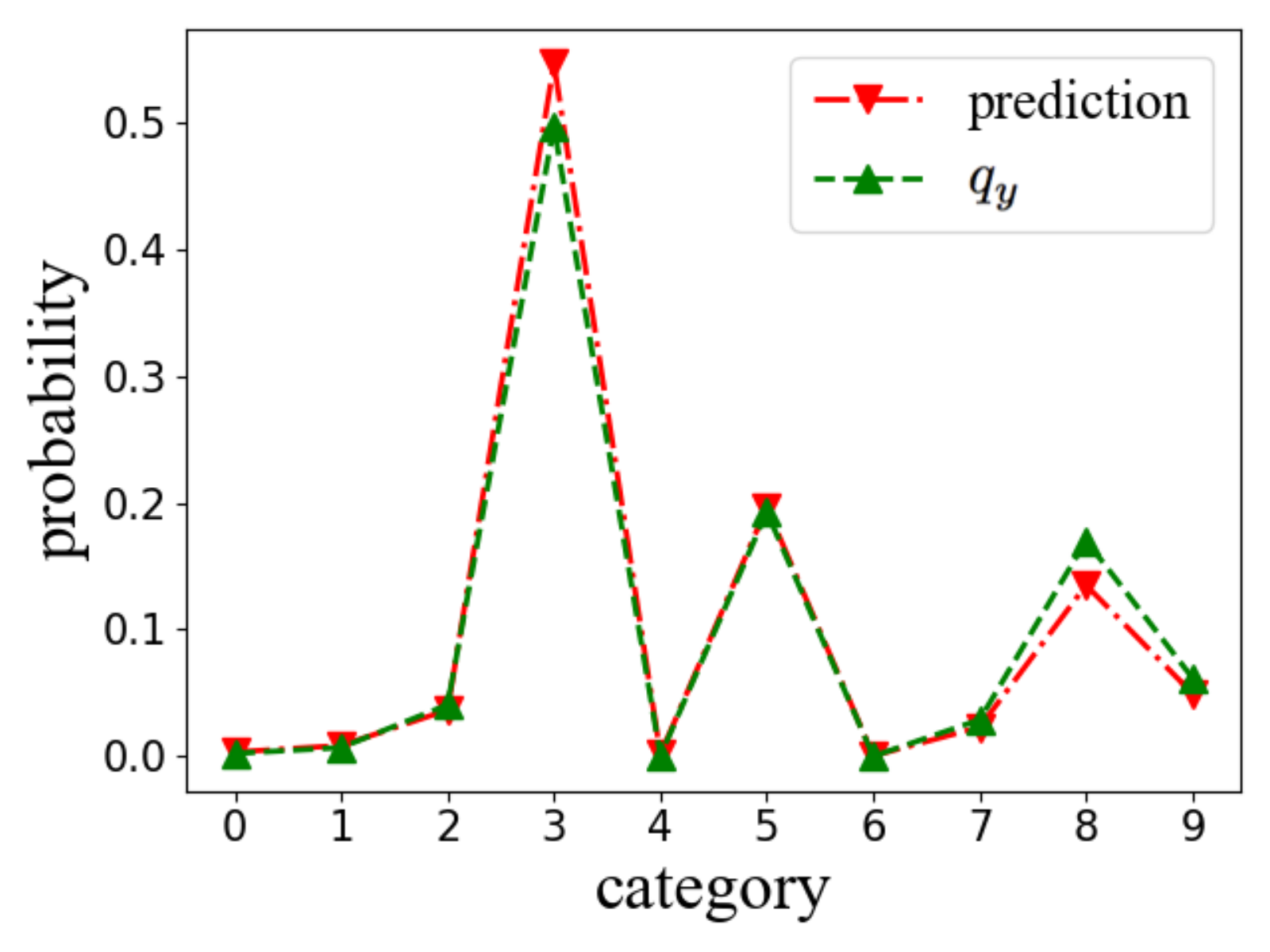}}}
    
    \subfigure[$\beta=0.4$, digit of $4$]{
    \centering{\includegraphics[width=0.22\columnwidth]{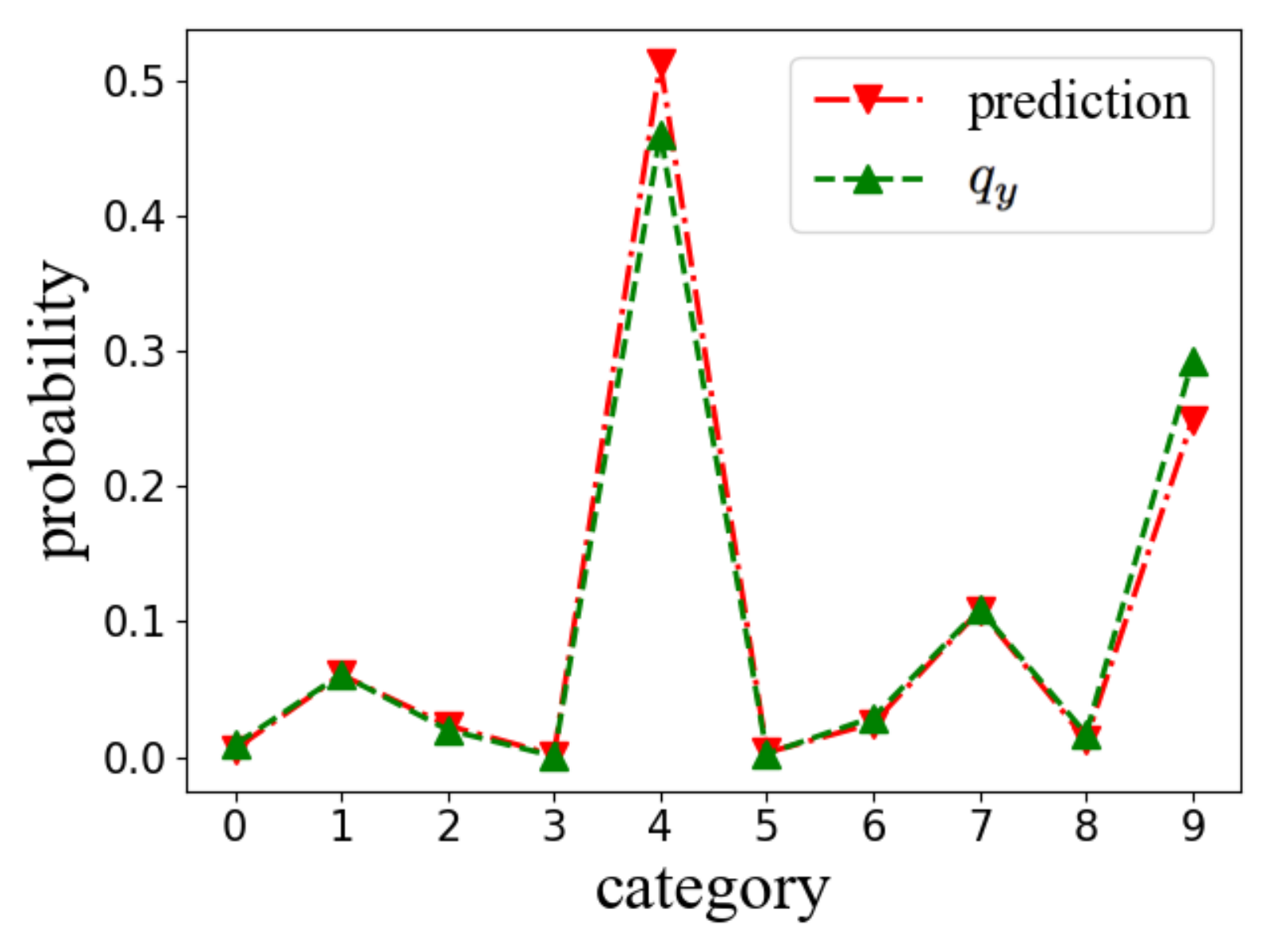}}}
    \subfigure[$\beta=0.4$, digit of $5$]{
    \centering{\includegraphics[width=0.22\columnwidth]{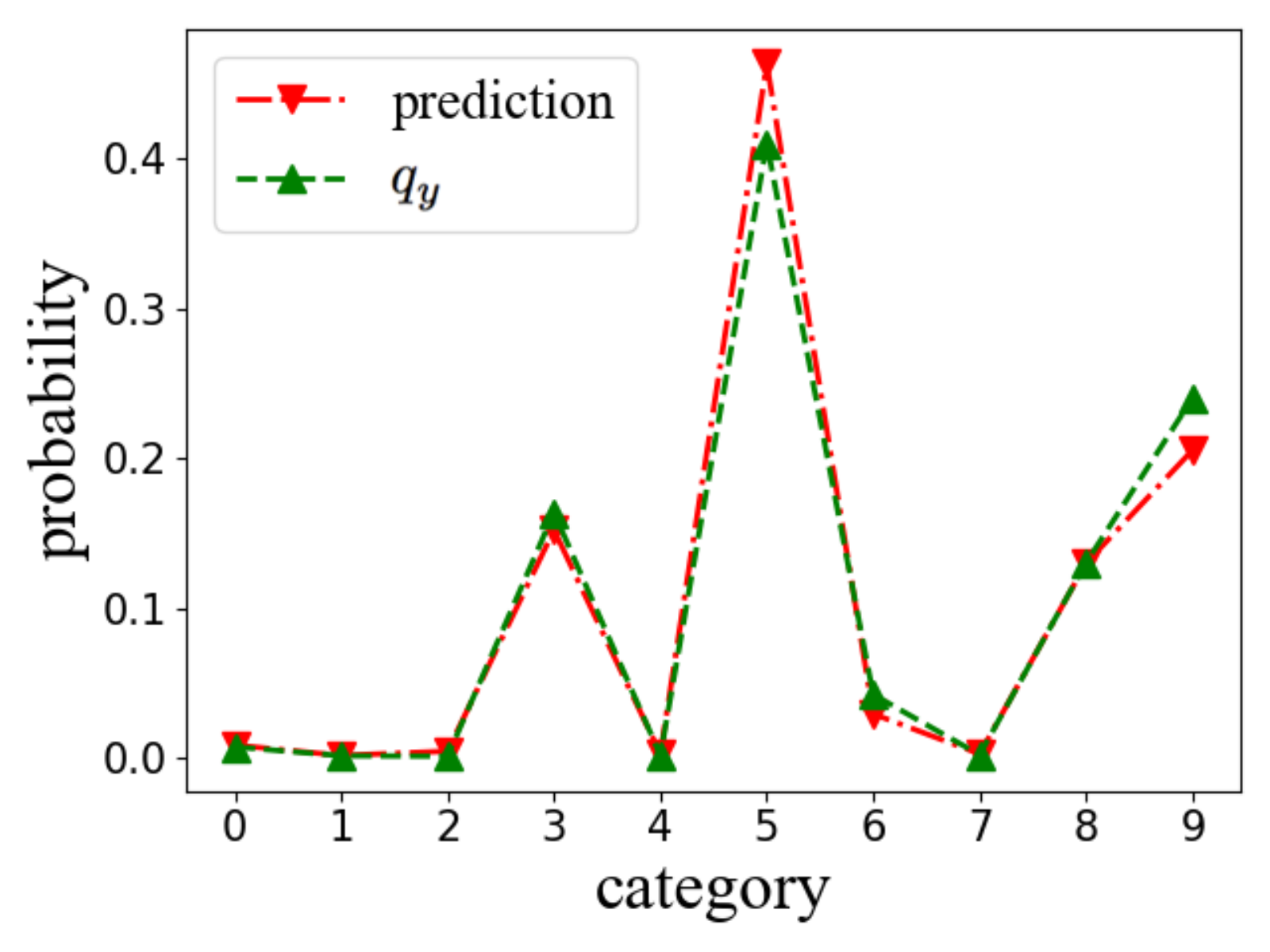}}}
    \subfigure[$\beta=0.4$, digit of $6$]{
    \centering{\includegraphics[width=0.22\columnwidth]{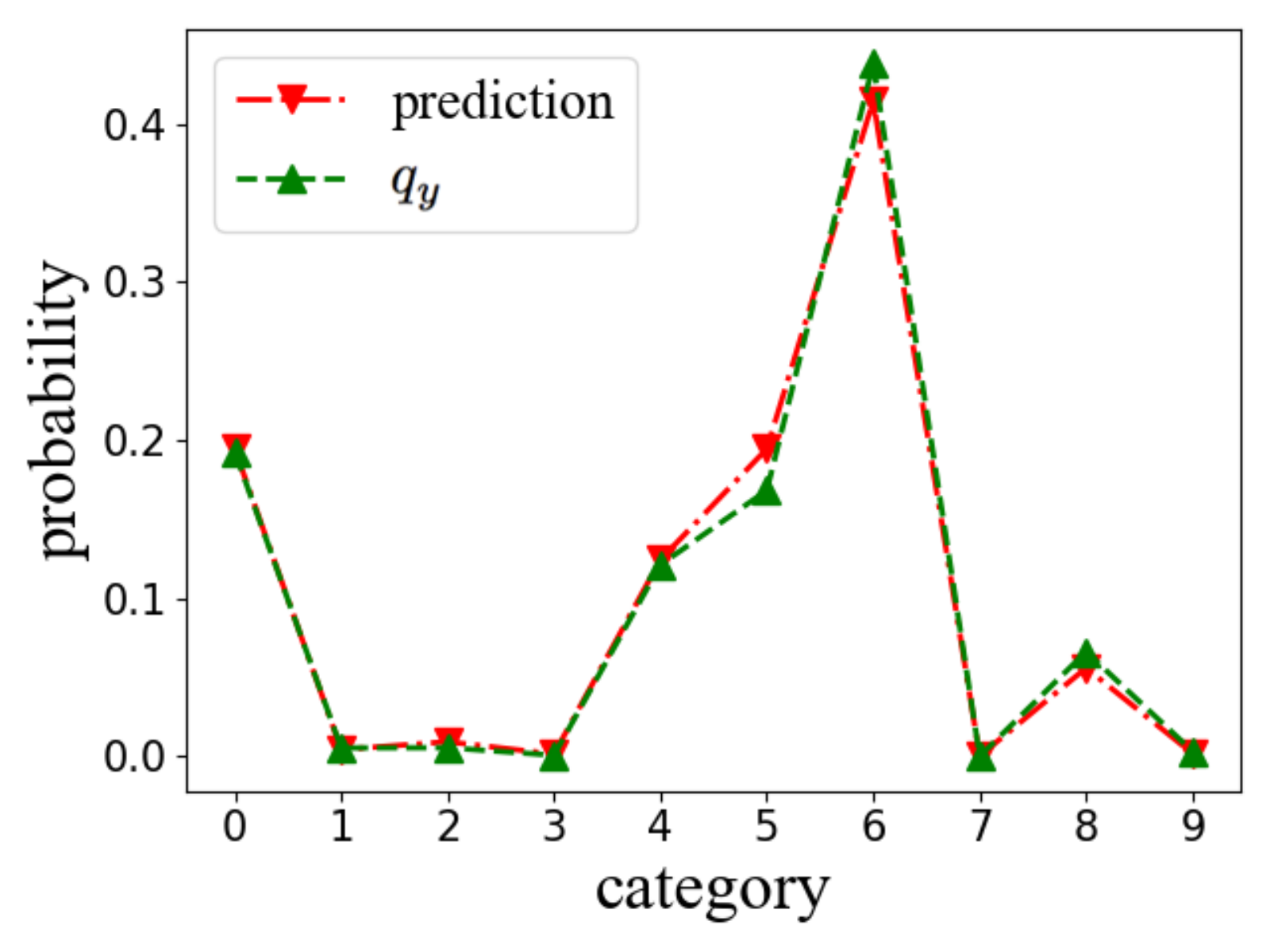}}}
    \subfigure[$\beta=0.4$, digit of $7$]{
    \centering{\includegraphics[width=0.22\columnwidth]{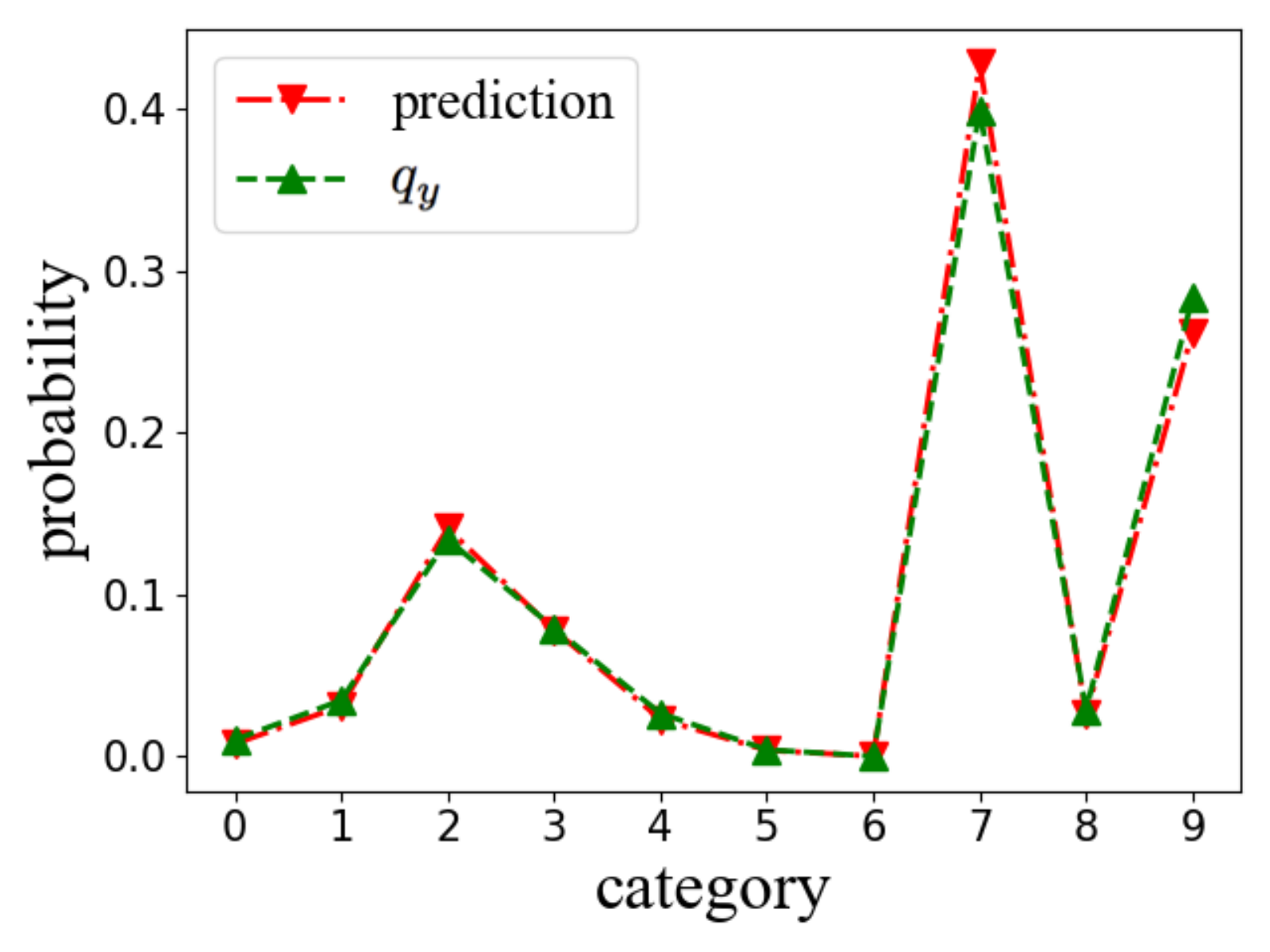}}}
    
    \subfigure[$\beta=0.4$, digit of $8$]{
    \centering{\includegraphics[width=0.22\columnwidth]{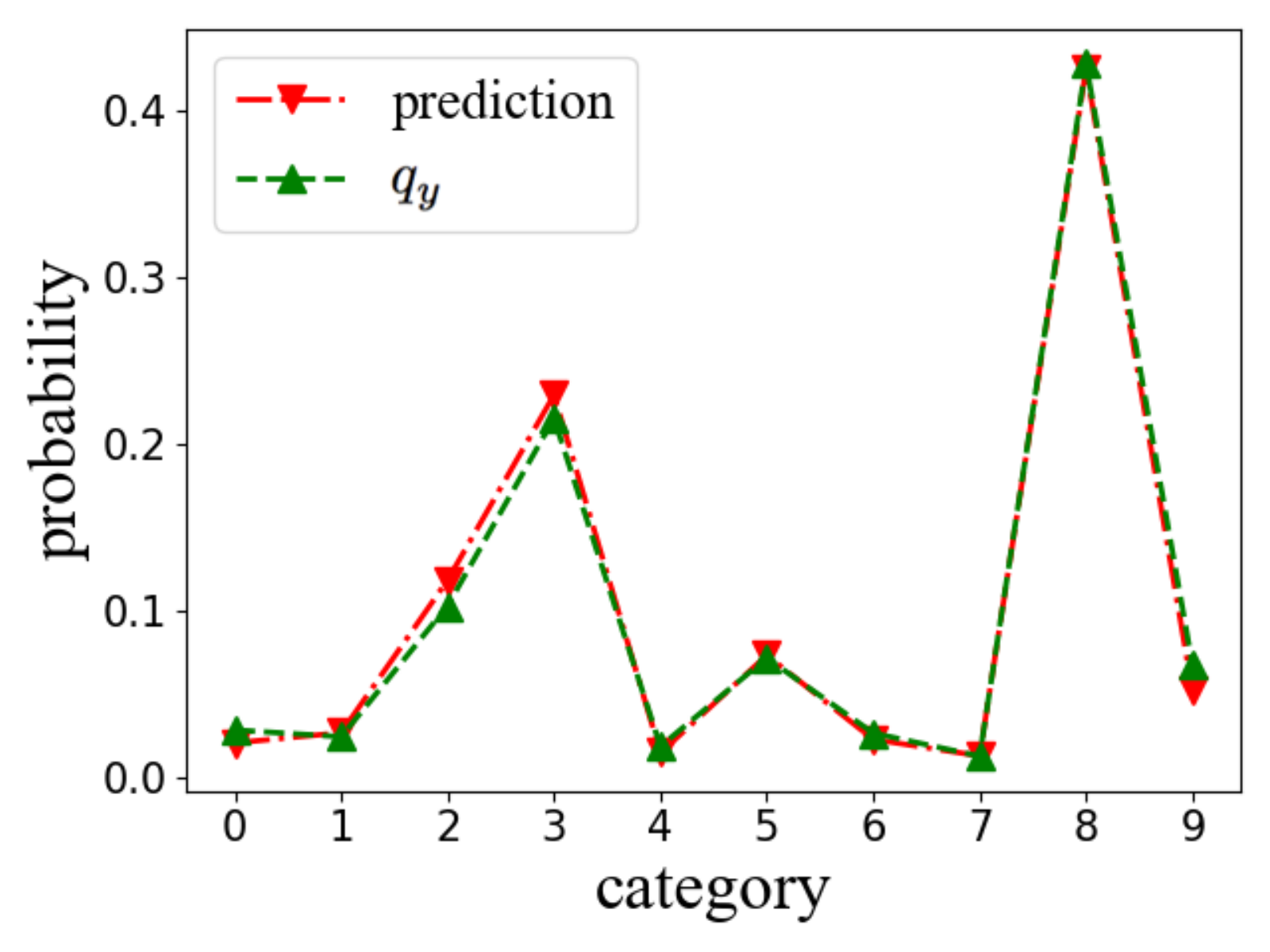}}}
    \subfigure[$\beta=0.4$, digit of $9$]{
    \centering{\includegraphics[width=0.22\columnwidth]{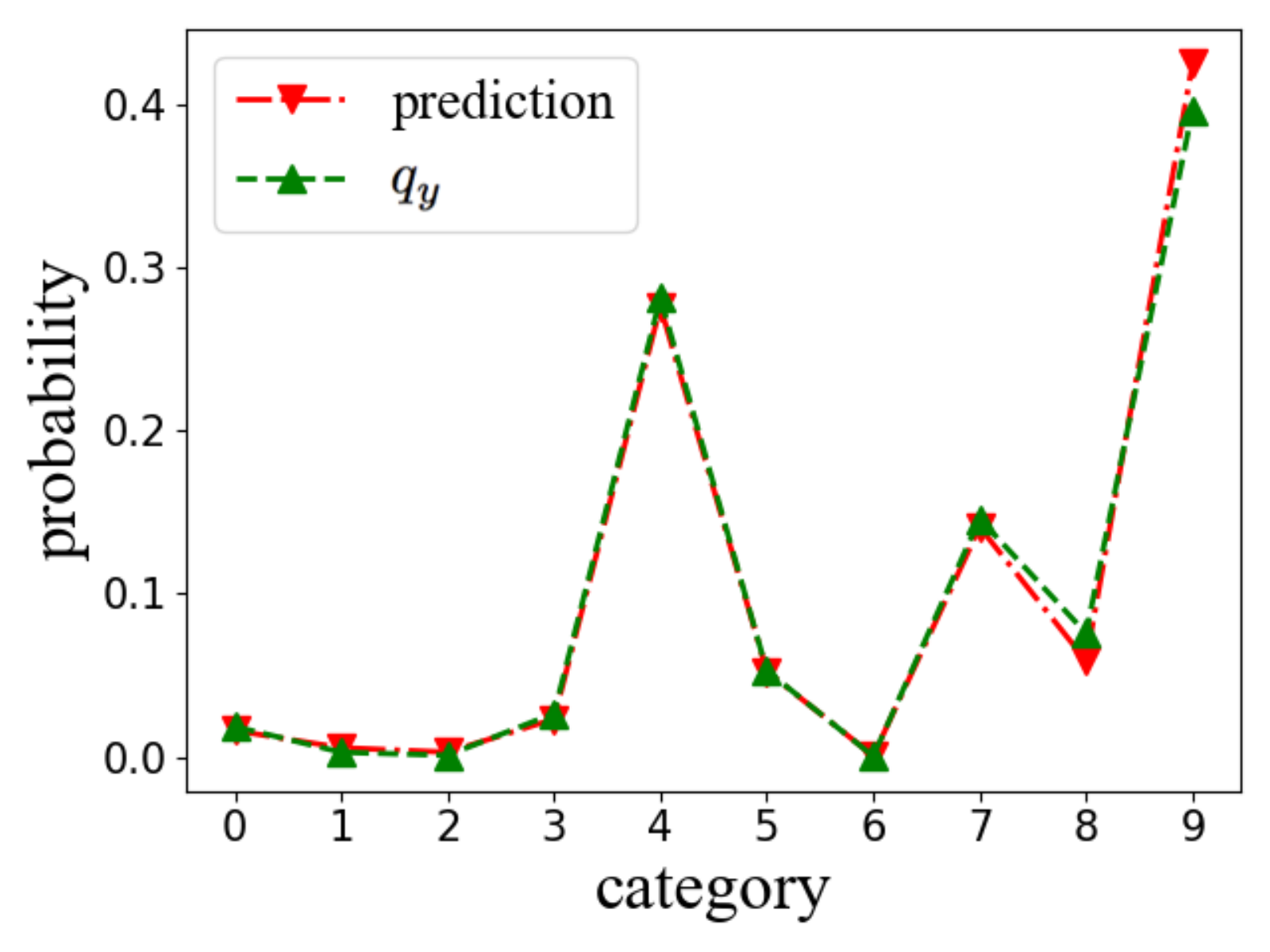}}}
    \subfigure[$\beta=0.4$, instance]{
    \centering{\includegraphics[width=0.22\columnwidth]{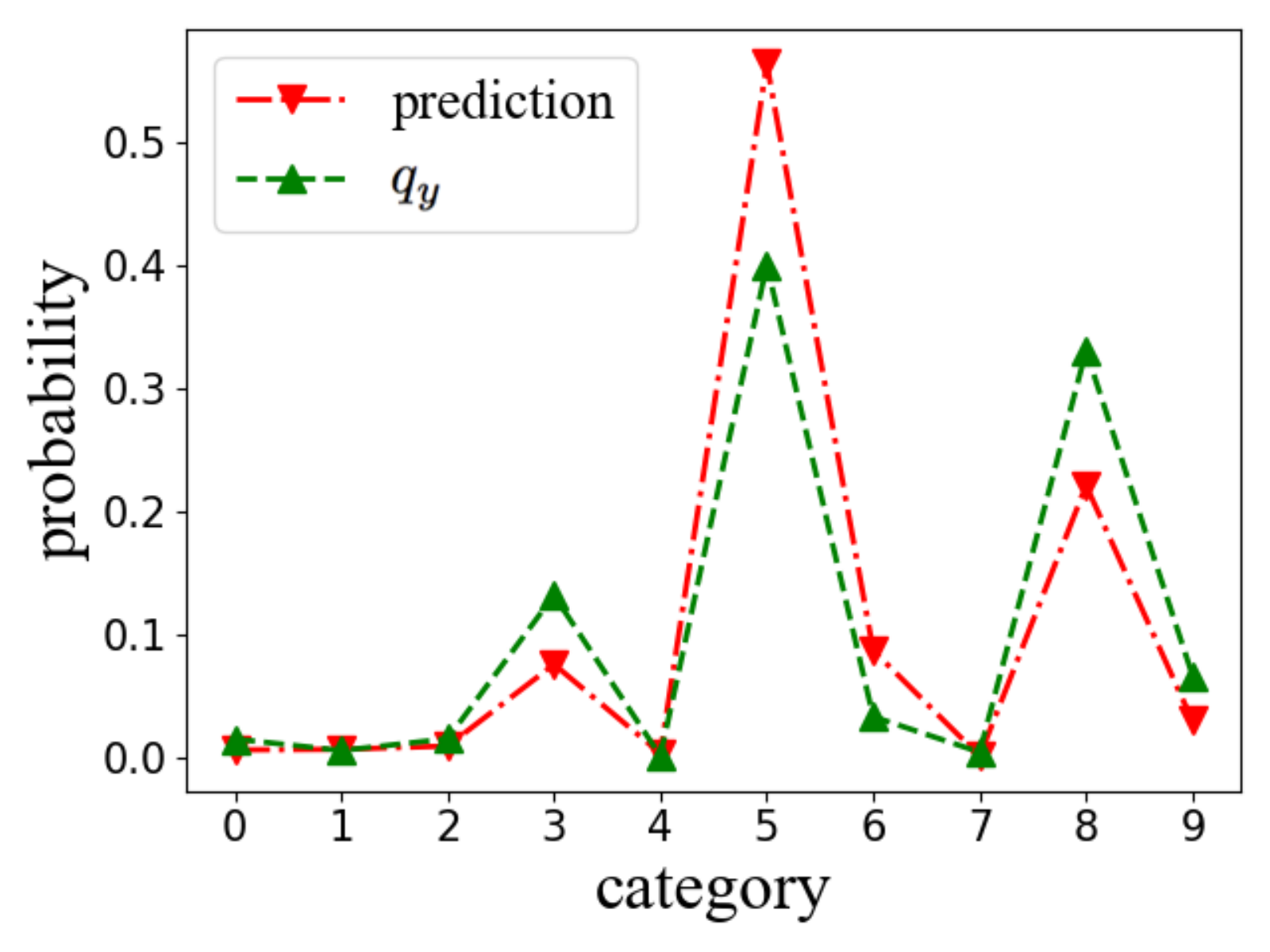}}}
    \subfigure[$\beta=0.4$, instance]{
    \centering{\includegraphics[width=0.22\columnwidth]{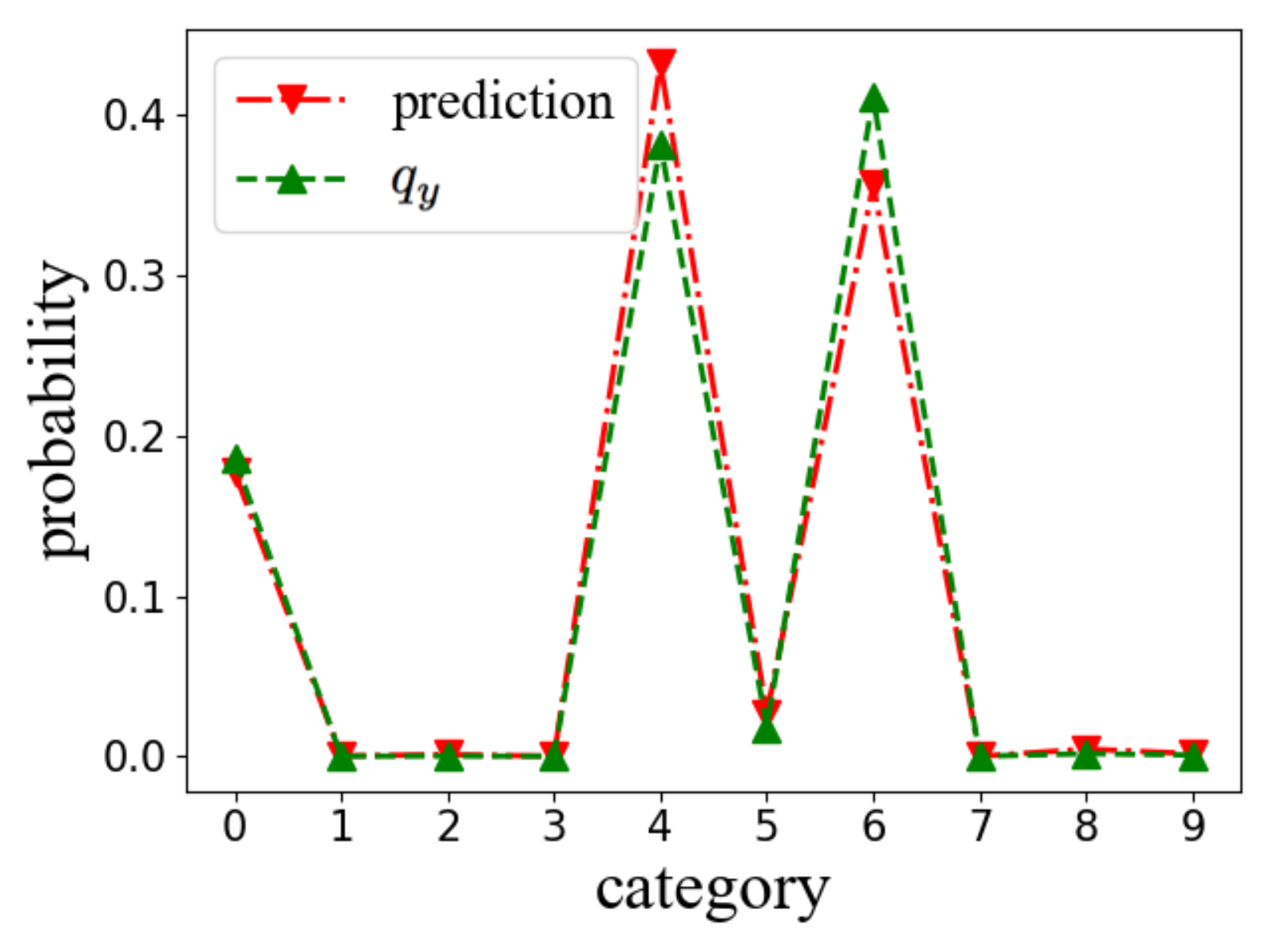}}}
    
    \subfigure[$\beta=0.8$, digit of $0$]{
    \centering{\includegraphics[width=0.22\columnwidth]{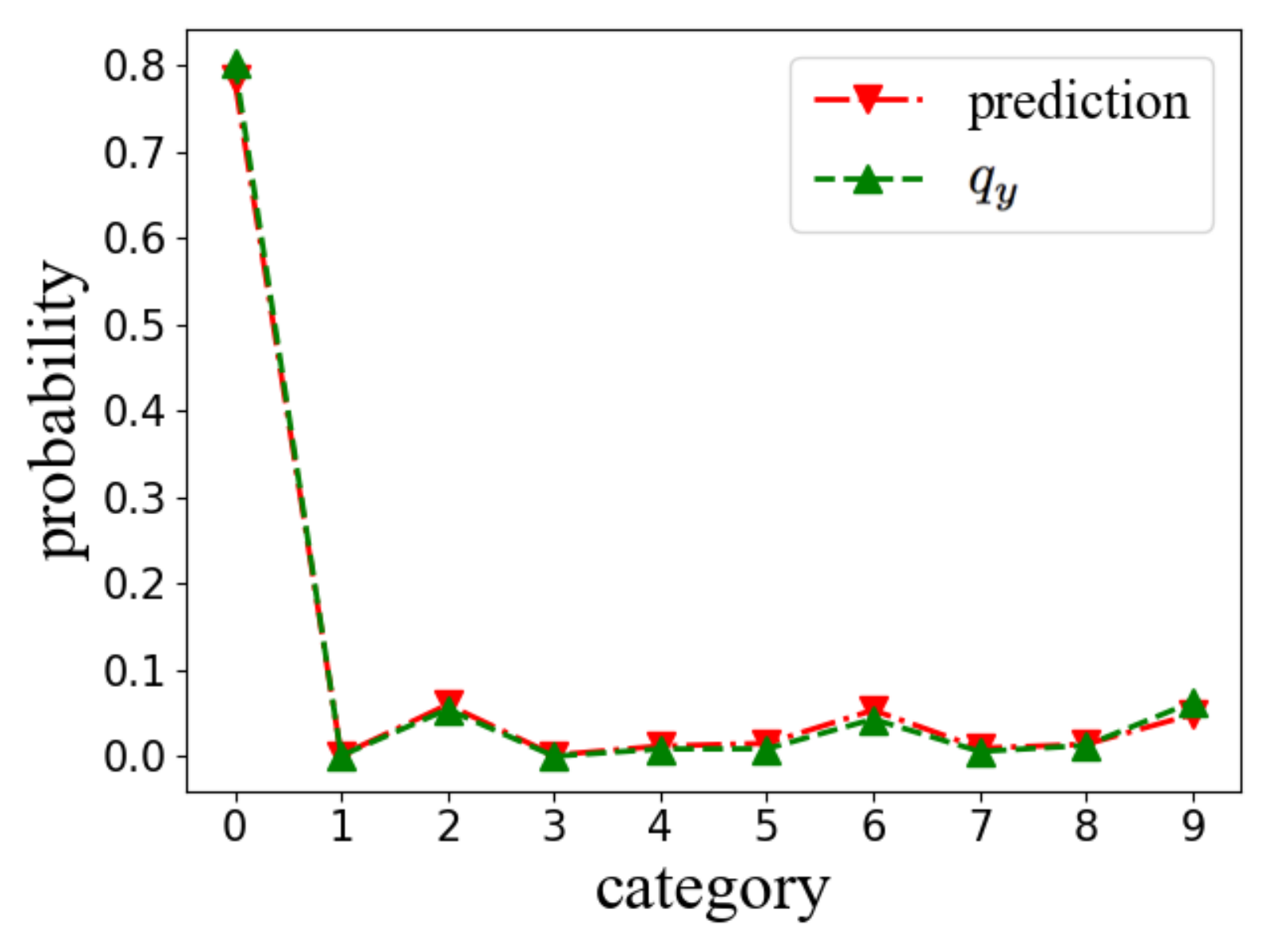}}}
    \subfigure[$\beta=0.8$, digit of $1$]{
    \centering{\includegraphics[width=0.22\columnwidth]{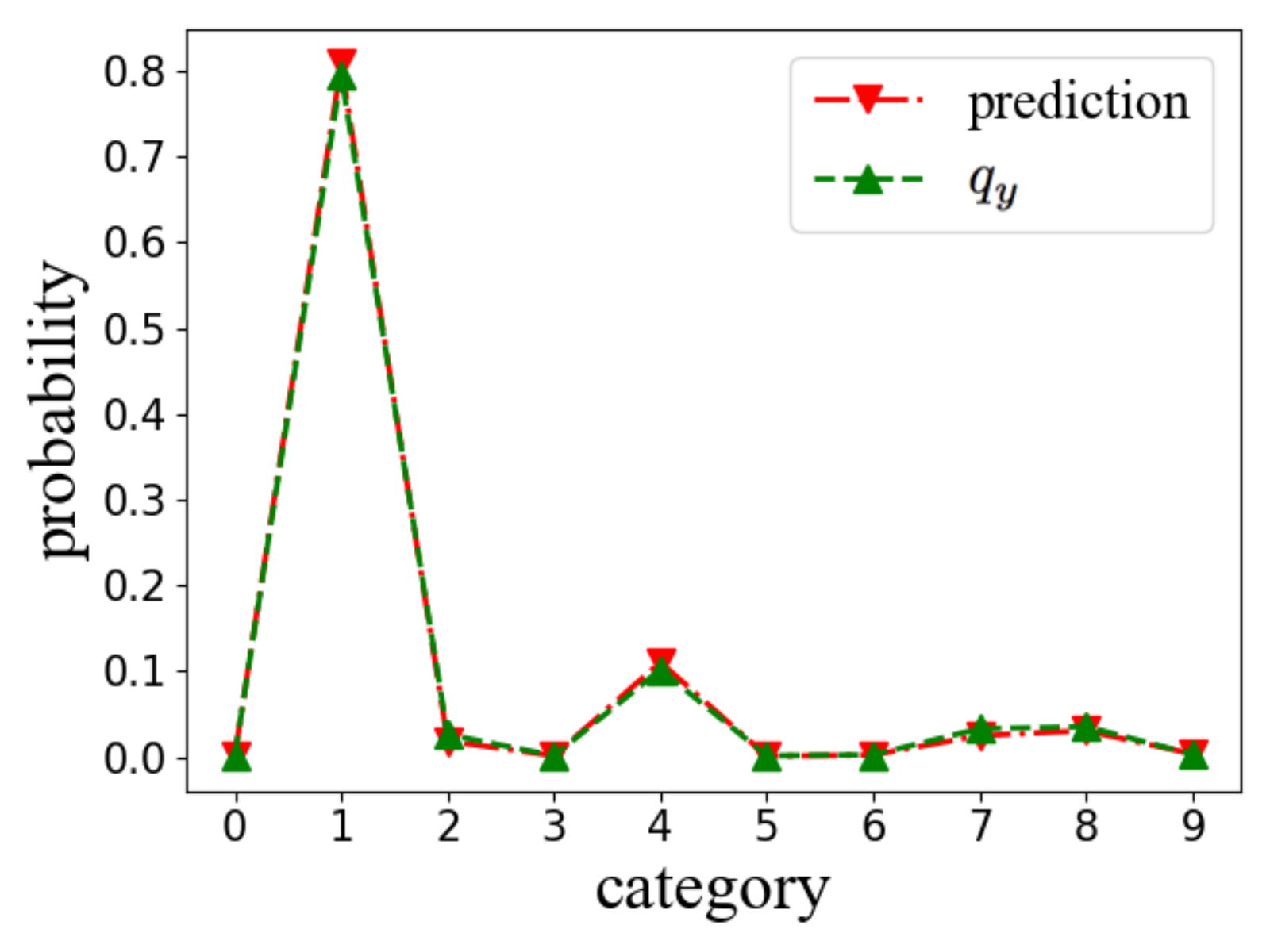}}}
    \subfigure[$\beta=0.8$, digit of $2$]{
    \centering{\includegraphics[width=0.22\columnwidth]{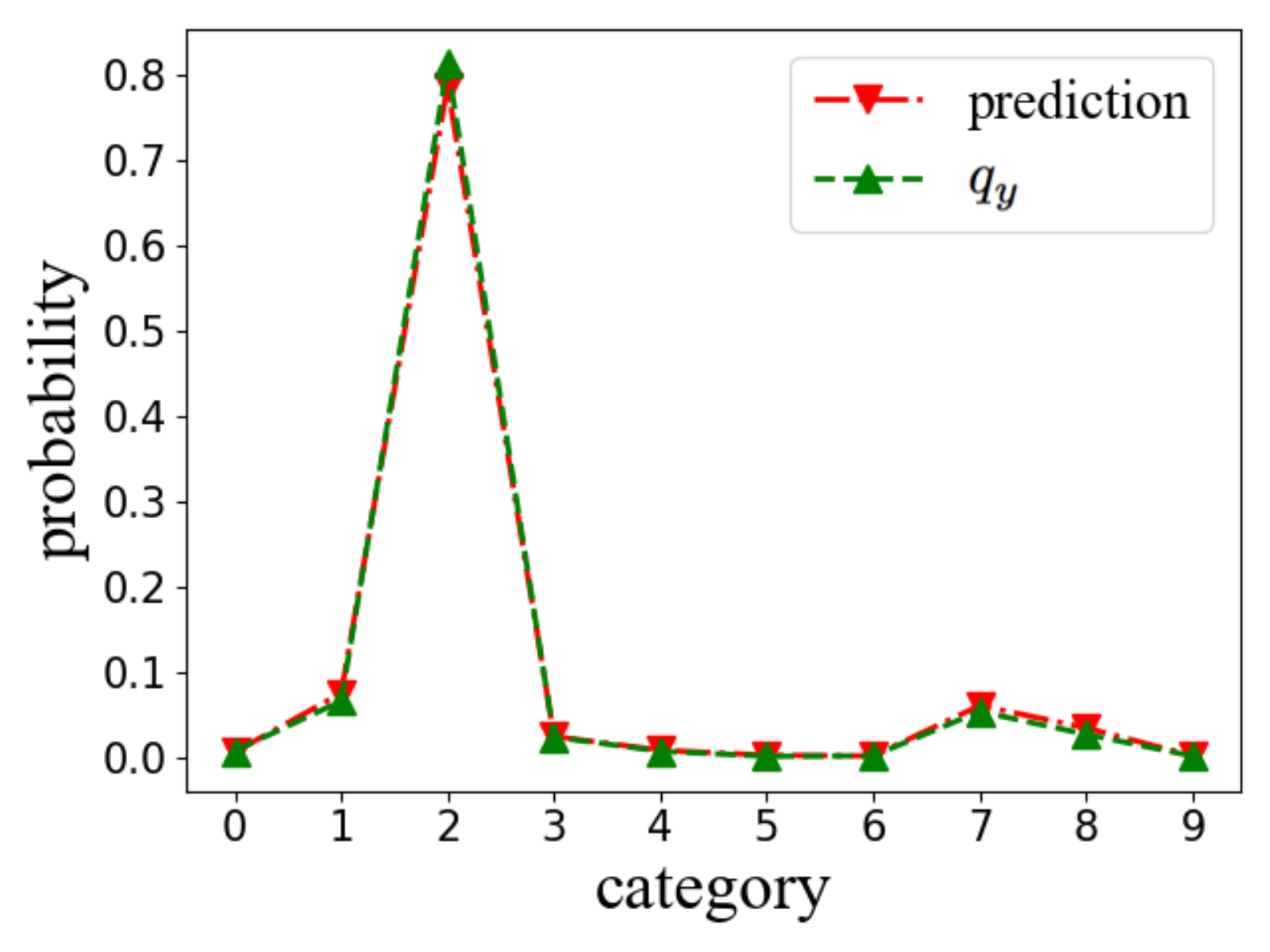}}}
    \subfigure[$\beta=0.8$, digit of $3$]{
    \centering{\includegraphics[width=0.22\columnwidth]{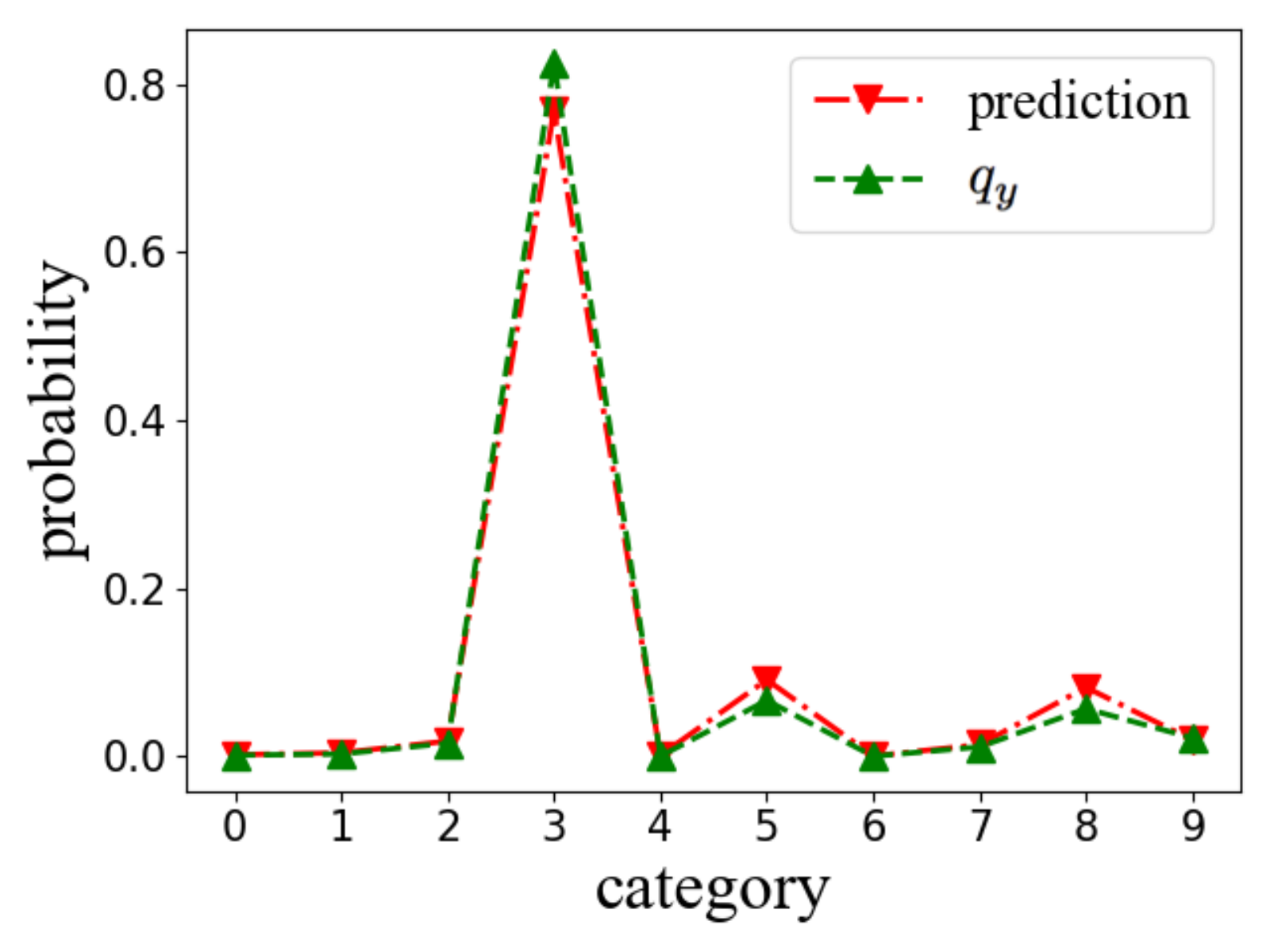}}}
    
    \subfigure[$\beta=0.8$, digit of $4$]{
    \centering{\includegraphics[width=0.22\columnwidth]{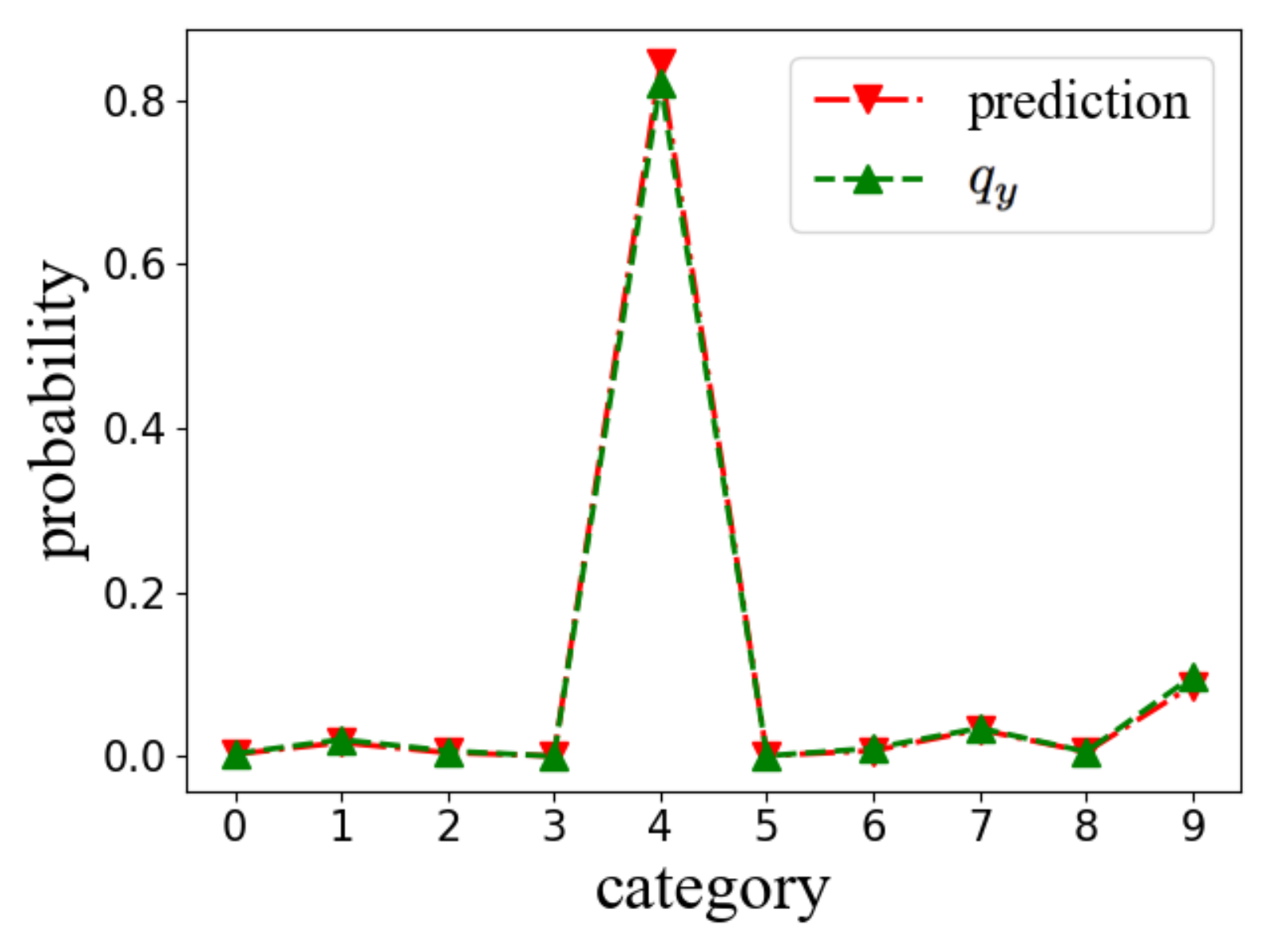}}}
    \subfigure[$\beta=0.8$, digit of $5$]{
    \centering{\includegraphics[width=0.22\columnwidth]{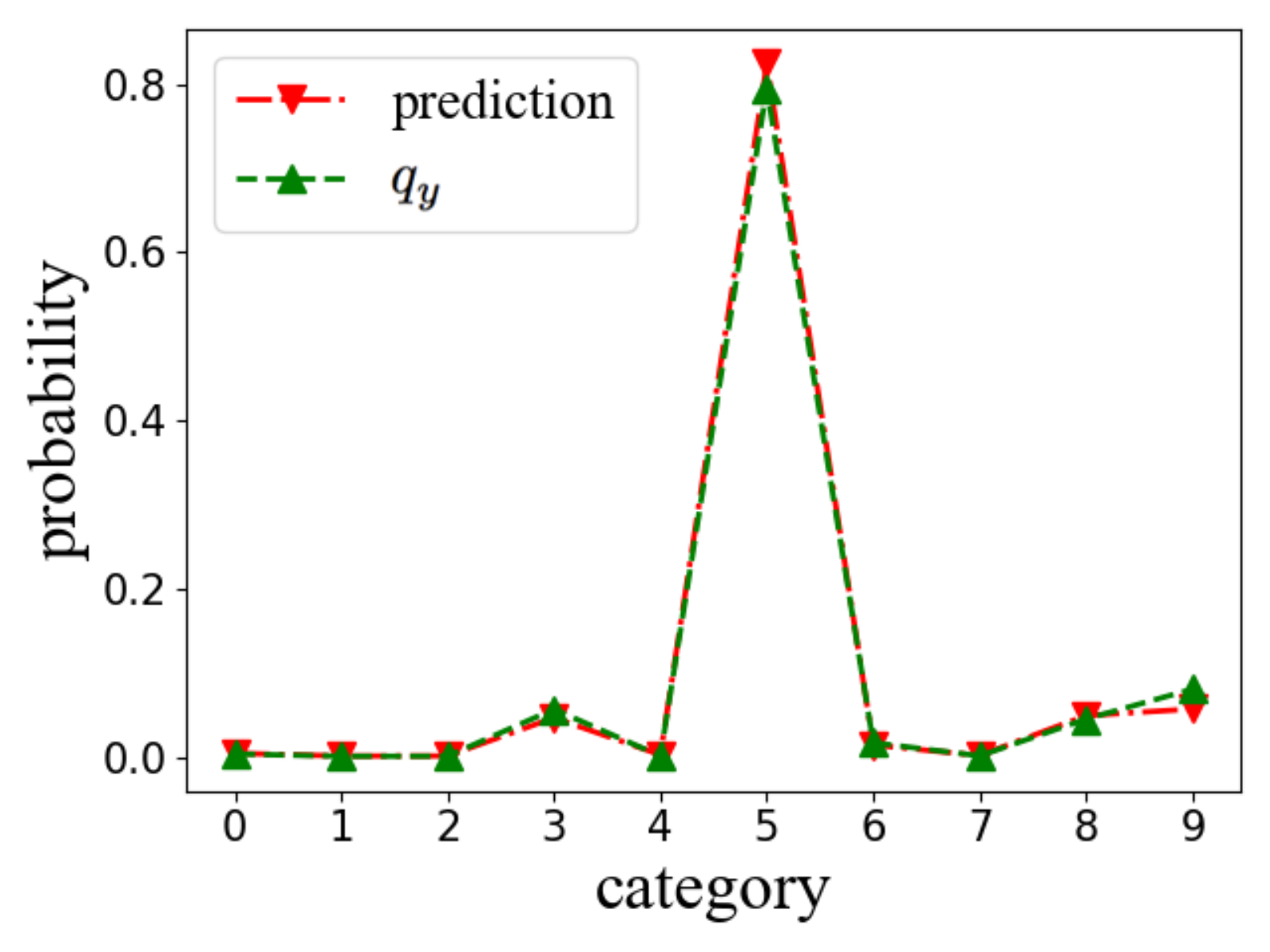}}}
    \subfigure[$\beta=0.8$, digit of $6$]{
    \centering{\includegraphics[width=0.22\columnwidth]{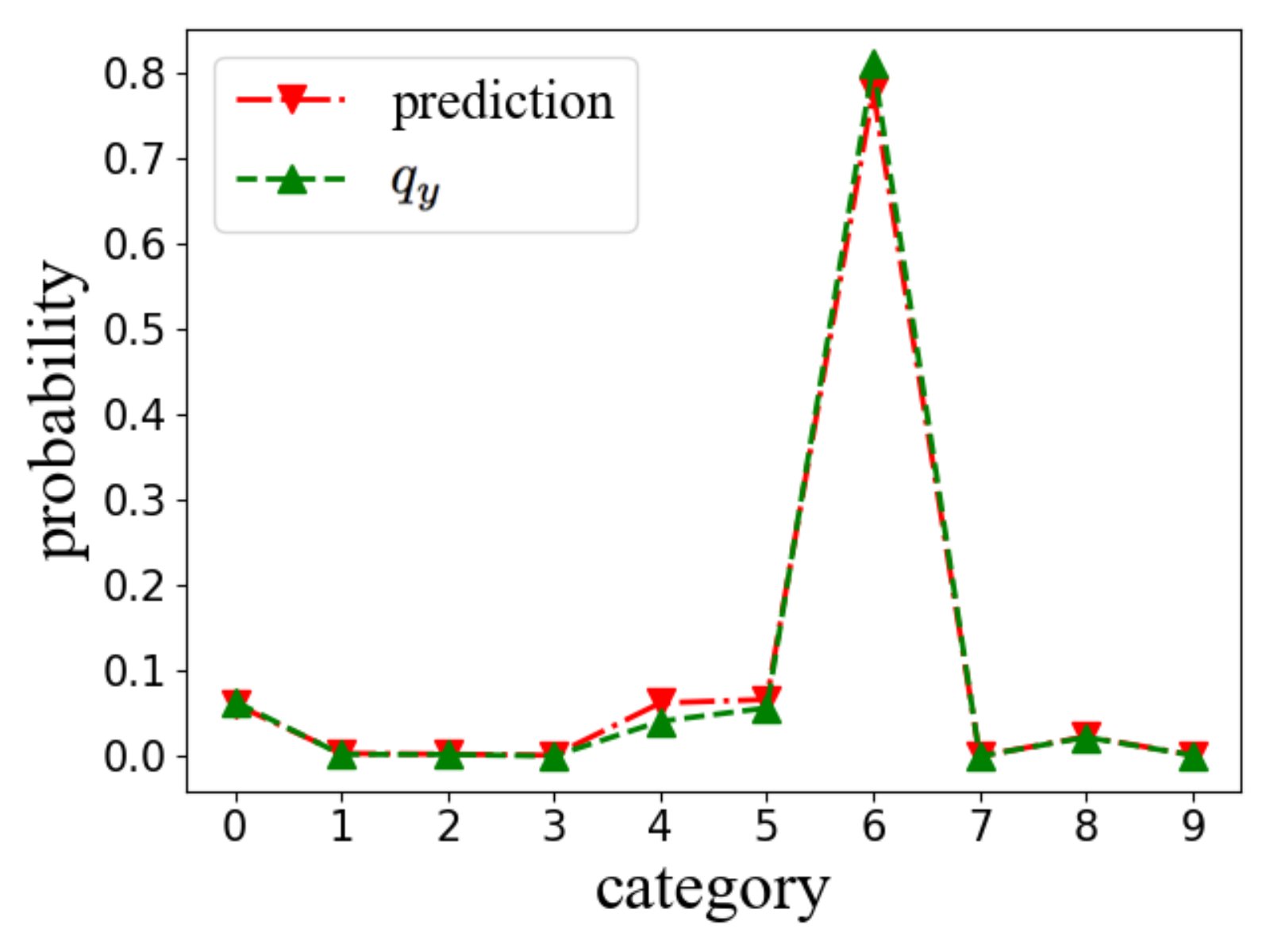}}}
    \subfigure[$\beta=0.8$, digit of $7$]{
    \centering{\includegraphics[width=0.22\columnwidth]{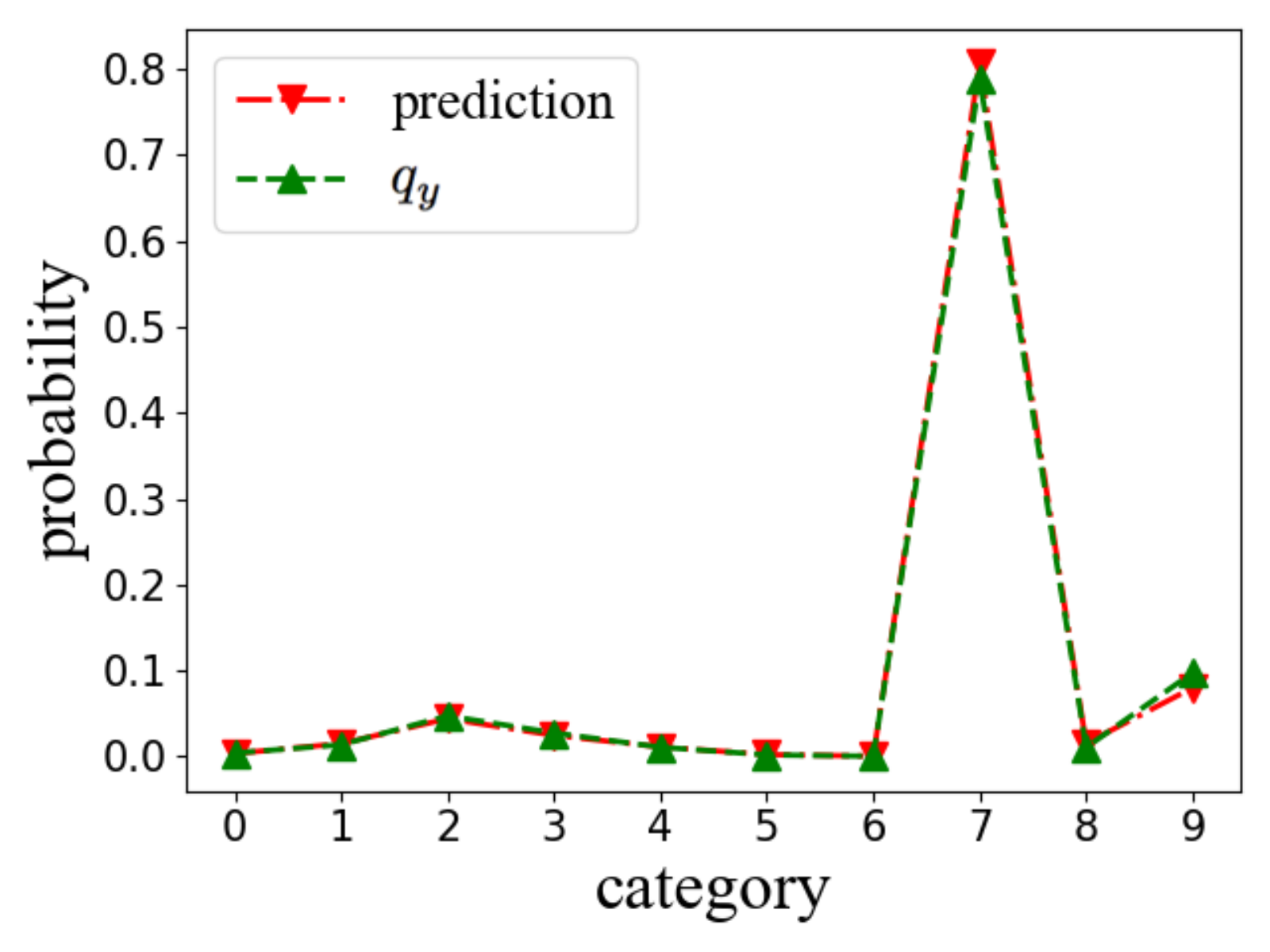}}}
    
    \subfigure[$\beta=0.8$, digit of $8$]{
    \centering{\includegraphics[width=0.22\columnwidth]{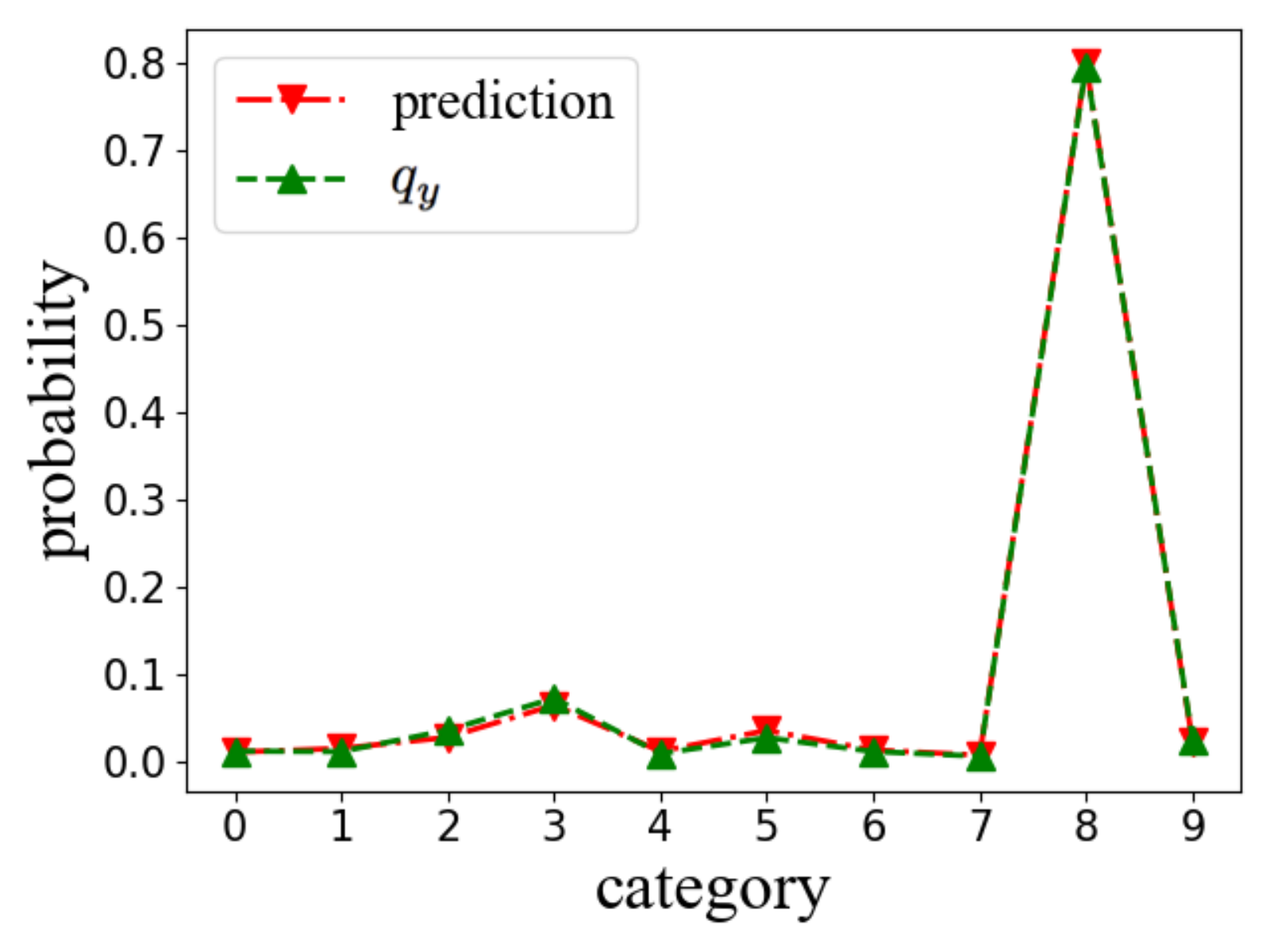}}}
    \subfigure[$\beta=0.8$, digit of $9$]{
    \centering{\includegraphics[width=0.22\columnwidth]{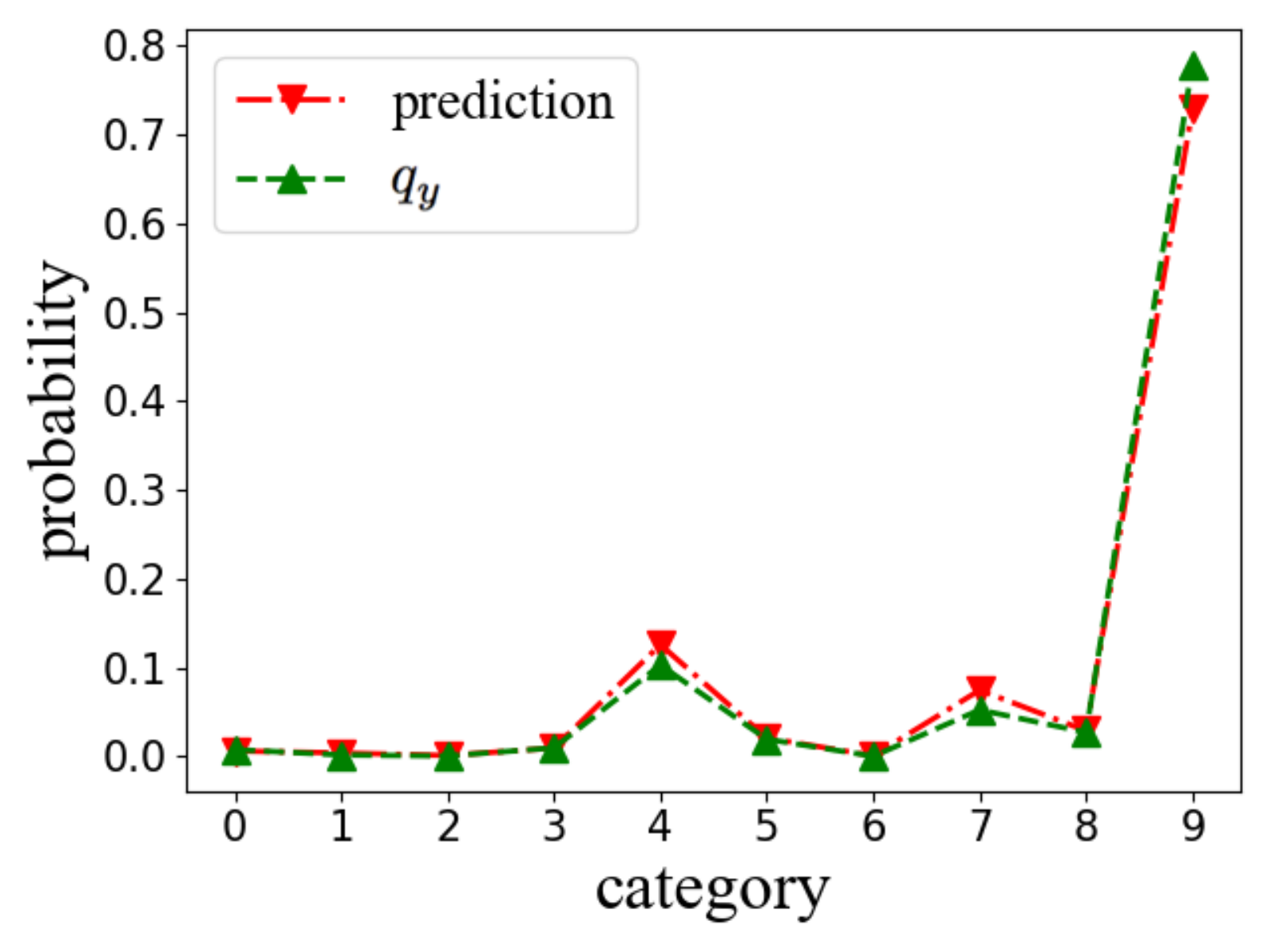}}}
    \subfigure[$\beta=0.8$, instance]{
    \centering{\includegraphics[width=0.22\columnwidth]{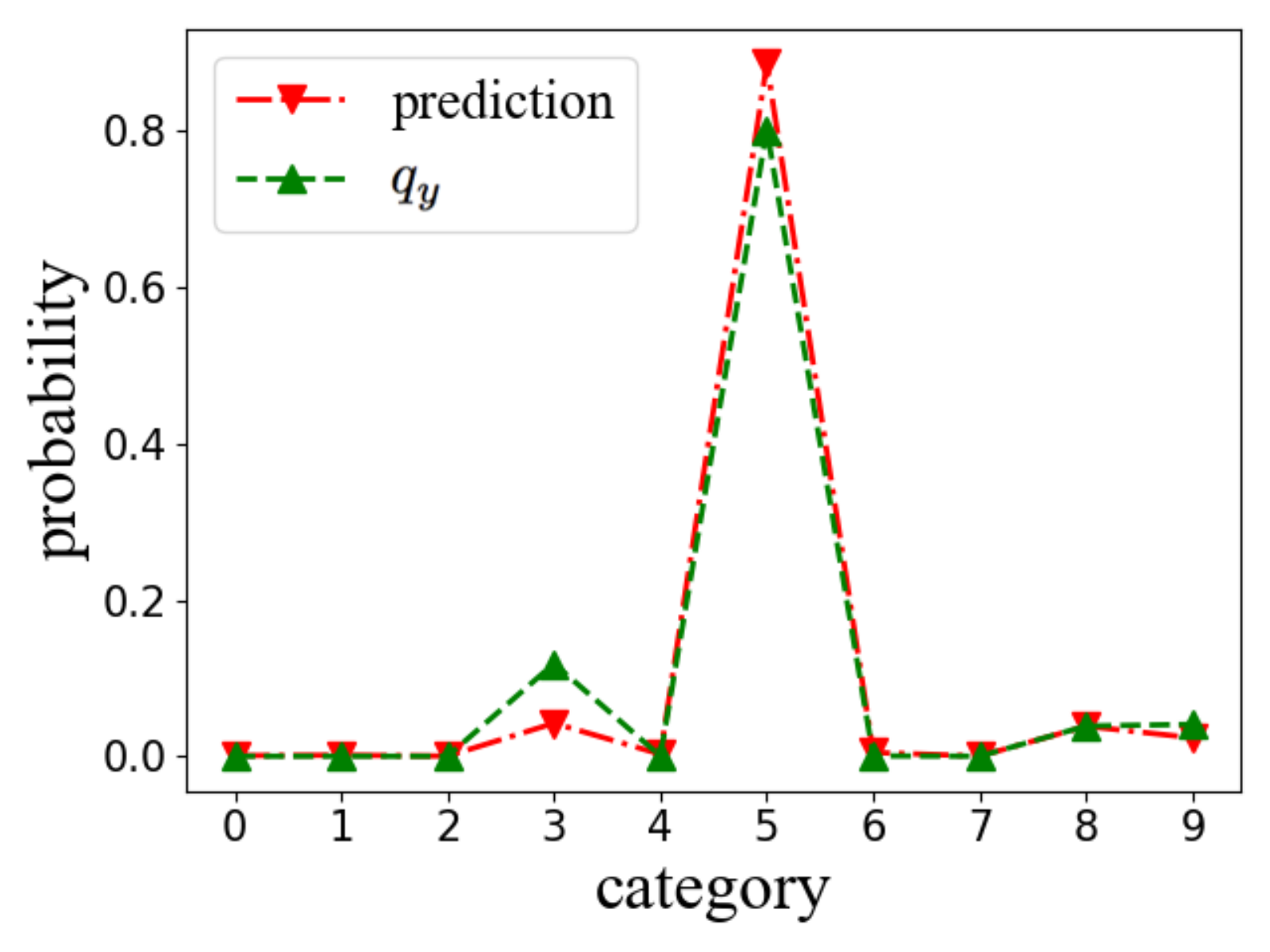}}}
    \subfigure[$\beta=0.8$, instance]{
    \centering{\includegraphics[width=0.22\columnwidth]{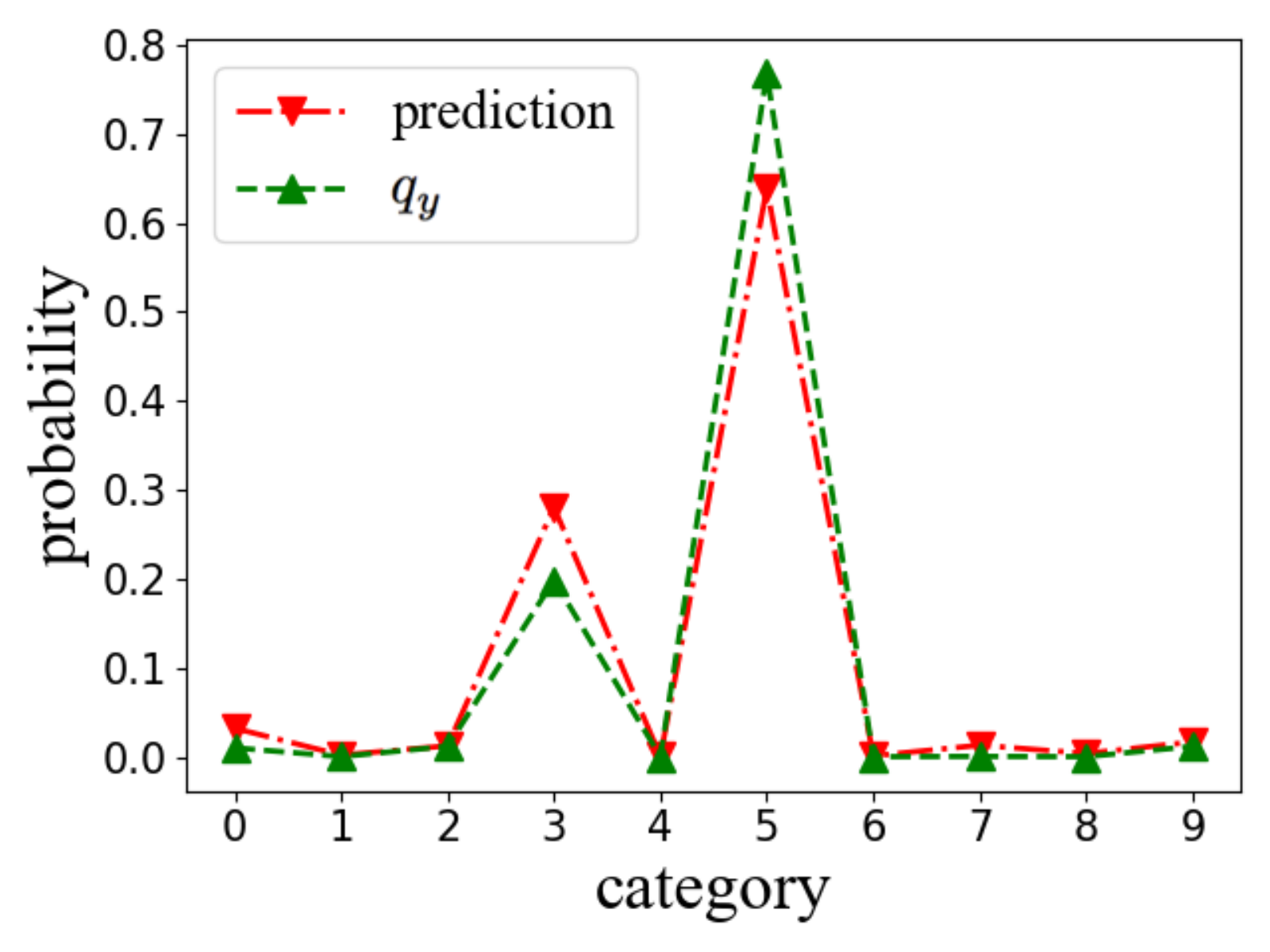}}}

\caption{Generated noise when $\beta=0.4$ and $\beta=0.8$. (k), (l), (w) and (x): instances of digit $5$. Others are the average of predicted probabilities and $q_y$ for all images of different digits. The red lines represent $f(y; \theta)$ whereas the green lines represent ${q}_{y}$.}
\label{fig:generated_noise}
\end{center}
\end{figure}

We show more comparison between predictive probability and $q_y$ for uniform noise, biased noise and generated noise in \cref{fig:uniform_noise}, \cref{fig:biased_noise} and \cref{fig:generated_noise}, respectively.

\subsection{Pixel-wise Variance Estimation of Single Image Super-resolution}
\label{sec:more_results_sr}

\begin{figure}[t]
\begin{center}
\centerline{\includegraphics[width=\columnwidth]{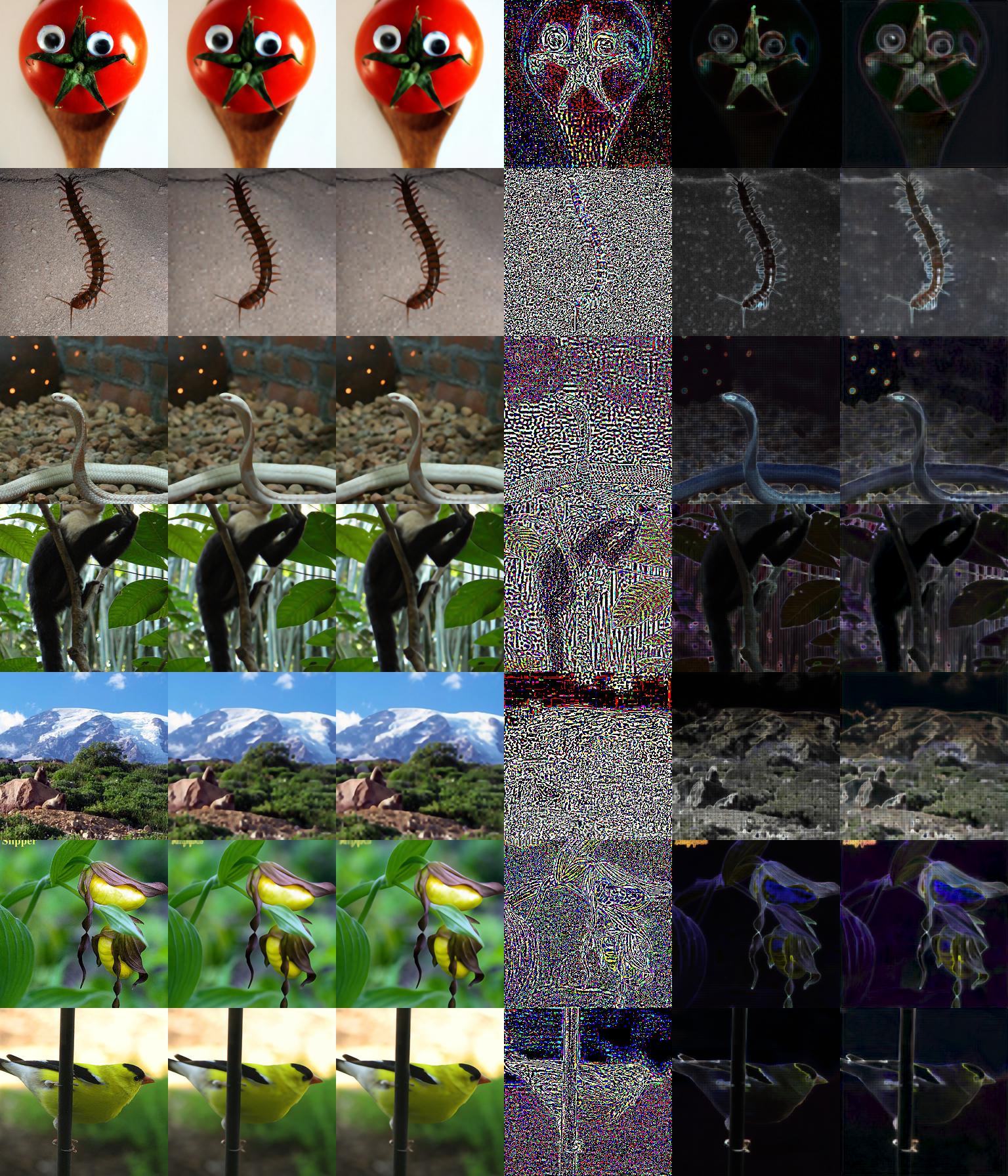}}
\caption{From left to right: the high-resolution image $x$ (column 1); estimated $\mathbb{E}_{x\mid y}\left[x\right]$ by DDPM-SR; (column 2); estimated $\mathbb{E}_{x\mid y}\left[x\right]$ by $f_{\mathrm{mean}}$ (column 3); square root of label when training $f_{\mathrm{var}}$, $\left| x - \mathbb{E}_{x\mid y}\left[x\right] \right|$ (column 4); estimated pixel-wise variance by DDPM-SR (column 5); estimated pixel-wise variance by $f_{\mathrm{var}}$ (column 6).}
\label{fig:super-resolution_samples_1}
\end{center}
\end{figure}

\begin{figure}[t]
\begin{center}
\centerline{\includegraphics[width=\columnwidth]{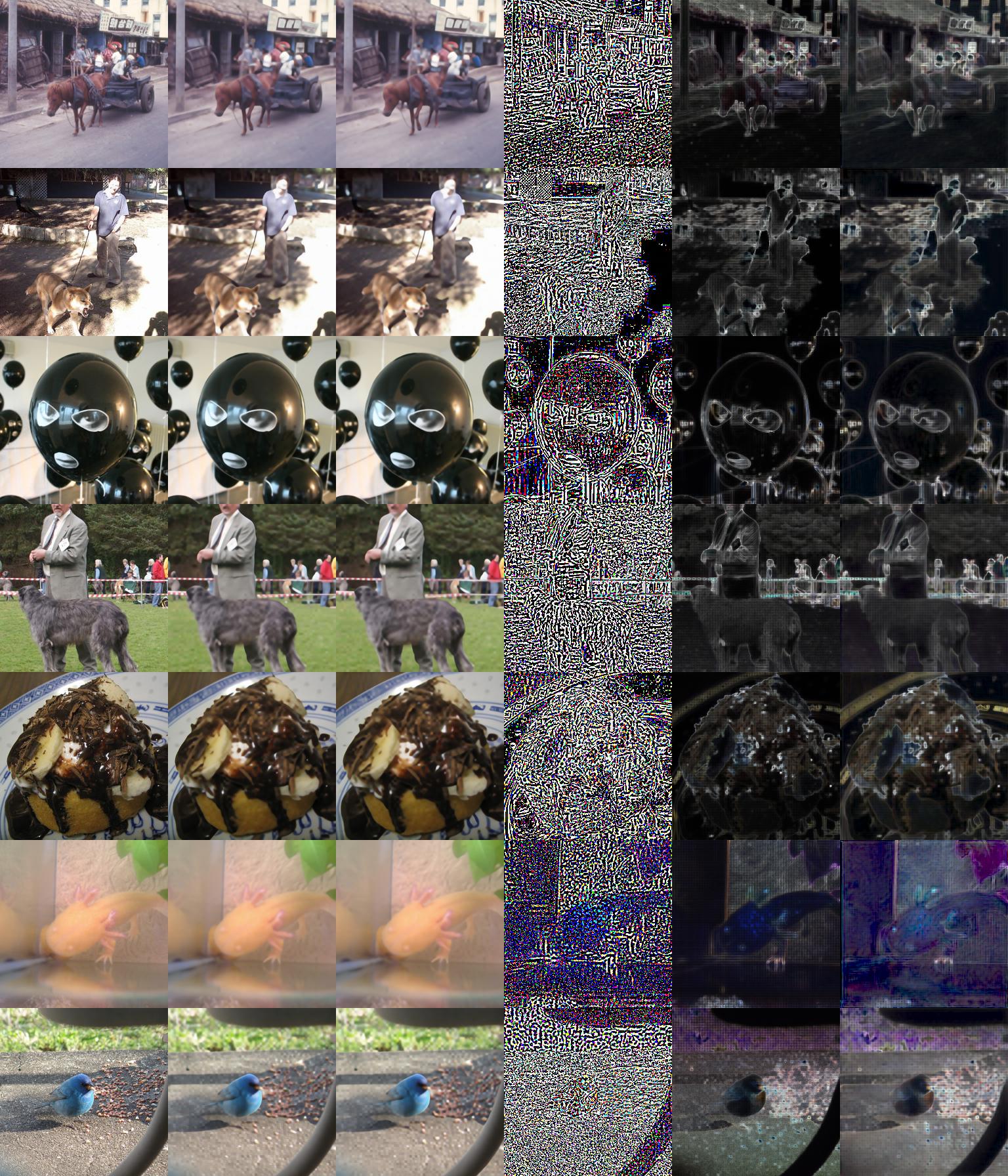}}
\caption{From left to right: the high-resolution image $x$ (column 1); estimated $\mathbb{E}_{x\mid y}\left[x\right]$ by DDPM-SR; (column 2); estimated $\mathbb{E}_{x\mid y}\left[x\right]$ by $f_{\mathrm{mean}}$ (column 3); square root of label when training $f_{\mathrm{var}}$, $\left| x - \mathbb{E}_{x\mid y}\left[x\right] \right|$ (column 4); estimated pixel-wise variance by DDPM-SR (column 5); estimated pixel-wise variance by $f_{\mathrm{var}}$ (column 6).}
\label{fig:super-resolution_samples_2}
\end{center}
\end{figure}

\begin{figure}[t]
\begin{center}
\centerline{\includegraphics[width=\columnwidth]{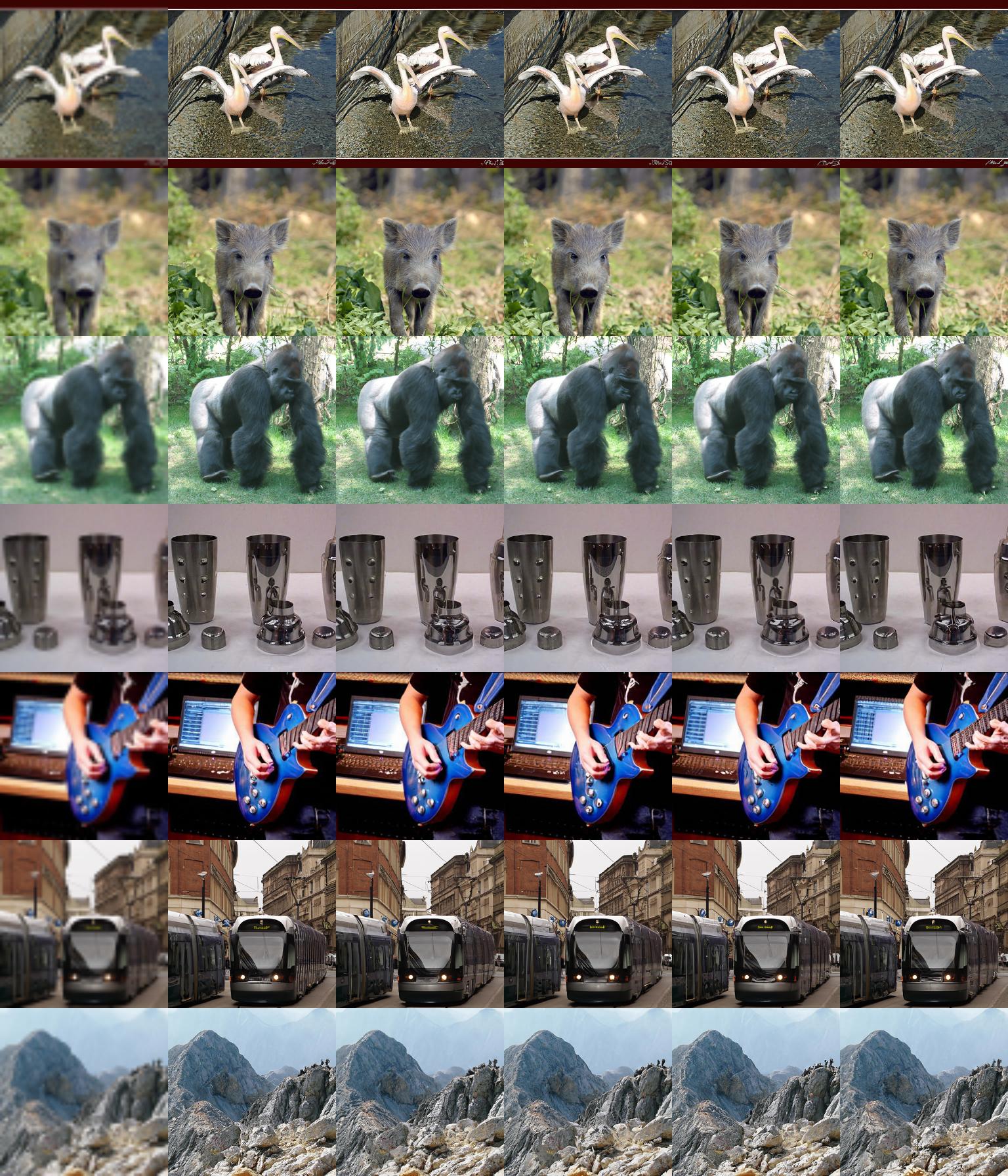}}
\caption{The first column is the upsampled low-resolution images from resolution of $64 \times 64$ to $256 \times 256$. Other columns are high-resolution images generated by DDPM-SR. We can observe the difference of details among generated images, such as grass, rock, eyes, windows and so on.}
\label{fig:super-resolution_samples_ddpm}
\end{center}
\end{figure}

We show more results of DDPM-SR, $f_{\text{mean}}$ and $f_{\text{var}}$ in \cref{fig:super-resolution_samples_1} and \cref{fig:super-resolution_samples_2}. To illustrate the image quality of DDPM-SR, we show some super-resolution samples in \cref{fig:super-resolution_samples_ddpm}.

\end{document}